\documentclass[journal]{IEEEtran}

\AtBeginDocument{%
  \providecommand\BibTeX{{%
    \normalfont B\kern-0.5em{\scshape i\kern-0.25em b}\kern-0.8em\TeX}}}

\usepackage{tikz}
\usepackage{amsmath,amssymb,graphicx}
\usepackage{caption}
\usepackage{subcaption}
\usepackage{algorithmic}
\usepackage{algorithm}
\usepackage{listings}
\usepackage{mdframed}
\usepackage[numbers,square]{natbib}
\usepackage{url}
\usepackage{breqn}
\usepackage{textcomp}

\usepackage{version}

\newif\ifdevel
\develfalse

\newcommand{\textnote}[2][]{\ifdevel\textit{\textcolor{green}{(\ifthenelse{\equal{#1}{}}{anonym}{#1})} \textcolor{blue}{#2$\triangleleft$}}\fi}

\newcommand{\oldwork}[1]{}

\newif\iftrackchanges
\trackchangestrue

\ifdevel
\newenvironment{potentialRemove}{\color{red}}{}
\else

\fi

\excludeversion{doRemove}

\newcommand{\emphBulletPoint}[1]{\textbf{#1}}

\renewcommand{\vec}[1]{\boldsymbol{#1}}
\newcommand{\matr}[1]{\boldsymbol{#1}}

\DeclareMathOperator*{\argmax}{arg\,max}
\DeclareMathOperator*{\argmin}{arg\,min}
\newcommand{\NN}[0]{\Phi}
\newcommand{\nl}[0]{s}
\newcommand{\sel}[0]{\tau}

\newcommand{\etal}{\textit{et al}. }
\newcommand{\ie}{\textit{i}.\textit{e}., }
\newcommand{\eg}{\textit{e}.\textit{g}., }

\newcommand{\expec}[2]{\mathrm{E}_{#1}\left[ #2 \right]}

\definecolor{pgmBgColor}{rgb}{0.7,0.87,0.95}
\newcommand{\pgmBgColor}[0]{pgmBgColor}
\definecolor{pgmLineColor}{rgb}{0.1,0.4,0.6}
\newcommand{\pgmLineColor}[0]{pgmLineColor}

\newcommand{\Bernoulli}{\text{Bernoulli}}

\makeatletter
\def\@firstoftwo@second#1#2{%
  \def\temp##1.##2\@nil{##2}%
   \temp#1\@nil}
\newcommand\sref[1]{%
   \expandafter\@setref\csname r@#1\endcsname\@firstoftwo@second{#1}%
}

\makeatother

\begin{document}

\title{Hands-on Bayesian Neural Networks -- A Tutorial for Deep Learning Users}

\author{Laurent Valentin Jospin\thanks{Corresponding author: Laurent Valentin Jospin (Email: laurent.jospin@research.uwa.edu.au)}, University of Western Australia, Australia \\
Hamid Laga, Murdoch University, Australia
\\Farid Boussaid, University of Western Australia, Australia
\\Wray Buntine, Monash University, Australia
\\Mohammed Bennamoun, University of Western Australia, Australia}

\newcommand{\shortauthors}{Jospin et al.}

\maketitle

\begin{abstract}
Modern deep learning methods constitute incredibly powerful tools to tackle a myriad of challenging problems. However, since deep learning methods operate as black boxes, the uncertainty associated with their predictions is often challenging to quantify. Bayesian statistics offer a formalism to understand and quantify the uncertainty associated with deep neural network predictions. This tutorial provides deep learning practitioners with an overview of the relevant literature and a complete toolset to design, implement, train, use and evaluate Bayesian neural networks, \ie stochastic artificial neural networks trained using Bayesian methods.
\end{abstract}


\section{introduction}

Deep learning has led to a revolution in machine learning, providing solutions to tackle problems that were traditionally difficult to solve. However, deep learning models are prone to overfitting, which adversely affects their generalization capabilities \cite{szegedy2013intriguing}. They also tend to be overconfident about their predictions when they provide a confidence interval. This is problematic for applications where silent failures can lead to dramatic outcomes, \eg autonomous driving \cite{10.1145/3194085.3194087}, medical diagnosis \cite{8241753} or finance \cite{CAVALCANTE2016194}. Consequently, many approaches have been proposed to mitigate this risk \cite{8371683}. Among them, the Bayesian paradigm provides a rigorous framework to analyze and train uncertainty-aware neural networks, and more generally, to support the development of learning algorithms.

The Bayesian paradigm in statistics contrasts with the frequentist paradigm, with a major area of distinction in hypothesis testing \cite{etzBayes}. It is based on two simple ideas. The \textbf{first} is that probability is a measure of belief in the occurrence of events, rather than the limit in the frequency of occurrence when the number of samples goes toward infinity, as assumed in the frequentist paradigm. The \textbf{second} idea is that prior beliefs influence posterior beliefs. Bayes' theorem, which states that:
\begin{equation}
    \label{eq:bayes}
    P(H|D) = \dfrac{P(D|H)P(H)}{P(D)} = \dfrac{P(D, H)}{\int_{H} P(D, H')dH'},
\end{equation}

\noindent summarizes this interpretation. Formula \eqref{eq:bayes} is still true in the frequentist interpretation, where $H$ and $D$ are considered as sets of outcomes. The Bayesian interpretation considers $H$ to be a hypothesis about which one holds some prior belief, and $D$ to be some data that will update one's belief about $H$. The probability distribution $P(D|H)$ is called the likelihood. It encodes the aleatoric uncertainty in the model, \ie the uncertainty due to the noise in the process. $P(H)$ is the prior and $P(D) = \int_{H} P(D, H')dH'$ the evidence. $P(H|D)$ is called the posterior. It encodes the epistemic uncertainty, \ie the uncertainty due to the lack of data. $P(D|H)P(H) = P(D,H)$ is the joint probability of $D$ and $H$. 

\begin{figure}[t]
    \centering
    \includegraphics[width=0.5\textwidth]{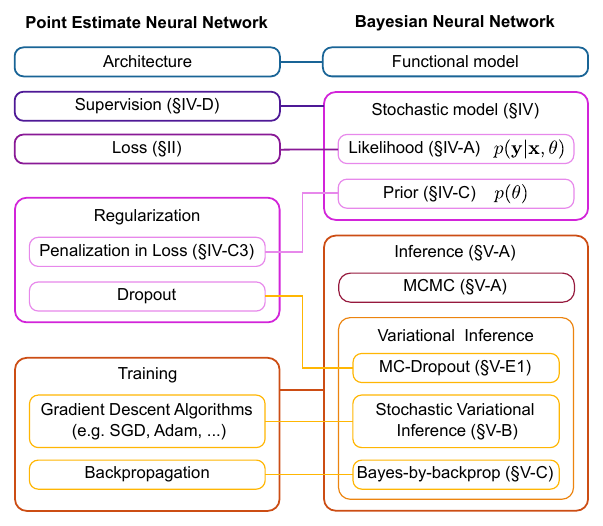}
    \caption{Illustration of the correspondence between the concepts used in deep learning for point-estimate neural networks and their counterparts in Bayesian neural networks (BNNs).}
    \label{fig:correspondances}
\end{figure}

\begin{figure*}[t]
    \centering
    
    \includegraphics{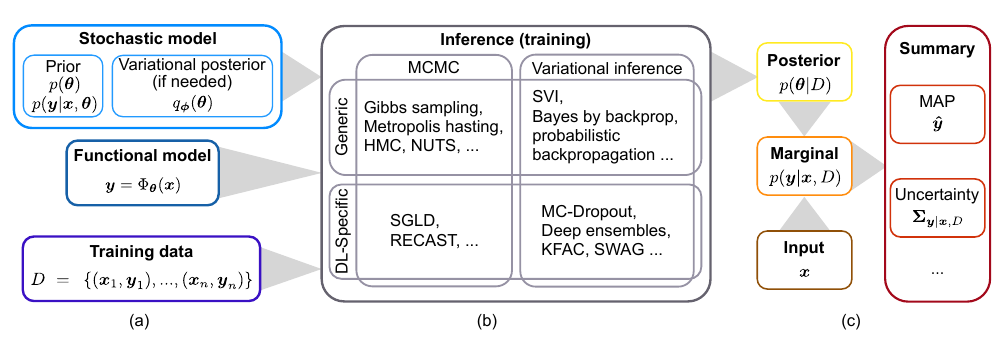}
    \caption{Workflow to design (a), train (b) and use a BNN for predictions (c).}
    \label{fig:workflow}
\end{figure*}

Using Bayes' formula to train a predictor can be understood as learning from the data $D$. In other words, the Bayesian paradigm not only offers a solid approach for the quantification of uncertainty in deep learning models but also provides a mathematical framework to understand many regularization techniques and learning strategies that are already used in classic deep learning \cite{polson2017deep} (Section~\ref{sec:reg-prior}).

Bayesian neural networks (BNNs)~\cite{LampinenDLReview2001, TitteringtonDL2004, Goan2020} are stochastic neural networks trained using a Bayesian approach. There is  a rich literature about BNNs and the related field of Bayesian deep learning, which is referred to by Wang and Yeung~\cite{WangDLSurvey2016} as the conjoint use of deep learning for perception and traditional Bayesian models for inference.\footnote{Note that some other authors use a different definition of Bayesian deep learning, which is closer to the idea of a BNN \cite{WilsonDL2020}).} However, navigating through this literature is challenging without some prior background in Bayesian statistics. This brings an additional layer of complexity for deep learning practitioners interested in building and using BNNs.

This paper, conceived as a tutorial, presents a unified workflow to design, implement, train and evaluate a BNN (Figure~\ref{fig:workflow}). It also provides an overview of the relevant literature where a large number of approaches have been developed to efficiently train and use BNNs. A good knowledge of those different methods is a prerequisite for an efficient use of BNNs in big data applications of deep learning. In this tutorial, we assume that the reader is already familiar with the concepts of traditional deep learning such as artificial neural networks, training algorithms, supervision strategies, and loss functions \cite{Goodfellow-et-al-2016}. This paper focuses on exploring the correspondences between traditional deep learning approaches and Bayesian methods (Figure~\ref{fig:correspondances}). It is intended to motivate and help researchers and students to use BNNs in measuring uncertainty for problems in their respective fields of study and research, helping them relate their existing knowledge in deep learning to the relevant Bayesian methods.

The remaining parts of this paper are organized as follows. Section \ref{sec::definitions} introduces the concept of a BNN. Section \ref{sec::motivations} presents the motivations for BNNs as well as their applications. Section \ref{sec::theory} explains how to design the stochastic model associated with a BNN. Section \ref{sec:inference} explores the most important algorithms used for Bayesian inference and how they were adapted for deep learning. Section \ref{sec:simplifications} reviews BNN simplification methods. Section \ref{sec:evaluation} presents the methods used to evaluate the performance of a BNN. Finally, Section \ref{sec::conclusion} concludes the paper. The supplementary material contains a gallery of practical examples illustrating the theoretical concepts presented in Sections \ref{sec::definitions}, \ref{sec::theory} and \ref{sec:inference} of the main paper. Each example source code is also available online on GitHub to provide implementation examples of the most important algorithms to work with BNNs.

\section{What is a Bayesian Neural Network?}
\label{sec::definitions}

A BNN is defined slightly differently across the literature, but a commonly agreed definition is that a BNN is a stochastic artificial neural network trained using Bayesian inference.

\begin{figure}[t]
    \centering
    
    \begin{subfigure}[b]{0.15\textwidth}
    \includegraphics[width=\textwidth]{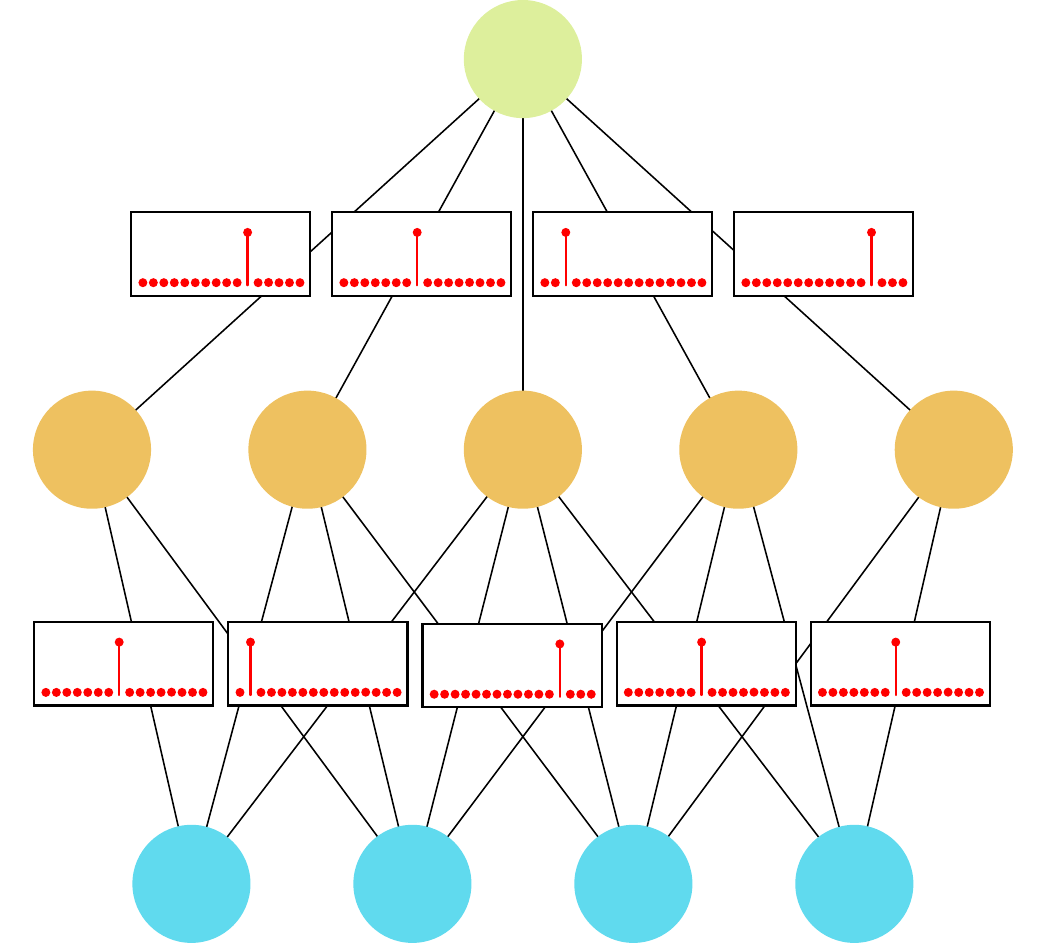}
    \caption{}
    \label{fig:nn:pe}
    \end{subfigure}
    \hfill
    \begin{subfigure}[b]{0.15\textwidth}
    \includegraphics[width=\textwidth]{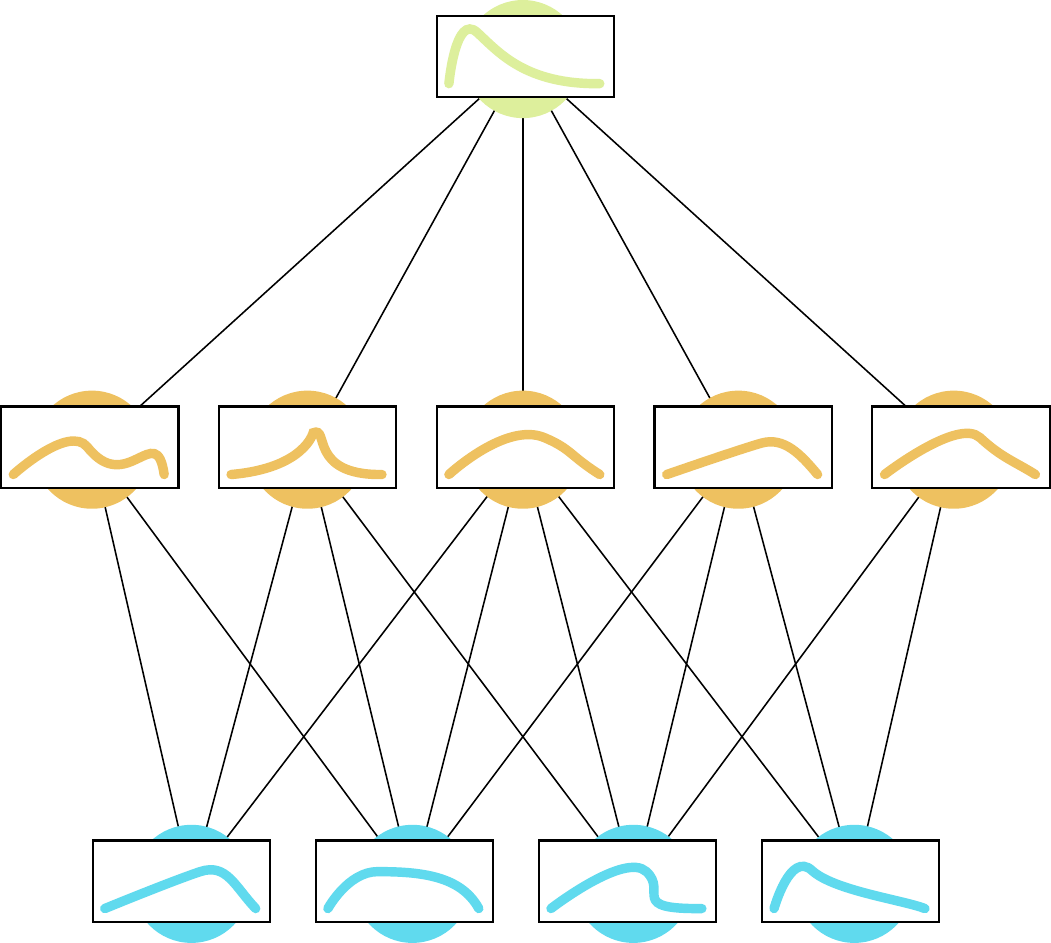}
    \caption{}
    \label{fig:nn:trans}
    \end{subfigure}
    \hfill
    \begin{subfigure}[b]{0.15\textwidth}
    \includegraphics[width=\textwidth]{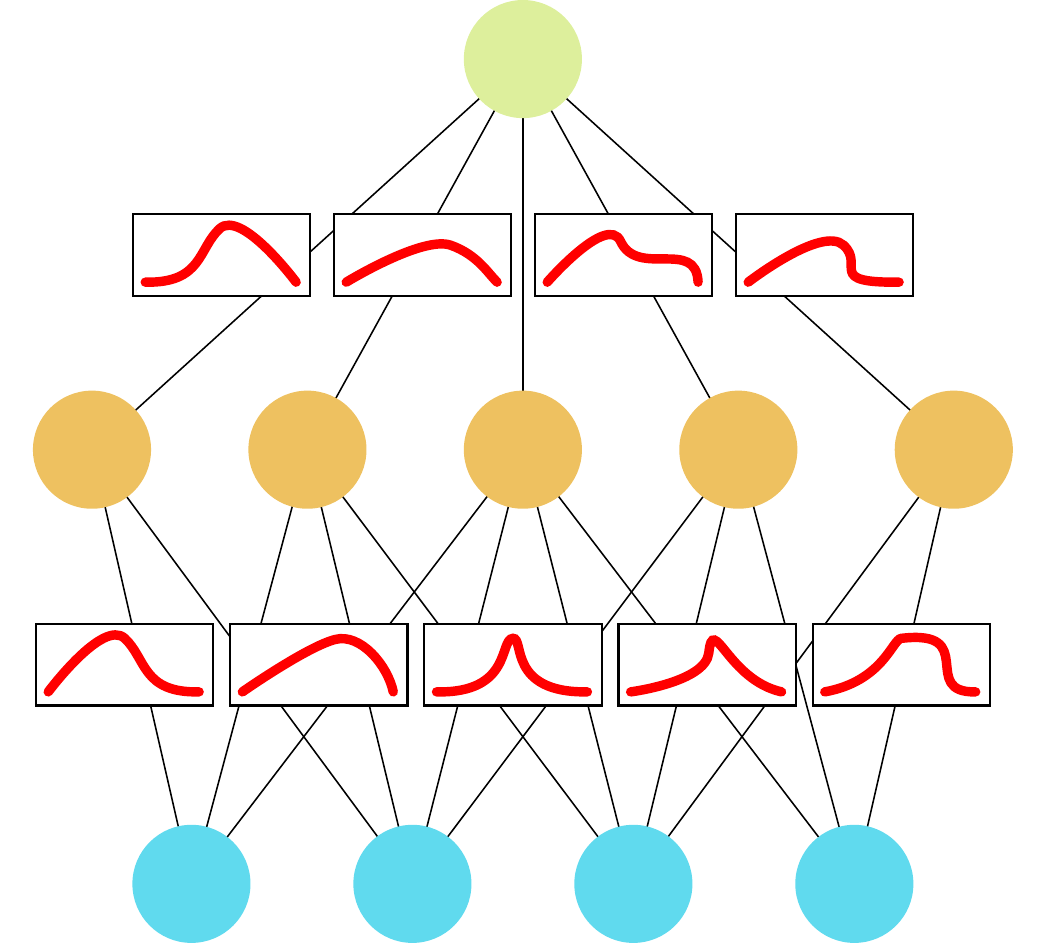}
    \caption{}
    \label{fig:nn:sto}
    \end{subfigure}
    
    \caption{(\subref{fig:nn:pe}) Point estimate neural network, (\subref{fig:nn:trans}) stochastic neural network with a probability distribution for the activations, and (\subref{fig:nn:sto}) stochastic neural network with a probability distribution over the weights.}
    \label{fig:nn}
\end{figure}

The goal of artificial neural networks (ANNs) is to represent an arbitrary function $\vec{y} = \NN(\vec{x})$. Traditional ANNs such as feedforward networks and recurrent networks are built using one input layer $ \vec{l}_0$, a succession of hidden layers $ \vec{l}_i, i=1,\dots, n-1$, and one output layer $ \vec{l}_n$. (Here, $n+1$ is the total number of layers.) In the simplest architecture of feedforward networks, each layer $\vec{l}$ is represented as a linear transformation, followed by a nonlinear operation $\nl$, also known as an \emph{activation function}:
\begin{equation}
    \label{eq:dnn}
    \begin{array}{l}
        \vec{l}_0  = \vec{x}, \\
        \vec{l}_i = \nl_i(\matr{W}_i \vec{l}_{i-1} ~+~ \vec{b}_i) \quad \forall i \in [1,n],\\
        \vec{y} = \vec{l}_n .
    \end{array}
\end{equation}
\noindent Here, $\vec{\theta} = (\matr{W}, \vec{b})$ are the parameters of the network, where  $\matr{W}$ are the weights of the network connections and $\vec{b}$ the biases. A given ANN architecture represents a set of functions isomorphic to the set of possible parameters $\vec{\theta}$. Deep learning is the process of regressing the parameters $\vec{\theta}$ from the training data $D$, where $D$ is composed of a series of input $\vec{x}$ and their corresponding labels $\vec{y}$. The standard approach is to approximate a minimal cost point estimate of the network parameters $\hat{\vec{\theta}}$, \ie a single value for each parameter (Figure~\ref{fig:nn:pe}), using the backpropagation algorithm, with all other possible parametrizations of the network discarded. The cost function is often defined as the log likelihood of the training set, sometimes with a regularization term included. From a statistician's point of view, this is a maximum likelihood estimation (MLE), or a maximum a posteriori (MAP) estimation when regularization is used.

The point estimate approach, which is the traditional approach in deep learning, is relatively easy to deploy with modern algorithms and software packages, but tends to lack explainability \cite{Yang2021}. The final model might also generalize in unforeseen and overconfident ways on out-of-training-distribution data points \cite{Guo2017,Nixon_2019_CVPR_Workshops}. This property, in addition to the inability of ANNs to say \textit{``I don't know"}, is 
problematic for many critical applications.
Of all the techniques that exist to mitigate this \cite{DBLP:conf/iclr/HendrycksG17}, stochastic neural networks have proven to be one of the most generic and flexible.

\textbf{Stochastic neural networks} are a type of ANN built by introducing stochastic components into the network. This is performed by giving the network either a stochastic activation (Figure~\ref{fig:nn:trans}) or stochastic weights (Figure~\ref{fig:nn:sto}) to simulate multiple possible models $\vec{\theta}$ with their associated probability distribution $p(\vec{\theta})$. Thus, BNNs can be considered a special case of \textbf{ensemble learning} \cite{Zhou:2012}.

The main motivation behind ensemble learning comes from the observation that aggregating the predictions of a large set of average-performing but independent predictors can lead to better predictions than a single well-performing expert predictor \cite{GALTON1907, Breiman1996}. Stochastic neural networks might improve their performance over their point estimate counterparts in a similar fashion, but this is not their main aim. Rather, the main goal of using a stochastic neural network architecture is to obtain a better idea of the uncertainty associated with the underlying processes. This is accomplished by comparing the predictions of multiple sampled model parametrizations $\vec{\theta}$. If the different models agree, then the uncertainty is low. If they disagree, then the uncertainty is high. This process can be summarized as follows:
\begin{equation}
\label{eq:samplingforward}
    \begin{array}{l}
         \vec{\theta} \sim p(\vec{\theta}), \\
         \vec{y} = \NN_{\vec{\theta}}(\vec{x}) ~+~ \vec{\epsilon} ,
    \end{array}
\end{equation}

\noindent where $\vec{\epsilon}$ represents random noise to account for the fact that the function $\NN$ is only an approximation.
A BNN can then be defined as any stochastic artificial neural network trained using Bayesian inference \cite{doi:10.1162/neco.1992.4.3.448}.

\emphBulletPoint{To design a BNN}, the first step is the choice of a deep neural network architecture, \ie \textbf{a functional model}. Then, one has to choose a \textbf{stochastic model}, \ie a prior distribution over the possible model parametrization $p(\vec{\theta})$ and a prior confidence in the predictive power of the model $p(\vec{y} | \vec{x}, \vec{\theta})$ (Figure~\ref{fig:workflow}a). The model parametrization can be considered to be the hypothesis $H$ and the training set is the data $D$. The choice of a BNN's stochastic model is somehow equivalent to the choice of a loss function when training a point estimate neural network; see Section~\ref{sec:reg-prior}. In the rest of this paper, we will denote the model parameters by $\vec{\theta}$, the training set by $D$, the training inputs by $D_{\vec{x}}$, and the training labels by $D_{\vec{y}}$. By applying Bayes' theorem, and enforcing independence between the model parameters and the input, the Bayesian posterior can be written as:
\begin{equation}\small
\label{eq:deeplearningbayesianposterior}
    p(\vec{\theta}|D) = \dfrac{p(D_{\vec{y}}|D_{\vec{x}},\vec{\theta})p(\vec{\theta})}{\int_{\vec{\theta}} p(D_{\vec{y}}|D_{\vec{x}},\vec{\theta'})p(\vec{\theta'}) d\vec{\theta'}} \propto p(D_{\vec{y}}|D_{\vec{x}},\vec{\theta})p(\vec{\theta}) .
\end{equation}

\noindent The Bayesian posterior for complex models such as artificial neural networks is a high dimensional and highly non-convex probability distribution \cite{izmailov2021bayesian}. This complexity makes computing and sampling it using standard methods an intractable problem, especially because computing the evidence $\int_{\vec{\theta}} p(D_{\vec{y}}|D_{\vec{x}},\vec{\theta'})p(\vec{\theta'}) d\vec{\theta'}$ is difficult. To address this problem, two broad approaches have been introduced: \textbf{(1)} Markov chain Monte Carlo and \textbf{(2)} variational inference. These are presented in more details in Section \ref{sec:inference}.

When \emphBulletPoint{using a BNN for prediction}, the probability distribution  $p(\vec{y}|\vec{x},D)$~\cite{WilsonDL2020}, called the marginal and which quantifies the model's uncertainty on its prediction, is of particular interest. Given $p(\vec{\theta}|D)$, $p(\vec{y}|\vec{x},D)$ can be computed as:
\begin{equation}
    p(\vec{y}|\vec{x},D) = \int_{\vec{\theta}} p(\vec{y}|\vec{x},\vec{\theta'})p(\vec{\theta'}|D) d\vec{\theta'} .
\end{equation}

\noindent In practice, $p(\vec{y}|\vec{x}, D)$ is sampled indirectly using Equation \eqref{eq:samplingforward}. The final prediction can be summarized by statistics computed using a Monte Carlo approach (Figure~\ref{fig:workflow}c). A large set of weights $\vec{\theta}_i$ is sampled from the posterior and used to compute a series of possible outputs $\vec{y}_i$, as shown in Algorithm~\ref{alg:BNN_inf}, which corresponds to samples from the marginal.

\begin{algorithm}[ht]
\begin{algorithmic}
{
    \STATE Define $p(\vec{\theta}|D) = \dfrac{p(D_{\vec{y}}|D_{\vec{x}},\vec{\theta})p(\vec{\theta})}{\int_{\vec{\theta}} p(D_{\vec{y}}|D_{\vec{x}},\vec{\theta'})p(\vec{\theta'}) d\vec{\theta'}}$;
    \FOR{$i=0$ \TO $N$}
        \STATE Draw $\vec{\theta}_i \sim p(\vec{\theta}|D)$;
        \STATE $\vec{y}_i = \NN_{\vec{\theta}_i}(\vec{x})$;
    \ENDFOR
    
    \RETURN $Y = \{\vec{y}_i | i \in [0,N)\},~\Theta = \{\vec{\theta}_i | i \in [0,N)\}$;
}
\end{algorithmic}
\caption{Inference procedure for a BNN.
\label{alg:BNN_inf}}
\end{algorithm}
\noindent In Algorithm~\ref{alg:BNN_inf}, $Y$ is a set of samples from $p(\vec{y}|\vec{x}, D)$ and $\Theta$ a collection of samples from $p(\vec{\theta}|D)$. Usually, aggregates are computed on those samples to summarize the uncertainty of the BNN and obtain an estimator for the output $\vec{y}$. This estimator is denoted by $\vec{\hat{y}}$.

When performing \textbf{regression}, the procedure that is usually used to summarize the predictions of a BNN is model averaging \cite{gal2015bayesian}:
\begin{equation}
\label{eq:deeplearningmodelaveraging}
         \vec{\hat{y}} = \dfrac{1}{|\Theta|} \sum_{\vec{\theta}_i \in \Theta} \NN_{\vec{\theta}_i}(\vec{x}).
\end{equation}

\noindent This approach is so common in ensemble learning that it is sometimes called ensembling. To quantify uncertainty, the covariance matrix can be computed as follows:
\begin{equation}
    \matr{\Sigma}_{\vec{y}|\vec{x}, D} = \dfrac{1}{|\Theta|-1} \sum_{\vec{\theta}_i \in \Theta}
    \left( \NN_{\vec{\theta}_i}(\vec{x}) - \vec{\hat{y}} \right) \left( \NN_{\vec{\theta}_i}(\vec{x}) - \vec{\hat{y}} \right)^\intercal.
\end{equation}

\noindent When performing \textbf{classification}, the average model prediction will give the relative probability of each class, which can be considered a measure of uncertainty:
\begin{equation}
\label{eq:deeplearningmodelaveragingcat}
         \vec{\hat{p}} = \dfrac{1}{|\Theta|} \sum_{\vec{\theta}_i \in \Theta} \NN_{\vec{\theta}_i}(\vec{x}) .
\end{equation}

\noindent The final prediction is taken as the most likely class:
\begin{equation}
\label{eq:classpredictor}
         \vec{\hat{y}} = \argmax_{i} p_i \in \vec{\hat{p}} .
\end{equation}

This definition considers BNNs as discriminative models, \ie models that aim to reconstruct a target variable $\vec{y}$ given observations $\vec{x}$. This excludes generative models, although there are examples of generative ANNs based on the Bayesian formalism, \eg Variational autoencoders \cite{kingma2014stochastic}. Those are out of the scope of this tutorial.

\section{Advantages of Bayesian methods for deep learning}
\label{sec::motivations}
One of the major critiques of Bayesian methods is that they rely on prior knowledge. This is especially true in deep learning, as deriving any insight about plausible parametrization for a given model before training is very challenging. Thus, why use Bayesian methods for deep learning? Discriminative models implicitly represent the conditional probability $p(\vec{y} | \vec{x}, \vec{\theta})$, and
Bayes' formula is an appropriate tool to invert conditional probabilities, even if one has little insight about $p(\vec{\theta})$ a priori. While there are strong theoretical principles and schema upon which
this Bayes' formula can be based \cite{robert2007bayesian}, this section focuses on some practical benefits of using BNNs.

\textbf{First}, Bayesian methods provide a natural approach to \textbf{quantify uncertainty} in deep learning since
BNNs have better calibration than classical neural networks \cite{MitrosDL2019,kristiadi2020bayesian, NIPS2019_9547}, \ie their uncertainty is more consistent with the observed errors. They are less often overconfident or underconfident.

\textbf{Second}, a BNN allows distinguishing between the \textbf{epistemic uncertainty} $p(\vec{\theta}|D)$ and the \textbf{aleatoric uncertainty} $p(\vec{y} | \vec{x}, \vec{\theta})$ \cite{KIUREGHIAN2009105}. This makes BNNs very data-efficient since they can learn from a small dataset without overfitting \cite{depeweg2017decomposition}. At prediction time, out-of-training distribution points will have high epistemic uncertainty instead of blindly giving a wrong prediction.

\textbf{Third}, the no-free-lunch theorem for machine learning \cite{nofreelunch} can be interpreted as stating that any supervised learning algorithm includes some implicit prior. Bayesian methods, when used correctly, will at least make the prior explicit. \textbf{Integrating prior knowledge} into ANNs, which work as black boxes, is difficult but not impossible. In Bayesian deep learning, priors are often considered as soft constraints, analogous to regularization, or data transformations such as data augmentation in traditional deep learning; see Section~\ref{sec:prior}. Most regularization methods used for point estimate neural networks can be understood from a Bayesian perspective as setting a prior; see Section~\ref{sec:reg-prior}.

\textbf{Finally}, the Bayesian paradigm enables \textbf{the analysis of learning methods}. A number of those methods initially not presented as Bayesian can be \textbf{implicitly understood} as being approximate Bayesian, \eg regularization (Section~\ref{sec:reg-prior}) or ensembling (Section~\ref{sec::approx::SGDD::VI}). In fact, most of the BNNs used in practice rely on methods that are approximately or implicitly Bayesian (Section~\ref{sec::approx}) since the exact algorithms are computationally too expensive. The Bayesian paradigm also provides a systematic framework to design new learning and regularization strategies, even for point estimate models.

BNNs have been used in many fields to quantify uncertainty, \eg in computer vision \cite{Kendall2017}, network traffic monitoring \cite{4049810}, aviation \cite{ZHANG2020113246}, civil engineering \cite{doi:10.1080/15732479.2014.951867, BATENI2007102}, hydrology \cite{hydrologyBNN}, astronomy \cite{Cobb_2019}, electronics \cite{Aminian2001}, and medicine \cite{Beker2020}. BNNs are useful in \textbf{(1)} active learning \cite{Gal2017, tran2019bayesian} where an oracle (\eg a human annotator, a crowd, an expensive algorithm) can label new points from an unlabeled dataset $U$. The model needs to determine which points should be submitted to the oracle to maximize its performance while minimizing the calls to the oracle. BNNs are also useful in \textbf{(2)} online learning \cite{opper1998bayesian}, where the model is retrained multiple times as new data become available. \emphBulletPoint{For active learning}, data points in the training set with high epistemic uncertainty are scheduled to be labeled with higher priority; see Algorithm~\ref{alg:BNN_active_learning}. In contrast, \emphBulletPoint{in online learning}, previous posteriors can be recycled as priors when new data become available to avoid the so-called problem of catastrophic forgetting \cite{10.5555/3327144.3327290}; see Algorithm~\ref{alg:BNN_online_learning}.

\begin{algorithm}[t]
\begin{algorithmic}
{
    \WHILE{$U \neq \varnothing$ and $\matr{\Sigma}_{\vec{y}|\vec{x}_{max}, D} < \text{threshold}$ and $C < \text{MaxC}$}
        \STATE Draw $\Theta = \{\vec{\theta}_i \sim p(\vec{\theta}|D)| i \in [0,N)\}$;
        \FOR{$\vec{x} \in U$}
            \STATE $\matr{\Sigma}_{\vec{y}|\vec{x}, D} = \dfrac{1}{|\Theta|-1} \sum_{\vec{\theta}_i \in \Theta}
    \left( \NN_{\vec{\theta}_i}(\vec{x}) - \vec{\hat{y}} \right) \left( \NN_{\vec{\theta}_i}(\vec{x}) - \vec{\hat{y}} \right)^\intercal$;
            \IF{$\matr{\Sigma}_{\vec{y}|\vec{x}, D} > \matr{\Sigma}_{\vec{y}|\vec{x}_{max}, D}$}
                \STATE $\vec{x}_{max} = \vec{x}$;
            \ENDIF
        \ENDFOR
        \STATE $D_{\vec{x}} = D_{\vec{x}} \cup \{\vec{x}_{\text{max}}\}$;
        \STATE $D_{\vec{y}} = D_{\vec{y}} \cup \{\text{Oracle}(\vec{x}_{\text{max}})\}$;
        \STATE $U = U \setminus \{\vec{x}_{\text{max}}\}$;
        \STATE $C = C+1$;
    \ENDWHILE
}
 \end{algorithmic}
\caption{Active learning loop with a BNN. \label{alg:BNN_active_learning}}
\end{algorithm}

\begin{algorithm}[t]
\begin{algorithmic}
{
    \STATE Define $p(\vec{\theta}) = p(\vec{\theta})_0$;
    \WHILE{\TRUE}
        \STATE Define $p(\vec{\theta}|D_i) = \dfrac{p(D_{\vec{y},i}|D_{\vec{x},i},\vec{\theta})p(\vec{\theta})_i}{\int_{\vec{\theta}}. p(D_{\vec{y},i}|D_{\vec{x},i},\vec{\theta'})p(\vec{\theta'})_i d\vec{\theta'}}$;
        \STATE Define $p(\vec{\theta})_{i+1} = p(\vec{\theta}|D_i)$;
    \ENDWHILE
}
\end{algorithmic}
\caption{Online learning loop with a BNN.
\label{alg:BNN_online_learning}}
\end{algorithm}

\section{Setting the stochastic model for a Bayesian Neural Network}
\label{sec::theory}
Designing a BNN requires choosing a \textbf{functional model} and a \textbf{stochastic model}. This tutorial will not cover the design of the functional model, \textbf{as almost any model used for point estimate networks can be used as a functional model for a BNN}. Furthermore, a rich literature on the subject exists already; see, for example, \cite{10.1145/3234150}. Instead, this section will focus on how to design the stochastic model. Section \ref{sec:pgm} introduces probabilistic graphical models (PGMs), a tool used to represent the relationships between the model's stochastic variables. Section \ref{sec:pgmimpl} details how to derive the posterior for a BNN from its PGM. Section \ref{sec:prior} discusses how to choose the probability laws used as priors. Finally, Section \ref{sec:degreesup} presents how the choice of a PGM can affect the degree of supervision or incorporate other forms of prior knowledge into the model.

\subsection{Probabilistic graphical models}
\label{sec:pgm}

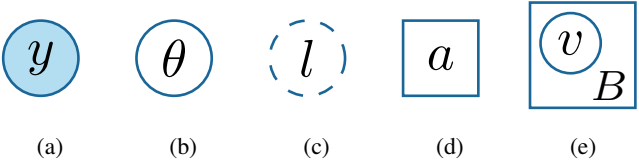
\begin{figure}[t]
    \centering
    
    \hfill
    \begin{subfigure}[b]{0.08\textwidth}
        \centering
        \scalebox{2}{\begin{tikzpicture}[scale=1,y=-1cm] 
    \fill[\pgmBgColor] (0,0) circle (0.25);
    \draw[\pgmLineColor,line width=0.5pt] (0,0) circle (0.25);
    \draw (0,0) node[anchor=center,align=center]{$y$};
    
    \draw (0,0.4);
\end{tikzpicture}}
        \caption{}
        \label{fig:pgmdef:obs}
    \end{subfigure}
    \hfill
    \begin{subfigure}[b]{0.08\textwidth}
        \centering
        \scalebox{2}{\begin{tikzpicture}[scale=1,y=-1cm] 
    \draw[\pgmLineColor,line width=0.5pt] (0,0) circle (0.25);
    \draw (0,0) node[anchor=center,align=center]{$\theta$};
    
    \draw (0,0.4);
\end{tikzpicture}}
        \caption{}
        \label{fig:pgmdef:nobs}
    \end{subfigure}
    \hfill
    \begin{subfigure}[b]{0.08\textwidth}
        \centering
        \scalebox{2}{\begin{tikzpicture}[scale=1,y=-1cm] 
    \draw[\pgmLineColor,line width=0.5pt, dashed] (0,0) circle (0.25);
    \draw (0,0) node[anchor=center,align=center]{$l$};
    
    \draw (0,0.4);
\end{tikzpicture}}
        \caption{}
        \label{fig:pgmdef:det}
    \end{subfigure}
    \hfill
    \begin{subfigure}[b]{0.08\textwidth}
        \centering
        \scalebox{2}{\begin{tikzpicture}[scale=1,y=-1cm] 
    \draw[\pgmLineColor,line width=0.5pt] (-0.25,-0.25) rectangle (0.25, 0.25);
    \draw (0,0) node[anchor=center,align=center]{$a$};
    
    \draw (0,0.4);
\end{tikzpicture}}
        \caption{}
        \label{fig:pgmdef:param}
    \end{subfigure}
    \hfill
    \begin{subfigure}[b]{0.08\textwidth}
        \centering
        \scalebox{2}{\begin{tikzpicture}[scale=1,y=-1cm] 
    \draw[\pgmLineColor,line width=0.5pt] (-0.35,-0.35) rectangle (0.35, 0.35);
    \draw[\pgmLineColor,line width=0.5pt] (-0.08,-0.08) circle (0.2);
    \draw (-0.08,-0.08) node[anchor=center,align=center]{$v$};
    \draw (0.42,0.42) node[anchor=south east,align=center]{\footnotesize $B$};
\end{tikzpicture}}
        \caption{}
        \label{fig:pgmdef:plate}
    \end{subfigure}
    \hfill
    
    \caption{The different symbols PGM, (\subref{fig:pgmdef:obs}) observed variables are in colored circles, (\subref{fig:pgmdef:nobs}) unobserved variables are in white circles, (\subref{fig:pgmdef:det}) deterministic variables are in dashed circles and (\subref{fig:pgmdef:param}) parameters are in rectangles. Plates, represented as a rectangle around a subgraph, indicate multiple independent instances of the subgraph for a batch of variables $B$ (\subref{fig:pgmdef:plate}).}
    \label{fig:pgmdef}
\end{figure}

Probabilistic graphical models (PGMs) use graphs to represent the interdependence of multivariate stochastic variables and subsequently decompose their probability distributions. PGMs cover a large variety of models. The type of PGMs this tutorial focuses on are Bayesian belief networks (BBN), which are PGMs whose graphs are acyclic and directed. 
We refer the reader to~\cite{Buntine_1994} for more details on how to represent learning algorithms using general PGMs.

In a PGM, variables $\vec{v}_i$ are the nodes in the graph. Different symbols are used to distinguish the nature of the considered variables (Figure~\ref{fig:pgmdef}). A directed link, which is the only type of link allowed in a BBN, means that the probability distribution of the target variable is defined conditioned on the source variable.
The fact that the BBN is acyclic allows the computation of the joint probability distribution of all the variables $\vec{v}_i$ in the graph:
\begin{equation}
    \label{eq:pgmdecomposition}
    p(\vec{v}_1, ... , \vec{v}_n) = \prod_{i=1}^{n} p(\vec{v}_i | \text{parents}(\vec{v}_i)).
\end{equation}

\noindent The type of distribution used to define the conditional probabilities $p(\vec{v}_i | \text{parents}(\vec{v}_i))$ depends on the context. Once the conditional probabilities are defined, the BBN describes a data generation process. Parents are sampled before their children. This is always possible since the graph is acyclic. All the variables together represent a sample from the joint probability distribution $p(\vec{v}_1,\dots, \vec{v}_n)$.

Models usually learn from multiple examples sampled from the same distribution. To highlight this fact, the plate notation (Figure~\ref{fig:pgmdef:plate}) has been introduced. A plate indicates that the variables $(\vec{v}_1, ... , \vec{v}_n)$ in the subgraph encapsulated by the plate are copied along a given batch dimension. A plate implies independence between all the duplicated nodes. This fact can be exploited to compute the joint probability of a batch $B = \left\{(\vec{v}_1, ... , \vec{v}_n)_b: b = 1, \dots, |B|\right\}$ as:
\begin{equation}
    \label{eq:pgmplatebatch}
    p(B) = \prod_{(\vec{v}_1, \dots, \vec{v}_n) \in B} p(\vec{v}_1, \dots , \vec{v}_n).
\end{equation}

\noindent In a PGM, the observed variables, depicted in Figure~\ref{fig:pgmdef:obs} using colored circles, are treated as the data. The unobserved, also called latent variables, represented by a white circle in Figure~\ref{fig:pgmdef:nobs}, are treated as the hypothesis. From the joint probability derived from the PGM, defining the posterior for the latent variables given the observed variables is straightforward using Bayes' formula:
\begin{equation}
    \label{eq:bayesfrompgm}
    p(\vec{v}_{\text{latent}}|\vec{v}_{\text{obs}}) \propto p(\vec{v}_{\text{obs}},\vec{v}_{\text{latent}}).
\end{equation}

\noindent The joint distribution $p(\vec{v}_{\text{obs}},\vec{v}_{\text{latent}})$ is then used by the different inference algorithms; see Section~\ref{sec:inference}.

\subsection{Defining the stochastic model of a BNN from a PGM}
\label{sec:pgmimpl}

\begin{figure}[t]
    \centering
    
    \includegraphics[width=0.48\textwidth]{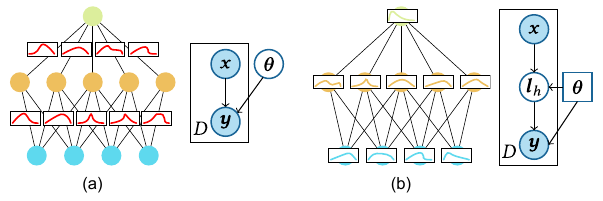}
    
    \caption{BBNs with (a) coefficients as stochastic variables and (b) activations as stochastic variables. }
    \label{fig:pgmbnn}
\end{figure}

Consider the two models presented in Figure~\ref{fig:pgmbnn}, with both the BNN and the corresponding BBN depicted. The BNN with stochastic weights (Figure~\ref{fig:pgmbnn}a), if meant to perform regression, could represent the following data generation process:
\begin{equation}
\begin{array}{l}
    \vec{\theta} \sim p(\vec{\theta}) = \mathcal{N}(\vec{\mu}, \matr{\Sigma}),\\
    \vec{y} \sim p(\vec{y}|\vec{x}, \vec{\theta}) = \mathcal{N}(\NN_{\vec{\theta}}(\vec{x}), \matr{\Sigma}) .
\end{array}
\end{equation}

\noindent The choice of using normal laws $\mathcal{N}(\vec{\mu}, \matr{\Sigma})$, with mean $\vec{\mu}$ and covariance $\matr{\Sigma})$, is arbitrary but is common in practice because of its good mathematical properties.

For classification, the model samples the prediction from a categorical law $\text{Cat}(p_i)$, \ie
\begin{equation}
\begin{array}{l}
    \vec{\theta} \sim p(\vec{\theta}) = \mathcal{N}(\vec{\mu}, \matr{\Sigma}),\\
    \vec{y} \sim p(\vec{y}|\vec{x}, \vec{\theta}) = \text{Cat}(\NN_{\vec{\theta}}(\vec{x})).
\end{array}
\end{equation}

\noindent Then, one can use the fact that multiple data points from the training set are independent, as indicated by the plate notation in Figure~\ref{fig:pgmbnn}, to write the probability of the training set as:
\begin{equation}
    p(D_{\vec{y}}|D_{\vec{x}}, \vec{\theta}) = \prod_{(\vec{x},\vec{y}) \in D} p(\vec{y}|\vec{x}, \vec{\theta}) .
\end{equation}

\noindent In the case of stochastic activations (Figure~\ref{fig:pgmbnn}b), the data generation process might become:
\begin{equation}
    \begin{array}{l}
        \vec{l}_0  = \vec{x}, \\
        \vec{l}_i \sim p(\vec{l}_i|\vec{l}_{i-1}) = \nl_i(\mathcal{N}(\matr{W}_i \vec{l}_{i-1} ~+~ \vec{b}_i, \Sigma)) \quad \forall i \in [1,n],\\
        \vec{y} = \vec{l}_n .
    \end{array}
\end{equation}

\noindent The formulation of the joint probability is slightly more complex as we have to account for the chain of dependencies spanned by the BBN over the multiple latent variables $\vec{l}_{[1,n-1]}$:
\begin{equation}
    p(D_{\vec{y}}, \vec{l}_{[1,n-1]}|D_{\vec{x}}) = \prod_{(\vec{l}_0,\vec{l}_n) \in D} \left( \prod_{i = 1}^n p(\vec{l}_i|\vec{l}_{i-1}) \right).
\end{equation}

\noindent It is sometimes possible, and often desirable, to define $p(\vec{l}_i|\vec{l}_{i-1})$ such that the BNNs described in Figure~\ref{fig:pgmbnn}a and in Figure~\ref{fig:pgmbnn}b can be considered equivalent. For instance, sampling $l$ as:
\begin{equation}
    \begin{array}{l}
        \matr{W}  \sim \mathcal{N}(\vec{\mu}_{\matr{W}}, \matr{\Sigma}_{\matr{W}}), \\
        \vec{b} \sim \mathcal{N}(\vec{\mu}_{\vec{b}}, \matr{\Sigma}_{\vec{b}}), \\
        \vec{l} = \nl(\matr{W} \vec{l}_{-1} + \vec{b})
    \end{array}
\end{equation}

\noindent is equivalent to sampling $l$ as: 
\begin{equation}
    \vec{l} \sim \nl( \mathcal{N}( \vec{\mu}_{\matr{W}} \vec{l}_{-1} + \vec{\mu}_{\vec{b}}, (\matr{I} \otimes \vec{l}_{-1})^\intercal \matr{\Sigma}_{\matr{W}} (\matr{I} \otimes \vec{l}_{-1}) + \matr{\Sigma}_{\vec{b}}) ) ,
\end{equation}

\noindent where $\otimes$ denotes a Kronecker product.

The basic Bayesian regression architecture shown in Figure~\ref{fig:pgmbnn}a is more common in practice. The alternative architecture shown in  Figure~\ref{fig:pgmbnn}b is sometimes used as it allows compressing the number of variational parameters when using variational inference~\cite{wen2018flipout}; see also Section~\ref{sec:inference}.

\subsection{Setting the priors}
\label{sec:prior}

Setting the prior of a deep neural network is often not an intuitive task. The main problem is that it is not truly explicit how models with a very large number of parameters and a nontrivial architecture such as an ANN will generalize for a given parametrization \cite{ZhangBHRV17}. 
In this Section, we first present the common practice, discuss the issues related to the statistical unidentifiability of ANNs, and then show the link between the prior for BNNs and regularization for the point estimate algorithms. Finally, we present a method to build the prior from high level knowledge.

\subsubsection{A good default prior}
\label{sec:priordefault}

For basic architectures such as Bayesian regression (Figure~\ref{fig:pgmbnn}a), a standard procedure is to use a normal prior with a zero mean $\vec{0}$ and a diagonal covariance  $\sigma \matr{I}$ on the coefficients of the network:
\begin{equation}
\label{eq:defaultprior}
    p(\vec{\theta}) = \mathcal{N}(\vec{0},\sigma \matr{I}) .
\end{equation}
\noindent This approach 
is equivalent to a weighted $\ell_2$ regularization (with weights $1/\sigma$) when training a point estimate network, as will be demonstrated in Section~\ref{sec:reg-prior}. The documentation of the probabilistic programming language Stan~\cite{carpenter2017stan} provides examples on how to choose $\sigma$ knowing the expected scale of the considered parameters~\cite{StanPriors}.

Although such an approach is often used in practice, there is no theoretical argument that makes it better than any other formulation \cite{silvestro2020prior}. The normal law is preferred due to its mathematical properties and the simple formulation of its log, which is used in most of the learning algorithms. 

\subsubsection{Addressing unidentifiability in Bayesian neural networks}
\label{sec:prioridenf}

One of the main problems with Bayesian deep learning is that deep neural networks are overparametrized models, \ie they have many equivalent parametrizations \cite{10.5555/2380985}. This is an example of statistical unidentifiability, which can lead to complex multimodal posteriors that are hard to sample and approximate when training a BNN \cite{izmailov2021bayesian}. There are two solutions to deal with this issue: \textbf{(1)} changing the functional model parametrization, or \textbf{(2)} constraining the support of the prior to remove unidentifiability.

The two most common classes of nonuniqueness in ANNs are weight-space symmetry and scaling symmetry \cite{pourzanjani2017bayesian}. Both are not a concern for point estimate neural networks but might be for BNNs. Weight-space symmetry implies that one can build an equivalent parametrization of an ANN with at least one hidden layer. This is achieved by permuting two rows in $(\matr{W}_i, \vec{b}_i)$, the weights and their corresponding bias $\vec{b}_i$, of one of the hidden layers as well as the corresponding columns in the following layer's weight matrix $\matr{W}_{i+1}$. This means that as the number of hidden layers and the number of units in the hidden layers grow, the number of equivalent representations, which would roughly correspond to the modes in the posterior distribution, grows factorially. A mitigation strategy is to enforce the bias vector in each layer to be sorted in an ascending or a descending order. However, the practical effects of doing so may be to degrade optimization: weight-space symmetry may implicitly support the exploration of the parameter space during the early stages of the optimization.

Scaling symmetry is an unidentifiability problem arising when using nonlinearities with the property $\nl(\alpha x)=\alpha \nl(x)$, which is the case of RELU and Leaky-RELU, two popular nonlinearities in modern machine learning. In this case, assigning the weights $\matr{W}_l, \matr{W}_{l+1}$ to two consecutive layers $l$ and $l+1$ becomes strictly equivalent to assigning $\alpha \matr{W}_l, (1/\alpha) \matr{W}_{l+1}$.
This can reduce the convergence speed for point estimate neural networks, a problem that is addressed in practice with various activation normalization techniques \cite{ba2016layer}. BNNs are slightly more complex as the scaling symmetry influences the posterior shape, making it harder to approximate. Givens transformations (also called Givens rotations) have been proposed as a mean to constrain the norm of the hidden layers \cite{pourzanjani2017bayesian} and address the scaling symmetry issue. In practice, using a Gaussian prior already reduces the scaling symmetry problem, as it favors weights with the same Frobenius norm on each layer. A soft version of the activation normalization can also be implemented by using a consistency condition; see Section~\ref{sec:consistentprior}. The additional complexity associated with sampling the network parameters in a constrained space to perfectly remove the scaling symmetry is computationally prohibitive. We provide, in the Practical Example III of the Supplementary Material , additional discussion on this issue using the "Paperfold" practical example.

\begin{figure*}[t]
    \centering
    
    \begin{subfigure}[b]{0.15\textwidth}
        \centering
        \scalebox{1.}{\begin{tikzpicture}[scale=1,y=1cm] 
    
    \draw[\pgmLineColor,line width=0.5pt] (0.75,1.75) circle (0.25);
    \draw (0.75,1.75) node[anchor=center,align=center]{$\vec{\theta}$};

    \draw[line width=0.5pt,->] (0.65, 1.52) -- (0.2, 0.9);

    \draw[\pgmLineColor,line width=0.5pt] (-0.6,-0.6) rectangle (0.4, 2.1);
    \draw (-0.68,-0.6) node[anchor=south west,align=center]{\footnotesize $D$};
    
    \fill[\pgmBgColor] (0,1.75) circle (0.25);
    \draw[line width=0.75pt] (0,1.75) circle (0.25);
    \draw (0,1.75) node[anchor=center,align=center]{$\vec{x}$};

    \draw[line width=0.5pt,->] (0, 1.5) -- (0, 1);

    \draw[\pgmLineColor,line width=0.75pt] (0,0.75) circle (0.25);
    \draw (0,0.75) node[anchor=center,align=center]{$\vec{y}$};

    \draw[line width=0.5pt,->] (0, 0.5) -- (0, 0);

    \fill[\pgmBgColor] (0,-0.25) circle (0.25);
    \draw[line width=0.75pt] (0,-0.25) circle (0.25);
    \draw (0,-0.25) node[anchor=center,align=center]{$\vec{\tilde{y}}$};

\end{tikzpicture}}
        \caption{Noisy labels}
        \label{fig:pgmnlssl:noisy}
    \end{subfigure}
    \begin{subfigure}[b]{0.22\textwidth}
        \centering
        \scalebox{1.}{\begin{tikzpicture}[scale=1,y=1cm] 
    
    \draw[\pgmLineColor,line width=0.5pt] (0.75,1.75) circle (0.25);
    \draw (0.75,1.75) node[anchor=center,align=center]{$\vec{\theta}$};

    \draw[line width=0.5pt,->] (0.65, 1.52) -- (0.2, 0.9);
    
    \fill[\pgmBgColor] (0,1.75) circle (0.25);
    \draw[\pgmLineColor,line width=0.75pt] (0,1.75) circle (0.25);
    \draw (0,1.75) node[anchor=center,align=center]{$\vec{x}$};

    \draw[line width=0.5pt,->] (0, 1.5) -- (0, 1);

    \fill[\pgmBgColor] (0,0.75) circle (0.25);
    \fill[white] (-0.25,1) rectangle (0.0,0.5);
    \draw[\pgmLineColor,line width=0.75pt] (0,0.75) circle (0.25);
    \draw (0,0.75) node[anchor=center,align=center]{$\vec{y}$};
    
    \draw (0,0);
    
\end{tikzpicture}}
        \caption{Semi-supervised learning}
        \label{fig:pgmnlssl:semis}
    \end{subfigure}
    \begin{subfigure}[b]{0.2\textwidth}
        \centering
        \scalebox{1.}{\begin{tikzpicture}[scale=1,y=1cm] 

    \fill[\pgmBgColor] (0,0.75) circle (0.25);
    \draw[\pgmLineColor,line width=0.75pt] (0,0.75) circle (0.25);
    \draw (0,0.75) node[anchor=center,align=center]{$\vec{y}$};
    
    \draw[\pgmLineColor,line width=0.5pt] (0.75,1.75) circle (0.25);
    \draw (0.75,1.75) node[anchor=center,align=center]{$\vec{\theta}$};

    \draw[line width=0.5pt,->] (0.65, 1.52) -- (0.2, 0.9);
    
    \fill[\pgmBgColor] (0,2.75) circle (0.25);
    \draw[\pgmLineColor,line width=0.75pt] (0,2.75) circle (0.25);
    \draw (0,2.75) node[anchor=center,align=center]{$\vec{x}$};
    

    \draw[line width=0.5pt,->] (0, 2.5) -- (0, 2.0);
    
    \fill[white] (0,1.75) circle (0.25);
    \draw[\pgmLineColor,line width=0.75pt] (0,1.75) circle (0.25);
    \draw (0,1.75) node[anchor=center,align=center]{$\vec{x'}$};

    \draw[line width=0.5pt,->] (0, 1.5) -- (0, 1);
    

    \draw[\pgmLineColor,line width=0.5pt] (-0.65,3.1) rectangle (0.4, 0.4);
    \draw (-0.7,0.4) node[anchor=south west,align=center]{\footnotesize $D$};

    
\end{tikzpicture}}
        \caption{Data augmentation}
        \label{fig:pgmdataaugm}
    \end{subfigure}
    \begin{subfigure}[b]{0.18\textwidth}
        \centering
        \scalebox{1.}{\begin{tikzpicture}[scale=1,y=1cm]

    \draw[\pgmLineColor,line width=0.5pt] (0.75,2.7) circle (0.25);
    \draw (0.75,2.7) node[anchor=center,align=center]{$\vec{\xi}$};

    \draw[line width=0.5pt,->] (0.75, 2.45) -- (0.75, 2);
    
    \draw[\pgmLineColor,line width=0.5pt] (-0.75,0.2) rectangle (1.1, 2.25);
    \draw (0.85,0.2) node[anchor=south,align=center]{$T$};
    
    \draw[\pgmLineColor,line width=0.5pt] (0.75,1.75) circle (0.25);
    \draw (0.75,1.75) node[anchor=center,align=center]{$\vec{\theta}$};

    \draw[line width=0.5pt,->] (0.65, 1.52) -- (0.2, 0.9);
    
    \draw[\pgmLineColor,line width=0.5pt] (-0.6,0.4) rectangle (0.4, 2.1);
    \draw (-0.68,0.4) node[anchor=south west,align=center]{\footnotesize $D$};
    
    \fill[\pgmBgColor] (0,1.75) circle (0.25);
    \draw[\pgmLineColor,line width=0.75pt] (0,1.75) circle (0.25);
    \draw (0,1.75) node[anchor=center,align=center]{$\vec{x}$};

    \draw[line width=0.5pt,->] (0, 1.5) -- (0, 1);

    \fill[\pgmBgColor] (0,0.75) circle (0.25);
    \draw[\pgmLineColor,line width=0.75pt] (0,0.75) circle (0.25);
    \draw (0,0.75) node[anchor=center,align=center]{$\vec{y}$};
    
\end{tikzpicture}}
        \caption{Meta-learning}
        \label{fig:pgmtlssl:ml}
    \end{subfigure}
    \begin{subfigure}[b]{0.2\textwidth}
        \centering
        \scalebox{1.}{\begin{tikzpicture}[scale=1,y=1cm] 
    
    \draw[\pgmLineColor,line width=0.5pt] (0.75,1.75) circle (0.25);
    \draw (0.75,1.75) node[anchor=center,align=center]{$\vec{\theta_s}$};

    \draw[line width=0.5pt,->] (0.65, 1.52) -- (0.2, 0.9);
    
    \draw[\pgmLineColor,line width=0.5pt] (0.75,0.75) circle (0.25);
    \draw (0.75,0.75) node[anchor=center,align=center]{$\vec{\theta_t}$};

    \draw[line width=0.5pt,->] (0.65, 0.52) -- (0.2, -0.1);
    
    \draw[\pgmLineColor,line width=0.5pt] (-1.5,0.75) circle (0.25);
    \draw (-1.5,0.75) node[anchor=center,align=center]{$\vec{\theta_p}$};

    \draw[line width=0.5pt,->] (-1.4, 0.52) -- (-0.95, -0.1);

    \draw[\pgmLineColor,line width=0.5pt] (-1.1,-0.6) rectangle (0.4, 2.1);
    \draw (-1.13,2.1) node[anchor=north west,align=center]{\footnotesize $D$};
    
    \fill[\pgmBgColor] (0,1.75) circle (0.25);
    \draw[\pgmLineColor,line width=0.75pt] (0,1.75) circle (0.25);
    \draw (0,1.75) node[anchor=center,align=center]{$\vec{x}$};

    \draw[line width=0.5pt,->] (0, 1.5) -- (0, 1);

    \draw[\pgmLineColor,line width=0.75pt, dashed] (0,0.75) circle (0.25);
    \draw (0,0.75) node[anchor=center,align=center]{$\vec{l}$};

    \draw[line width=0.5pt,->] (-0.2, 0.57) -- (-0.66, -0.02);
    \draw[line width=0.5pt,->] (0.0, 0.5) -- (0, 0);

    \fill[\pgmBgColor] (-0.75,-0.25) circle (0.25);
    \draw[\pgmLineColor,line width=0.75pt] (-0.75,-0.25) circle (0.25);
    \draw (-0.75,-0.25) node[anchor=center,align=center]{$\vec{y_p}$};

    \fill[\pgmBgColor] (0,-0.25) circle (0.25);
    \fill[white] (-0.25,-0.5) rectangle (0.0,0.0);
    \draw[\pgmLineColor,line width=0.75pt] (0,-0.25) circle (0.25);
    \draw (0,-0.25) node[anchor=center,align=center]{$\vec{y_t}$};

\end{tikzpicture}}
        \caption{Self-supervised learning}
        \label{fig:pgmtlssl:ssl}
    \end{subfigure}

    \caption{Different examples of PGMs to adapt the learning strategy for a given BNN (with stochastic weights).}
    \label{fig:pgm_models}
\end{figure*}
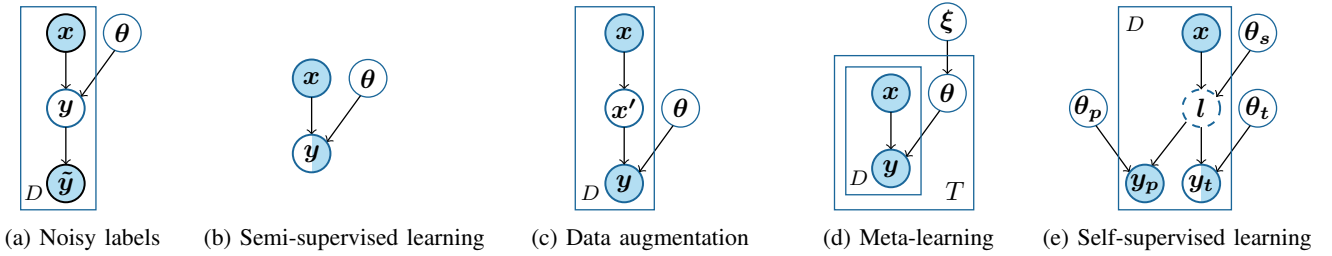

\subsubsection{The link between regularization and priors}
\label{sec:reg-prior}

The usual learning procedure for a point estimate neural network is to find the set of parameters $\vec{\theta}$ that minimize a loss function built using the training set $D$:
\begin{equation}
    \vec{\hat{\theta}} = \argmin_{\vec{\theta}} ~ \text{loss}_{D_x,D_y}(\vec{\theta}) .
\end{equation}

\noindent Assuming that the loss is defined as minus the log-likelihood function up to an additive constant, the problem can be rewritten as:
\begin{equation}
\label{eq:probloss}
    \vec{\hat{\theta}} = \argmax_{\vec{\theta}} ~ p(D_{\vec{y}}|D_{\vec{x}},\vec{\theta}) ,
\end{equation}
\noindent which would be the first half of the model according to the Bayesian paradigm. Now, assume that we also have a prior for $\vec{\theta}$, and we want to find the most likely point estimate from the posterior. The problem can be reformulated as:
\begin{equation}
    \vec{\hat{\theta}} = 
    \argmax_{\vec{\theta}} ~ p(D_{\vec{y}}|D_{\vec{x}},\vec{\theta})p(\vec{\theta}) .
\end{equation}
\noindent 
Next, one would go back to a log-likelihood formulation:
\begin{equation}
    \label{eq:regularizedopt}
    \vec{\hat{\theta}} = \argmin_{\vec{\theta}} ~ \text{loss}_{D_x,D_y}(\vec{\theta}) + \text{reg}(\vec{\theta}), 
\end{equation}

\noindent  which is easier to optimize. Equation \eqref{eq:regularizedopt} is how regularization is usually applied in machine learning and in many other fields. Another argument, less formal, is that regularization acts as a soft constraint on the search space, in a manner similar to what a prior does for a posterior.

\subsubsection{Prior with a consistency condition}
\label{sec:consistentprior}
Regularization can also be implemented with a consistency condition $C(\vec{\theta}, \vec{x})$, which is a function used to measure how well the model respects some hypothesis given a parametrization $\vec{\theta}$ and an input $\vec{x}$. For example, $C$ can be set to favor sparse or regular predictions to encourage monotonicity of predictions with respect to some input variables (\eg the probability of getting the flu increases with age), or to favor decision boundaries in low density regions when using semi-supervised learning; see Section~\ref{sec:degreesup:nlssl}. $C$ can be seen as the relative log likelihood of a prediction given the input $\vec{x}$ and parameter set $\vec{\theta}$. Thus, it can be included in the prior. To this end, $C$ should be averaged over all possible inputs:
\begin{equation}
    C(\vec{\theta}) = \int_{\vec{x}}  C(\vec{\theta}, \vec{x}) p(\vec{x}) d\vec{x} .
\end{equation}

\noindent In practice, as $p(\vec{x})$ is unknown, $C(\vec{\theta})$ is approximated from the features in the training set:
\begin{equation}
    C(\vec{\theta}) \approx \dfrac{1}{|D_{\vec{x}}|} \sum_{\vec{x} \in |D_{\vec{x}}|} C(\vec{\theta}, \vec{x}) .
\end{equation}

\noindent We can now write a function proportional to the prior with the consistency condition included:
\begin{equation}
    \label{eq:consistentprior}
    p(\vec{\theta} | D_{\vec{x}}) \propto p(\vec{\theta}) \operatorname{exp}\left( -\dfrac{1}{|D_{\vec{x}}|} \sum_{\vec{x} \in |D_{\vec{x}}|} C(\vec{\theta}, \vec{x}) \right) ,
\end{equation}

\noindent where $p(\vec{\theta})$ is the prior without the consistency condition.

\subsection{Degree of supervision and alternative forms of prior knowledge}
\label{sec:degreesup}

The architecture presented in Section~\ref{sec:pgmimpl} focuses mainly on the use of BNNs in a supervised learning setting. However, in real world applications, obtaining ground-truth labels can be expensive. Thus, new learning strategies should be adopted \cite{qi2019small}. We will now present how to adapt BNNs for different degrees of supervision. While doing so, we will also demonstrate how PGMs in general and BBNs in particular are useful in designing or interpreting learning strategies. In particular, the formulation of the Bayesian posterior, which is derived from the different PGMs presented in Figure~\ref{fig:pgm_models}, can also be used for a point estimate neural network to obtain a suitable loss function to search for an MAP estimator for the parameters (Section~\ref{sec:reg-prior}). We also provide a practical example in the Supplementary Material (Practical Example II) to illustrate how such strategies can be implemented for an actual BNN.

\subsubsection{Noisy labels and semi-supervised learning}
\label{sec:degreesup:nlssl}

\begin{figure}[b]
    \centering
    
    \begin{subfigure}[b]{0.15\textwidth}
        \centering
        \scalebox{1.}{\begin{tikzpicture}[scale=1,y=1cm] 
    
    \draw[\pgmLineColor,line width=0.5pt] (0.75,1.75) circle (0.25);
    \draw (0.75,1.75) node[anchor=center,align=center]{$\vec{\theta}$};

    \draw[line width=0.5pt,->] (0.65, 1.52) -- (0.2, 0.9);

    \draw[\pgmLineColor,line width=0.5pt] (-1.1,-0.6) rectangle (0.4, 2.1);
    \draw (-1.18,-0.6) node[anchor=south west,align=center]{\footnotesize $D$};
    
    \fill[\pgmBgColor] (0,1.75) circle (0.25);
    \draw[line width=0.75pt] (0,1.75) circle (0.25);
    \draw (0,1.75) node[anchor=center,align=center]{$\vec{x}$};

    \draw[line width=0.5pt,->] (0, 1.5) -- (0, 1);

    \draw[\pgmLineColor,line width=0.75pt] (0,0.75) circle (0.25);
    \draw (0,0.75) node[anchor=center,align=center]{$\vec{y}$};
    
    \draw[line width=0.5pt,->] (-0.65, 0.52) -- (-0.2, -0.1);

    \draw[\pgmLineColor,line width=0.75pt] (-0.75,0.75) circle (0.25);
    \draw (-0.75,0.75) node[anchor=center,align=center]{$\vec{\sigma}$};

    \draw[line width=0.5pt,->] (0, 0.5) -- (0, 0);

    \fill[\pgmBgColor] (0,-0.25) circle (0.25);
    \draw[line width=0.75pt] (0,-0.25) circle (0.25);
    \draw (0,-0.25) node[anchor=center,align=center]{$\vec{\tilde{y}}$};

\end{tikzpicture}}
        \caption{}
        \label{fig:pgmnlclass:NCAR}
    \end{subfigure}
    \hfill
    \begin{subfigure}[b]{0.15\textwidth}
        \centering
        \scalebox{1.}{\begin{tikzpicture}[scale=1,y=1cm] 
    
    \draw[\pgmLineColor,line width=0.5pt] (0.75,1.75) circle (0.25);
    \draw (0.75,1.75) node[anchor=center,align=center]{$\vec{\theta}$};

    \draw[line width=0.5pt,->] (0.65, 1.52) -- (0.2, 0.9);

    \draw[\pgmLineColor,line width=0.5pt] (-1.1,-0.6) rectangle (0.4, 2.1);
    \draw (-1.18,-0.6) node[anchor=south west,align=center]{\footnotesize $D$};
    
    \fill[\pgmBgColor] (0,1.75) circle (0.25);
    \draw[line width=0.75pt] (0,1.75) circle (0.25);
    \draw (0,1.75) node[anchor=center,align=center]{$\vec{x}$};

    \draw[line width=0.5pt,->] (0, 1.5) -- (0, 1);

    \draw[\pgmLineColor,line width=0.75pt] (0,0.75) circle (0.25);
    \draw (0,0.75) node[anchor=center,align=center]{$\vec{y}$};
    
    \draw[line width=0.5pt,->] (-0.25, 0.75) -- (-0.5, 0.75);
    \draw[line width=0.5pt,->] (-0.65, 0.52) -- (-0.2, -0.1);

    \draw[\pgmLineColor,line width=0.75pt] (-0.75,0.75) circle (0.25);
    \draw (-0.75,0.75) node[anchor=center,align=center]{$\vec{\sigma}$};

    \draw[line width=0.5pt,->] (0, 0.5) -- (0, 0);

    \fill[\pgmBgColor] (0,-0.25) circle (0.25);
    \draw[line width=0.75pt] (0,-0.25) circle (0.25);
    \draw (0,-0.25) node[anchor=center,align=center]{$\vec{\tilde{y}}$};

\end{tikzpicture}}
        \caption{}
        \label{fig:pgmnlclass:NAR}
    \end{subfigure}
    \hfill
    \begin{subfigure}[b]{0.15\textwidth}
        \centering
        \scalebox{1.}{\begin{tikzpicture}[scale=1,y=1cm] 
    
    \draw[\pgmLineColor,line width=0.5pt] (0.75,1.75) circle (0.25);
    \draw (0.75,1.75) node[anchor=center,align=center]{$\vec{\theta}$};

    \draw[line width=0.5pt,->] (0.65, 1.52) -- (0.2, 0.9);

    \draw[\pgmLineColor,line width=0.5pt] (-1.1,-0.6) rectangle (0.4, 2.1);
    \draw (-1.18,-0.6) node[anchor=south west,align=center]{\footnotesize $D$};
    
    \fill[\pgmBgColor] (0,1.75) circle (0.25);
    \draw[line width=0.75pt] (0,1.75) circle (0.25);
    \draw (0,1.75) node[anchor=center,align=center]{$\vec{x}$};

    \draw[line width=0.5pt,->] (-0.15, 1.55) -- (-0.6, 0.95);
    \draw[line width=0.5pt,->] (0, 1.5) -- (0, 1);

    \draw[\pgmLineColor,line width=0.75pt] (0,0.75) circle (0.25);
    \draw (0,0.75) node[anchor=center,align=center]{$\vec{y}$};
    
    \draw[line width=0.5pt,->] (-0.25, 0.75) -- (-0.5, 0.75);
    \draw[line width=0.5pt,->] (-0.65, 0.52) -- (-0.2, -0.1);

    \draw[\pgmLineColor,line width=0.75pt] (-0.75,0.75) circle (0.25);
    \draw (-0.75,0.75) node[anchor=center,align=center]{$\vec{\sigma}$};

    \draw[line width=0.5pt,->] (0, 0.5) -- (0, 0);

    \fill[\pgmBgColor] (0,-0.25) circle (0.25);
    \draw[line width=0.75pt] (0,-0.25) circle (0.25);
    \draw (0,-0.25) node[anchor=center,align=center]{$\vec{\tilde{y}}$};

\end{tikzpicture}}
        \caption{}
        \label{fig:pgmnlclass:NNAR}
    \end{subfigure}
    
    \caption{BBNs corresponding to (\subref{fig:pgmnlclass:NCAR}) the noise completely at random (NCAR), (\subref{fig:pgmnlclass:NAR}) noise at random (NAR) and (\subref{fig:pgmnlclass:NNAR}) noise not at random (NNAR) models from \cite{6685834}.}
    \label{fig:pgmnlclass}
\end{figure}
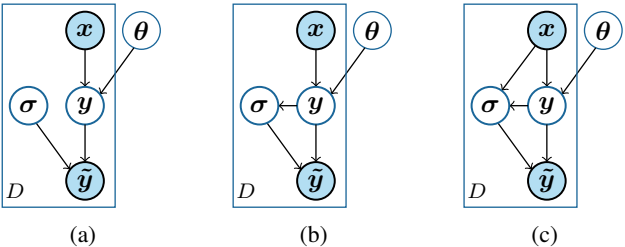

The inputs $D_{\vec{x}}$ in the training sets can be uncertain, either because the labels $D_{\vec{y}}$ are corrupted by noise \cite{NIPS2013_5073}, or because labels are missing for a number of points. In the case of \emphBulletPoint{noisy labels}, one should extend the BBN to add a new variable for the noisy labels $\vec{\tilde{y}}$ conditioned on $\vec{y}$ (Figure~\ref{fig:pgmnlssl:noisy}). It is common, as the noise level itself is often unknown, to add a variable $\vec{\sigma}$ to characterize the noise. Frenay~\etal~\cite{6685834}~proposed a taxonomy of the different approaches used to integrate $\vec{\sigma}$ in a PGM (Figure~\ref{fig:pgmnlclass}). They distinguish three cases: noise completely at random (NCAR); noise at random (NAR); and noise not at random (NNAR) models. In the NCAR model, the noise $\vec{\sigma}$ is independent of any other variable, \ie it is  homoscedastic. In the NAR model, $\vec{\sigma}$ is dependent on the true label $\vec{y}$ but remains independent of the features. NNAC models also account for the influence of the features $\vec{x}$, \eg if the level of noise in an image increases, then the probability that the image has been mislabeled also increases. Both NAR and NNAC models represent heteroscedastic, \ie the antonym of homoscedastic, noise.

These noise-aware PGMs are slightly more complex than a purely supervised BNN, as presented in Section \ref{sec:pgmimpl}. However, they can be treated in a similar fashion by deriving the formula for the posterior from the PGM (Equation \eqref{eq:bayesfrompgm}) and applying the chosen inference algorithm. For the NNAR model, the most generic stochastic model of the three described above (since the NCAR and NAR models are special cases of the NNAR model), the posterior becomes:
\begin{equation}\small
\label{eq:deeplearningbayesianposteriornoisy}
    p(\vec{y}, \vec{\sigma},  \vec{\theta}|D) ~\propto~ 
    p(D_{\vec{\tilde{y}}}|\vec{y},\vec{\sigma})p(\vec{\sigma}|D_{\vec{x}},\vec{y})p(\vec{y}|D_{\vec{x}},\vec{\theta})p(\vec{\theta}).
\end{equation}

\noindent During the prediction phase, $\vec{y}$ and $\vec{\sigma}$ can simply be discarded for each tuple $(\vec{y}, \vec{\sigma}, \vec{\theta})$ sampled from the posterior.

In the case of \emphBulletPoint{partially labeled data} (Figure~\ref{fig:pgmnlssl:semis}), also known as semi-supervised learning, the dataset $D$ is split into labeled $L$ and unlabeled $U$ examples. In theory, this PGM can be considered equivalent to the one used in the supervised learning case depicted in Figure~\ref{fig:pgmbnn}a, but in this case the unobserved data $U$ would bring no information. The additional information of unlabeled data comes from the prior and only the prior. Similar to traditional machine learning, the most common approaches to implement semi-supervised learning in Bayesian learning are either to use some type of data-driven regularization \cite{Tommi03oninformation} or to rely on pseudo labels \cite{sohn2020fixmatch}. 

\textbf{Data-driven regularization} implies modifying the prior assumptions, and thus the stochastic model, to be able to extract meaningful information from the unlabeled dataset $U$. There are two common ways to approach this process. \textbf{The first} one is to condition the prior distribution of the model parameters on the unlabeled examples to favor certain properties of the model, such as a decision boundary in a low density region, using a distribution $p(\vec{\theta} | U)$ instead of $p(\vec{\theta})$. This implies formulating the stochastic model as:
\begin{equation}
    \label{eq:sslcond}
    p(\vec{\theta} | D) 
    \propto p(L_{\vec{y}} | L_{\vec{x}}, \vec{\theta})p(\vec{\theta} | U) ,
\end{equation}

\noindent where $p(\vec{\theta} | U)$ is a prior with a consistency condition, as defined in Equation \eqref{eq:consistentprior}. The consistency condition usually expresses the fact that points that are close to each other should lead to the same prediction, \eg graph Laplacian norm regularization \cite{10.5555/1248547.1248632}.

\textbf{The second way} is to assume some kind of dependency across the observed and unobserved labels in the dataset. This type of semi-supervised Bayesian learning relies either on an undirected PGM \cite{JMLR:v12:yu11a} to build the prior or at least does not assume independence between different training pairs $(\vec{x}, \vec{y})$~\cite{6977247}. To keep things simple, we represent this fact by dropping the plate around $\vec{y}$ in Figure~\ref{fig:pgmnlssl:semis}. 
The posterior is written in the usual way (Equation \eqref{eq:deeplearningbayesianposterior}). The main difference is that $p(D_{\vec{y}}|D_{\vec{x}}, \vec{\theta})$ is chosen to enforce some kind of consistency across the dataset. For example, one can assume that two close points are likely to have similar labels $\vec{y}$ with a level of uncertainty that increases with the distance. 

Both approaches have a similar effect and the choice of one over the other will depend on the mathematical formulation favored to build the model.

The semi-supervised learning strategy can also be reformulated as having a weak predictor capable of generating some \textbf{pseudo labels} $\vec{\tilde{y}}$, sometimes with some confidence level. Many of the algorithms used for semi-supervised learning use an initial version of the model trained with the labeled examples \cite{lee2013pseudo} to generate the pseudo labels $\vec{\tilde{y}}$ and train the final model with $\vec{\tilde{y}}$. This is problematic for BNNs. When the prediction uncertainty is accounted for, reducing the uncertainty associated with the unlabeled data  becomes impossible, at least not without an additional hypothesis in the prior. Even if it is less current in practice, using a simpler model \cite{Li2019} to obtain the pseudo labels can help mitigate that problem.

\subsubsection{Data augmentation}

Data augmentation in machine learning is a strategy that is used to significantly increase the diversity of the data $D$ available to train deep models, without actually collecting new data. It relies on transformations that act on the input but have no or very low probability to change the label (or at least do so in a predictable way) to generate an augmented dataset $A(D)$. Examples of such transformations include applying rotations, flipping or adding noise in the case of images. Data augmentation is now at the forefront of state-of-the-art techniques in computer vision \cite{sohn2020fixmatch} and increasingly in natural language processing \cite{bari2020multimix}. 
 
The augmented dataset $A(D)$ could contain an infinite set of possible variants of the initial dataset $D$, \eg when using continuous transforms such as rotations or additional noise. To achieve this in practice, $A(D)$ is sampled on the fly during training, rather than generating in advance all possible augmentations in the training set.
This process is straightforward when training point estimate neural networks, but there are some subtleties when applying it in Bayesian statistics. The main concern is that the posterior of interest is $p(\vec{\phi}|D, Aug)$, where $Aug$ represents some knowledge about augmentation, not $p(\vec{\phi}|A(D), D)$, since $A(D)$ is not observed. From a Bayesian perspective, the additional information is brought by the knowledge of the augmentation process rather than by some additional data. Stated otherwise, the data augmentation is a part of the stochastic model (Figure~\ref{fig:pgmdataaugm}).

The idea is that if one is given data $D$, then one could also
have been given data $D'$, where each element in $D$ is replaced 
by an augmentation. Then, $D'$ is a different perspective
of the data $D$. To model this, we have the augmentation
distribution $p(\vec{x}'|\vec{x},Aug)$ that augments the observed data
using the augmentation model $Aug$ to generate
(probabilistically) $x'$, which represents data in the vicinity of $x$ (Figure~\ref{fig:pgmdataaugm}).
$x'$ can then be marginalized to simplify the stochastic model. The posterior is given by:
\begin{equation}\small
\label{eq:augmentation}
p( \vec{\theta} | \vec{x}, \vec{y}, Aug) \propto \left( \int_{\vec{x}'} p(\vec{y}|\vec{x}',\vec{\theta})p(\vec{x}'|\vec{x},Aug) d\vec{x}' \right) p(\vec{\theta}) .
\end{equation}

\noindent This is a probabilistic counterpart to vicinal risk
\cite{NIPS2000_1876}.

The integral in Equation (\ref{eq:augmentation}) can be approximated using Monte Carlo integration by sampling a small set of augmentations  $A_x$
according to $p( x' | x, Aug)$ and averaging:
\begin{equation}
\label{eq:augmentedtraining}
p( y | x, \vec{\theta} , Aug)  \approx \dfrac{1}{|A_x|} \sum_{x' \in A_x} p( y | x', \vec{\theta}) .
\end{equation}

\noindent When training using a Monte-Carlo-based estimate of the loss, $A_x$ can contain as few as a single element as long as it is resampled for each optimization iteration. This greatly simplifies the evaluation of Equation \eqref{eq:augmentedtraining}.

An extension of this approach works in the
context of semi-supervised learning. The prior can be designed to encourage consistency of predictions under augmentation \cite{xieArxiv19,sohn2020fixmatch},
using unlabeled data to build the samples for the consistency condition, as defined in Equation \eqref{eq:consistentprior}. Note that this does not add labeling to the unlabeled examples but only adds a term to encourage consistency between the labels for an unlabeled data point and its augmentation.

\subsubsection{Meta-learning, transfer learning, and self-supervised learning}

\textbf{Meta-learning} \cite{hospedales2020metalearning}, in the broadest sense, is the use of machine learning algorithms to assist in the training and optimization of other machine learning models. The meta knowledge acquired by meta-learning can be distinguished from standard knowledge in the sense that it is applicable to a set of related tasks rather than a single task.

\textbf{Transfer learning} designates methods that reuse some intermediate knowledge acquired on a given problem to address a different problem. In deep learning, it is used mostly for domain adaptation, when labeled data are abundant in a domain that is in some way similar to the domain of interest but scarce in the domain of interest \cite{5288526}. Alternatively, pre-trained models \cite{qiu2020pretrained} could be used to study large architectures whose complete training would be very computationally expensive.

\textbf{Self-supervised learning} is a learning strategy where the data themselves provide the labels \cite{9086055}. Since the labels directly obtainable from the data do not match the task of interest, the problem is approached by learning a pretext (or proxy) task in addition to the task of interest.
The use of self-supervision is now generally regarded as an essential step in
some areas. For instance, in natural language processing, most state-of-the-art methods use these pre-trained models \cite{qiu2020pretrained}. In addition, modern deep learning-based 3D object reconstruction~\cite{han2021image} and disparity estimation in stereo vision~\cite{laga2020survey} rely on self-supervised learning to overcome the time-consuming manual annotation of training data.

A common approach for meta-learning in Bayesian statistics is to recast the problem as hierarchical Bayes \cite{DBLP:conf/iclr/GrantFLDG18}, with the prior $p(\vec{\theta}_t|\vec{\xi})$ for each task conditioned on a new global variable $\vec{\xi}$ (Figure~\ref{fig:pgmtlssl:ml}). $\vec{\xi}$ can represent continuous metaparameters or discrete information about the structure of the BNN, \ie to learn probable functional models, or the underlying subgraph of the PGM, \ie to learn probable stochastic models. Multiple levels can be added to organize the tasks in a more complex hierarchy if needed. Here, we present only the case with one level since the generalization is straightforward. With this broad Bayesian understanding of meta-learning, both transfer learning and self-supervised learning are special cases of meta-learning. The general posterior becomes:
\begin{equation}
    p(\vec{\theta}, \vec{\xi} | D ) \propto \left( \prod_{t \in T} p(D^t_{\vec{y}}|D^t_{\vec{x}}, \vec{\theta}_t)p(\vec{\theta}_t|\vec{\xi}) \right) p(\vec{\xi}) .
\end{equation}

\noindent In practice, the problem is often approached with empirical Bayes (Section~\ref{sec:learnprior}), and only a point estimate $\vec{\hat{\xi}}$ is considered for the global variable, ideally the MAP estimate obtained by marginalizing $p(\vec{\theta}, \vec{\xi} | D )$ and selecting the most likely point, but this is not always the case.

In transfer learning, the usual approach would be to set $\vec{\hat{\xi}} = \vec{\theta}_m$, with $\vec{\theta}_m$ being the coefficients of the main task. The new prior can then be obtained from $\vec{\hat{\xi}}$, for example:
\begin{equation}
    p(\vec{\theta}|\vec{\xi}) = \mathcal{N}((\sel(\vec{\xi}), \vec{0}),\sigma \matr{I}) ,
\end{equation}

\noindent where $\sel$ is a selection of the parameters to transfer and $\sigma$ is a parameter to tune manually. Unselected parameters are assigned a new prior, with a mean of $0$ by convention. If a BNN has been trained for the main task, then $\sigma$ can be estimated from the previous posterior, with an increment to account for the additional uncertainty caused by the domain shift.

Self-supervised learning can be implemented in two steps. The \textbf{first} step learns the pretext task while the second one performs transfer learning. 
This can be considered overly complex but might be required if the pretext task has a high computational complexity (\eg BERT models in natural language processing \cite{qiu2020pretrained}). Recent contributions~\cite{9010283} have shown that jointly learning the pretext task and the final task (Figure~\ref{fig:pgmtlssl:ssl}) can improve the results obtained in self-supervised learning. This approach, which is closer to hierarchical Bayes, also allows setting the prior a single time while still retaining the benefits of self-supervised learning.

\section{Bayesian Inference algorithms}
\label{sec:inference}

A priori, a BNN does not require a learning phase as one just needs to sample the posterior and do model averaging; see Algorithm~\ref{alg:BNN_inf}. However, sampling the posterior is not easy in the general case. While the conditional probability $P(D|H)$ of the data and the probability $P(H)$ of the model are given by the stochastic model, the integral for the evidence term $\int_{H} P(D|H')P(H')dH'$ might be excessively difficult to compute. For nontrivial models, even if the evidence has been computed, directly sampling the posterior is prohibitively difficult due to the high dimensionality of the sampling space. Instead of using traditional methods, \eg  inversion sampling or rejection sampling to sample the posterior, dedicated algorithms are used. The most popular ones are Markov chain Monte Carlo (MCMC) methods \cite{10.1093/biomet/57.1.97}, a family of algorithms that exactly sample the posterior, or variational inference~\cite{BleiStat2018}, a method for learning an approximation of the posterior; see Figure~\ref{fig:workflow}. 

This section reviews these methods. First, in subsection \ref{sec:MCMC} and \ref{sec:vi}, we introduce MCMC and variational inference as they are used in traditional Bayesian statistics. Then, in subsection \ref{sec::approx}, we review different simplifications or approximations that have been proposed for deep learning. We also provide a practical example in the Supplementary Material (Practical example III), which compares different learning strategies.

\subsection{Markov Chain Monte Carlo (MCMC)}
\label{sec:MCMC}
The idea behind MCMC methods is to construct a Markov chain, a sequence of random samples $S_i$, which probabilistically depend only on the previous sample $S_{i-1}$, such that the $S_i$ are distributed following a desired distribution. Unlike standard sampling methods such as rejection or inversion sampling, most MCMC algorithms require an initial burn-in time before the Markov chain converges to the desired distribution. Moreover, the successive $S_i$'s might be autocorrelated. This means that a large set of samples $\Theta$ has to be generated and subsampled to obtain approximately independent samples from the underlying distribution. 
The final collection of samples $\Theta$ has to be stored after training, which is expensive for most deep learning models.

Despite their inherent drawbacks, MCMC methods can be considered among the best available and the most popular solutions for sampling from exact posterior distributions in Bayesian statistics \cite{Bardenet2017}.
However, not all MCMC algorithms are relevant for Bayesian deep learning. Gibbs sampling \cite{George92explainingthe}, for example, is very popular in general statistics and unsupervised machine learning but is very ill-suited for BNNs. The most relevant MCMC method for BNNs is the Metropolis-Hastings algorithm \cite{doi:10.1080/00031305.1995.10476177}. The property that makes the Metropolis-Hasting algorithm popular is that it does not require knowledge about the exact probability distribution $P(\vec{x})$ to sample from. Instead, a function $f(\vec{x})$ that is proportional to that distribution is sufficient. This is the case of a Bayesian posterior distribution, which is usually quite easy to compute except for the evidence term.

\emphBulletPoint{The Metropolis-Hasting algorithm}, see Algorithm \ref{alg:MH}, starts with a random initial guess, $\vec{\theta}_0$, and then samples a new candidate point $\theta'$ around the previous $\theta$, using a proposal distribution $Q(\theta'|\theta)$. If $\theta'$ is more likely than $\theta$ according to the target distribution, it is accepted. If it is less likely, it is accepted with a certain probability or rejected otherwise.

\begin{algorithm}[ht]
\begin{algorithmic}
{\small
    \STATE Draw $\vec{\theta}_0 \sim Initial~probability~distribution$;
    \WHILE{$n=0$ \TO $N$}
        \STATE Draw $\vec{\theta}' \sim Q(\vec{\theta}'|\vec{\theta}_{n})$;
        \STATE $p = \min\left(1, \dfrac{Q(\vec{\theta}'|\vec{\theta}_{n})}{Q(\vec{\theta}_{n}|\vec{\theta}')} \dfrac{f(\vec{\theta}')}{f(\vec{\theta}_{n})} \right)$;
        \STATE Draw $k \sim \Bernoulli(p)$;
        \IF{$k$}
            \STATE $\vec{\theta}_{n+1} = \vec{\theta}'$;
            \STATE $n = n+1$;
        \ENDIF
    \ENDWHILE
}
\end{algorithmic}
\caption{Metropolis-Hasting algorithm.
\label{alg:MH}}
\end{algorithm}

The acceptance probability $p$ can be simplified if $Q$ is chosen to be symmetric, \ie $Q(\vec{\theta}'|\vec{\theta}_{n}) = Q(\vec{\theta}_{n}|\vec{\theta}')$. The formula for the acceptance rate then becomes:
\begin{equation}
    p = \min\left(1, \dfrac{f(\vec{\theta}')}{f(\vec{\theta}_{n})} \right) .
\end{equation}

\noindent In this situation, the algorithm is simply called \emphBulletPoint{the Metropolis method}. Common choices for $Q$ can be a normal distribution $Q(\vec{\theta}'|\vec{\theta}_{n}) = \mathcal{N}(\vec{\theta}_{n}, \sigma^2)$, or a uniform distribution $Q(\vec{\theta}'|\vec{\theta}_{n}) = \mathcal{U}(\vec{\theta}_{n}-\vec{\varepsilon}, \vec{\theta}_{n}+\vec{\varepsilon})$, centered around the previous sample. To deal with non-symmetric proposal distributions, \eg to accommodate a constraint in the model such as a bounded domain, one has to take into account the correction term imposed by the full Metropolis-Hasting algorithm.

The spread of $Q(\vec{\theta}'|\vec{\theta}_{n})$ has to be tweaked. If it is too large, the rejection rate will be too high. If it is too small, the samples will be more autocorrelated. There is no general method to tweak those parameters. However, a clever strategy to obtain the new proposed sample $\vec{\theta}'$ can reduce their impact. This is why the Hamiltonian Monte-Carlo method has been proposed.

\emphBulletPoint{The Hamiltonian Monte Carlo algorithm (HMC)} \cite{neal2011mcmc}  is another example of Metropolis-Hasting algorithms for continuous distributions. It is designed with a clever scheme to draw a new proposal $\vec{\theta}'$ to ensure that as few samples as possible are rejected and there is as few correlation as possible between samples. In addition, the HMC's  burn-in time is extremely short compared to  the standard Metropolis-Hasting algorithm.

Most software packages for Bayesian statistics implement the \emphBulletPoint{No-U-Turn sampler} (NUTS for short) \cite{hoffman2014no}, which is an improvement over the classic HMC algorithm allowing the hyperparameters of the algorithm to be automatically tweaked instead of manually setting them.

\subsection{Variational inference}
\label{sec:vi}
MCMC algorithms are the best tools for sampling from the exact posterior. 
However, their lack of scalability has made them less popular for BNNs, given the size of the models under consideration. Variational inference \cite{BleiStat2018}, which scales better than MCMC algorithms, gained considerable popularity. Variational inference is not an exact method. Rather than allowing sampling from the exact posterior, the idea is to have a distribution $q_{\phi}(H)$, called the variational distribution, parametrized by a set of parameters $\vec{\phi}$. The values of the parameters $\vec{\phi}$ are then learned such that the variational distribution $q_{\phi}(H)$ is as close as possible to the exact posterior $P(H|D)$. The measure of closeness that is commonly used is the Kullback-Leibler divergence (KL-divergence) \cite{10.1214/aoms/1177729694}. It measures the differences between probability distributions based on Shannon's information theory \cite{6773024}. The KL-divergence represents the average number of additional bits required to encode a sample from $P$ using a code optimized for $q$. For Bayesian inference, it is computed as:
\begin{equation}
    \label{eq:KLdiv}
    D_{KL}(q_{\phi} || P) = \int_{H} q_{\phi}(H') \log \left( \dfrac{q_{\phi}(H')}{P(H'|D)} \right) dH'.
\end{equation}

\noindent There is an apparent problem here, which is, to compute $D_{KL}(q_{\phi} || P)$, one needs to compute $P(H|D)$ anyway. To overcome this, a different, easily derived formula called the \emph{evidence lower bound}, or ELBO, serves as a loss:
\begin{equation}\small
    \label{eq:elbo}
    \int_{H} q_{\phi}(H') \log \left( \dfrac{P(H',D)}{q_{\phi}(H') } \right) dH' = \log(P(D)) - D_{KL}(q_{\phi} || P) .
\end{equation}
Since $\log(P(D))$ only depends on the prior, minimizing $D_{KL}(q_{\phi} || P) $
is equivalent to maximizing the ELBO.

The most popular method to optimize the ELBO is stochastic variational inference (SVI) \cite{Hoffman2013}, which is in fact the stochastic gradient descent method applied to variational inference. This allows the algorithm to scale to the  large datasets that are encountered in modern machine learning, since the ELBO can be computed on a single mini-batch at each iteration.

Convergence, when learning the posterior with SVI, will be slow compared to the usual gradient descent. Moreover, most implementations use a small number of samples to evaluate the ELBO, often just one, before taking a gradient step. In other words, the ELBO estimate will be noisy at each iteration.

In traditional machine learning and statistics, $q_{\phi}(H)$ is mostly constructed from distributions in the exponential family, \eg multivariate normal \cite{GravesNIPS2011}, Gamma and Dirichlet distributions. The ELBO can then be dramatically simplified into components \cite{ghahramani00propagation} leading to a generalization of the well-known expectation-maximization algorithm. To account for correlations between the large number of parameters, certain approximations are made. For instance, block diagonal~\cite{ritter2018a} or low rank plus diagonal \cite{MaddoxDL2019} covariance matrices can be used to reduce the number of variational parameters $\vec{\phi}$ from $\mathcal{O}(n^2)$ to $\mathcal{O}(n)$, where $n$ is the number of model parameters $\vec{\theta}$. Appendix~\ref{app:matrSimpl} gives more details on how these simplifications are implemented in practice.

\subsection{Bayes by backpropagation}
\label{sec:bayesbackprop}

\begin{figure}[tb]
    \centering
        \includegraphics[width=0.45\textwidth]{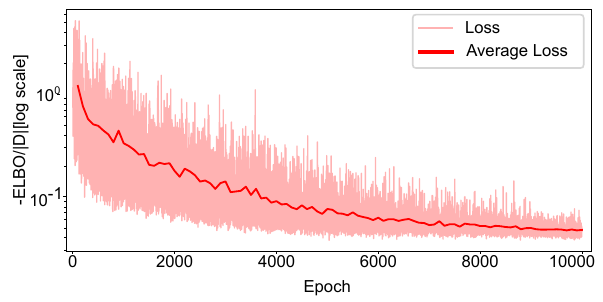}
    
    \caption{Typical training curve for Bayes-by-backprop.}
    \label{fig:trainingcurve}
\end{figure}

Variational inference offers a good mathematical tool for Bayesian inference, but it needs to be adapted to deep learning. The main problem is that stochasticity stops backpropagation from functioning at the internal nodes of a network \cite{Buntine_1994}. Different solutions have been proposed to mitigate this problem, including probabilistic backpropagation \cite{hernndezlobato2015probabilistic} or Bayes-by-backprop \cite{pmlr-v37-blundell15}. The latter may appear more familiar to deep learning practitioners. We will thus focus on Bayes-by-backprop in this tutorial. Bayes-by-backprop is indeed a practical implementation of SVI combined with a reparametrization trick \cite{kingma2019introduction} to ensure backpropagation works as usual.

The idea is to use a random variable $\varepsilon \sim q(\varepsilon)$ as a nonvariational source of noise. $\vec{\theta}$ is not sampled directly but obtained via a deterministic transformation $t(\varepsilon, \vec{\phi})$ such that $\vec{\theta} = t(\varepsilon, \vec{\phi})$ follows $q_{\phi}(\vec{\theta})$. $\varepsilon$ is sampled and thus changes at each iteration but can still be considered a constant with regard to other variables. All other transformations being non-stochastic, backpropagation works as usual for the variational parameters $\vec{\phi}$, meaning the training loop can be implemented analogous to the training loop of a non-stochastic neural network; see Algorithm~\ref{alg:bayesbackprop}. The general formula for the ELBO becomes:
\begin{equation}\small
    \label{eq:elbo_general}
    \int_{\varepsilon} q_{\phi}(t(\varepsilon, \vec{\phi})) \log \left( \dfrac{P(t(\varepsilon, \vec{\phi}),D)}{q_{\phi}(t(\varepsilon, \vec{\phi})) } \right) \left| \text{Det}(\nabla_{\varepsilon} t(\varepsilon, \vec{\phi}))\right|  d\varepsilon.
\end{equation}

\noindent This is tedious to work with. Instead, to estimate the gradient of the ELBO, Blundell \etal  \cite{pmlr-v37-blundell15} proposed to use the fact that if $q_{\phi}(\vec{\theta})d\vec{\theta} = q(\varepsilon) d\varepsilon $, then for a differentiable function $f(\vec{\theta},\vec{\phi})$, we have:
\begin{equation}\footnotesize
    \label{eq:elbo_trick}
    \dfrac{\partial}{\partial \vec{\phi}} \int_{\vec{\phi}} q_{\phi}(\vec{\theta}') f(\vec{\theta}',\vec{\phi}) d\vec{\theta}' = \int_{\varepsilon} q(\varepsilon) \left( \dfrac{\partial f(\vec{\theta},\vec{\phi})}{\partial \vec{\theta}}\dfrac{\partial \vec{\theta}}{\partial \vec{\phi}} + \dfrac{\partial f(\vec{\theta},\vec{\phi})}{\partial \vec{\phi}} \right) d\varepsilon.
\end{equation}

\noindent A proof is provided in \cite{pmlr-v37-blundell15}. We also provide in Appendix \ref{app:proofProbInt} an alternative proof to give more details on when we can assume $q_{\phi}(\vec{\theta})d\vec{\theta} = q(\varepsilon) d\varepsilon $. A sufficient condition is for $t(\varepsilon, \vec{\phi})$ to be invertible with respect to $\varepsilon$ and the distributions $q(\varepsilon)$ and $q_{\phi}(\vec{\theta})$ to not be degenerated.

For the case where the weights are treated as stochastic variables, and thus the hypothesis $H$, the training loop can be implemented as described in Algorithm \ref{alg:bayesbackprop}.

\begin{algorithm}[h]
\begin{algorithmic}
{\small
    \STATE $\vec{\phi} = \vec{\phi}_0$;
    \FOR{$i=0$ \TO $N$}
        \STATE Draw $\varepsilon \sim q(\varepsilon)$;
        \STATE $\vec{\theta} = t(\varepsilon, \vec{\phi})$;
        \STATE $f(\vec{\theta}, \vec{\phi}) = \log(q_{\phi}(\vec{\theta})) - \log(p(D_{\vec{y}}|D_{\vec{x}},\vec{\theta})p(\vec{\theta}))$;
        \STATE $\Delta_{\phi} f = \text{backprop}_{\phi}(f)$;
        \STATE $\vec{\phi} = \vec{\phi} - \alpha \Delta_{\phi} f$;
    \ENDFOR
}
\end{algorithmic}
\caption{Bayes-by-backprop algorithm.}
\label{alg:bayesbackprop}
\end{algorithm}

The objective function $f$ corresponds to an estimate of the ELBO from a single sample. This means that the gradient estimate will be noisy. The convergence graph will also be much more noisy than in the case of classic backpropagation (Figure~\ref{fig:trainingcurve}). To obtain a better estimate of the convergence, one can average the loss over multiple epochs.

Since algorithm \ref{alg:bayesbackprop} is very similar to the classical training loop for point estimate deep learning, most techniques used for optimization in deep learning are straightforward to use for Bayes-by-backprop. For example, it is perfectly fine to use the ADAM optimizer \cite{kingma2014adam} instead of the stochastic gradient descent.

Note also that, if Bayes-by-backprop is presented for BNNs with stochastic weights,  adapting it for BNNs with stochastic activations is straightforward. In that case, the activations $\vec{l}$ represent the hypothesis $H$ and the weights $\vec{\theta}$ are part of the variational parameters $\vec{\phi}$.

\subsection{Learning the prior}
\label{sec:learnprior}

Learning the prior and the posterior afterwards is possible. This is meaningful if most aspects of the prior can be set using prior knowledge, and only a limited set of free parameters of the prior are learned before obtaining the posterior.
In standard Bayesian statistics, this is known as \textbf{empirical Bayes}. This is usually a valid approximation when the dimensions of the prior parameters being learned are significantly smaller than the dimensions of the model parameters.

Given a parametrized prior distribution $p_{\xi}(H)$, maximizing the likelihood of the data is a good method to learn the parameters $\vec{\xi}$:
\begin{eqnarray}
    \nonumber\vec{\hat{\xi}} &=& \argmax_{\xi} P(D|\vec{\xi})\\
                    &=& \argmax_{\xi} \int_{H} p_{\xi}(D|H')p_{\xi}(H')dH'.
\end{eqnarray}

\noindent In general, directly finding  $\vec{\hat{\xi}}$ is an intractable problem. However, 
 when using variational inference, the ELBO is the log likelihood of the data minus the KL-divergence of $q_{\phi}(\vec{\theta})$ and prior (Eq.~\ref{eq:elbo}):
\begin{equation}
    \log(P(D|\vec{\xi})) = \text{ELBO}  + D_{KL}(q_{\phi} || P).
\end{equation}

\noindent This property means that maximizing the ELBO, now a function of both $\vec{\xi}$ and $\vec{\phi}$, is equivalent to maximizing a lower bound on the log likelihood of the data.  This lower bound becomes tighter when $q_{\phi}$ is from a general family of probability distributions with more flexibility to fit the exact posterior $P(\vec{\theta}|D)$. The Bayes-by-backprop algorithm presented in Section \ref{sec:bayesbackprop} needs only to be slightly modified to include the additional parameters in the training loop; see Algorithm~\ref{alg:bbbpp}.

\begin{algorithm}[H]
\label{alg:priorbayesbackprop}
\begin{algorithmic}
{\small
    \STATE $\vec{\xi} = \vec{\xi}_0$;
    \STATE $\vec{\phi} = \vec{\phi}_0$;
    \FOR{$i=0$ \TO $N$}
        \STATE Draw $\varepsilon \sim q(\varepsilon)$;
        \STATE $\vec{\theta} = t(\varepsilon, \vec{\phi})$;
        \STATE $f(\vec{\theta}, \vec{\phi}, \vec{\xi}) = \log(q_{\phi}(\vec{\theta})) - \log(p_{\vec{\xi}}(D_{\vec{y}}|D_{\vec{x}},\vec{\theta})p_{\vec{\xi}}(\vec{\theta}))$;
        \STATE $\Delta_{\xi} f = \text{backprop}_{\xi}(f)$;
        \STATE $\Delta_{\phi} f = \text{backprop}_{\phi}(f)$;
        \STATE $\vec{\xi} = \vec{\xi} - \alpha_{\xi} \Delta_{\xi} f$;
        \STATE $\vec{\phi} = \vec{\phi} - \alpha_{\phi} \Delta_{\phi} f$;
    \ENDFOR
}
\end{algorithmic}
\caption{Bayes-by-backprop with parametric prior.}
\label{alg:bbbpp}
\end{algorithm}

\subsection{Inference algorithms adapted for deep learning}
\label{sec::approx}

\begin{figure*}[t]
    \centering
        \includegraphics[width=\textwidth]{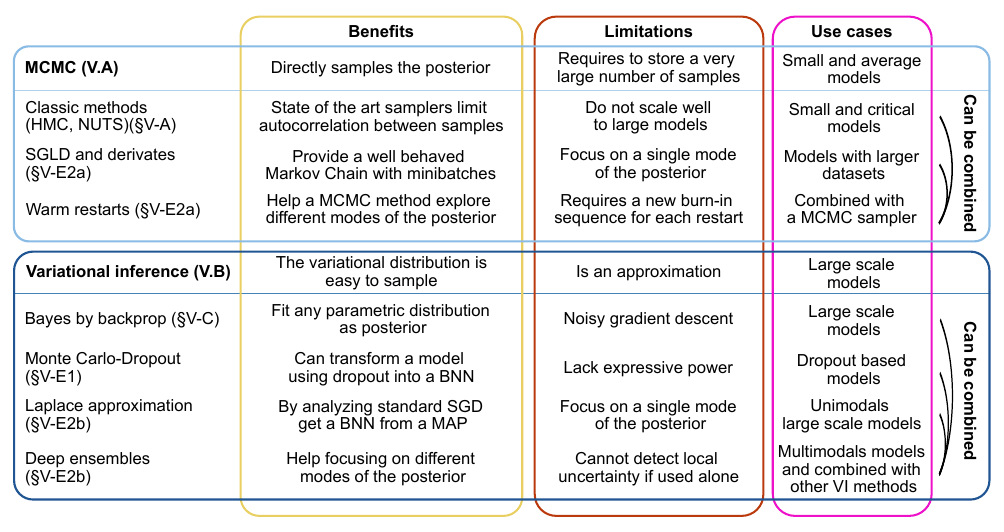}
    
    \caption{Summary of the different inference approaches used to train a BNN with their benefits, limitations and use cases.}
    \label{fig:approxclassification}
\end{figure*}

We presented thus far the fundamental theory to design and train BNNs. However, the aforementioned methods are still not easily applicable to most large scale architectures currently used in deep learning. Recent research has also shown that being only approximately Bayesian is sufficient to achieve a correctly calibrated model with uncertainty estimates \cite{kristiadi2020bayesian}.
This section presents how inference algorithms were adapted for deep learning, resulting in more efficient methods. Specific inference methods can still be classified as MCMC algorithms, \ie they generate a sequence of samples from the posterior, or as a form of variational inference, \ie they learn the parameters of an intermediate distribution to approximate the posterior. All methods are summarized in Figure~\ref{fig:approxclassification}.

\subsubsection{Bayes via Dropout}
\label{sec:mc-dropout}

Dropout has initially been proposed as a regularization method \cite{JMLR:v15:srivastava14a}. It works by applying multiplicative noise to the target layer. The most commonly used type of noise is Bernoulli noise, but other types such as the Gaussian noise for Gaussian Dropout \cite{JMLR:v15:srivastava14a} might be used instead.

Dropout is usually turned off at evaluation time, but leaving it on results in a distribution for the output predictions \cite{GalDL2015, Yingzhen2017}. It turns out that this procedure, called Monte Carlo Dropout, is in fact variational inference with a variational distribution defined for each weight matrix as:
\begin{equation}
\label{eq:dropoutvarpost}
\begin{array}{l}
     \vec{z}_{i,j} \sim \text{Bernoulli}(p_i), \\
     \matr{W}_i = \matr{M}_i \cdot \text{diag}(\vec{z}_i), \\
\end{array}
\end{equation}

\noindent with $\vec{z}_i$ being the random activation coefficients and $\matr{M}_i$ the matrix of weights before dropout is applied. $p_i$ is the activation probability for layer $i$ and can be learned or set manually.

When used to train a BNN, dropout should not be seen as a regularization method, as it is part of the variational posterior, not the prior. This means that it should be coupled with a different type of regularization \cite{pmlr-v80-hron18a}, \eg $\ell^2$ weight penalization. The equivalence between the objective function $\mathbf{L}_{\text{dropout}}$ used for training with dropout and $\ell^2$ weight regularization, which is defined as:
\begin{equation}
    \mathbf{L}_{\text{dropout}} = \dfrac{1}{N} \sum_{D} f(\vec{y}, \vec{\hat{y}}) + \lambda \sum_{\vec{\theta}} \theta_i^2,
\end{equation}
\noindent
and the ELBO, assuming a normal prior on the weights and the distribution presented in Equation \ref{eq:dropoutvarpost} as variational posterior, has been demonstrated in~\cite{GalDL2015}. The argument is similar to the one presented in Section \ref{sec:reg-prior}.

MC-Dropout is a very convenient technique to perform Bayesian deep learning. It is straightforward to implement and requires little additional knowledge or modeling effort compared to traditional methods. It often leads to a faster training phase compared to other variational inference approaches. If a model has been trained with dropout layers, which are quite widespread in today's deep learning architectures, and an additional form of regularization acting as prior, it can be used as a BNN without any need to be retrained.

On the other hand, MC-Dropout might lack some expressiveness and may not fully capture the uncertainty associated with the model predictions \cite{pmlr-v119-chan20a}. It also lacks flexibility compared to other Bayesian methods for online or active learning.

\subsubsection{Bayes via stochastic gradient descent}

Stochastic gradient descent (SGD) and related algorithms are at the core of modern machine learning. The initial goal of SGD is to provide an algorithm that converges to an optimal point estimate solution while having only noisy estimates of the gradient of the objective function. This is especially useful when the training data has to be split into mini-batches. The parameter update rule at time $t$ can be written as:
\begin{equation}
    \Delta \vec{\theta}_t = \frac{\epsilon_{t}}{2} \left( \dfrac{N}{n} \nabla \log(p(D_{t,\vec{y}}|D_{t,\vec{x}},\vec{\theta}_{i})) + \nabla \log( p(\vec{\theta}_t) ) \right),
\end{equation}

\noindent where $D_t$ is a mini-batch subsampled at time $t$ from the complete dataset $D$, $\epsilon_{t}$ is the learning rate at time $t$, $N$ is the size of the whole dataset and $n$ the size of the mini-batch.

SGD, or related optimization algorithms such as ADAM \cite{kingma2014adam}, can be reinterpreted as a Markov Chain algorithm \cite{mandt2017stochastic}. Usually, the hyperparameters of the algorithm are tweaked to ensure that the chain converges to a Dirac distribution, whose position gives the final point estimate. This is done by reducing $\epsilon_{t}$ toward zero while ensuring that $\sum_{t=0}^{\infty} \epsilon_{t} = \infty$. However, if the learning rate is reduced toward a strictly positive value, the underlying Markov Chain will converge to a stationary distribution. If a Bayesian prior is accounted for in the objective function, then this stationary distribution can be an approximation of the corresponding posterior.

\paragraph{MCMC algorithms based on the SGD dynamic}

\begin{algorithm}[b]
\begin{algorithmic}
{\small
    \STATE Draw $\vec{\theta}_0 \sim Initial~probability~distribution$;
    \FOR{$t=0$ \TO $E$}
        \STATE Select a mini-batch $D_{t,\vec{y}},D_{t,\vec{x}} \subset D$;
        \STATE $f(\vec{\theta}_{t}) = \dfrac{N}{n} \log(p(D_{t,\vec{y}}|D_{t,\vec{x}},\vec{\theta}_{t})) + \log(p(\vec{\theta}_{t}))$;
        \STATE $\Delta_{\vec{\theta}} f = \text{backprop}_{\vec{\theta}}(f)$;
        \STATE Draw $\eta_t \sim \mathcal{N}(0, \epsilon_{t})$;
        \STATE $\vec{\theta}_{t+1} = \vec{\theta}_{t} - \left( \dfrac{\epsilon_t}{2} \Delta_{\theta} f + \eta_t \right)$;
    \ENDFOR
}
\end{algorithmic}
\caption{Stochastic Gradient Langevin Dynamic (SGLD).}
\label{alg:sgld}
\end{algorithm}

To approximately sample the posterior using the SGD algorithm, a specific MCMC method, called stochastic gradient Langevin dynamic (SGLD) \cite{welling2011bayesian}, has been developed, see Algorithm~\ref{alg:sgld}. Coupling SGD with Langevin dynamic leads to a slightly modified update step:
\begin{equation}
    \begin{array}{l}
         \Delta \vec{\theta}_t = \frac{\epsilon_{t}}{2} \left( \dfrac{N}{n} \nabla \log( p(D_t, \vec{\theta}_t) ) + \nabla \log( p(\vec{\theta}_t) ) \right) + \eta_t, \\
         \eta_t \sim \mathcal{N}(0, \epsilon_{t}).
    \end{array}
\end{equation}

\noindent Welling \etal \cite{welling2011bayesian}  showed  that this method leads to a Markov Chain that samples the posterior if $\epsilon_{t}$ goes toward zero. However, in that case, the successive samples become increasingly autocorrelated. To address this problem, the authors proposed to stop reducing $\epsilon_{t}$ at some point, thus making the samples only an approximation of the posterior. Nevertheless, SGLD offers better theoretical guarantees compared to other MCMC methods when the dataset is split into mini-batches. This makes the algorithm useful in Bayesian deep learning.

To favor the exploration of the posterior, one can use warm restart of the algorithm \cite{seedat2019calibrated}, \ie restarting the algorithm at a new random position $\vec{\theta}_0$ and with a large learning rate $\epsilon_0$. This offers multiple benefits. The main one is to avoid the mode collapse problem \cite{NIPS2017_7219}. In the case of a BNN, the true Bayesian posterior is usually a complex multimodal distribution, as multiple and sometimes not equivalent parametrizations $\vec{\theta}$ of the network can fit the training set. Favoring exploration over precise reconstruction can help to achieve a better picture of those different modes. Then, as parameters sampled from the same mode are likely to make the model generalize in a similar manner, using warm restarts enables a much better estimate of the epistemic uncertainty when processing unseen data, even if this approach provides only a very rough approximation of the exact posterior.

Similar to other MCMC methods, this approach still suffers from a huge memory footprint. This is why a number of authors have proposed methods that are more similar to traditional variational inference than to an MCMC algorithm.

\paragraph{Variational Inference based on SGD dynamic}
\label{sec::approx::SGDD::VI}

\begin{figure}[t]
    \centering
    \includegraphics[width=0.45\textwidth]{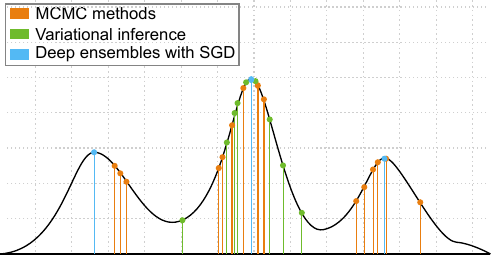}
    \caption{Different techniques for sampling the posterior. MCMC algorithms sample the true posterior but successive samples might be correlated, Variational Inference uses a parametric distribution that can suffer from mode collapse while deep ensembles focus on the modes of the distribution.}
    \label{fig:posteriorexpl}
\end{figure}

Instead of an MCMC algorithm, SGD dynamic can be used as a variational inference method to learn a distribution by using Laplace approximation. Laplace approximation fits a Gaussian posterior by using the maximum a posteriori estimate as the mean and the inverse of the Hessian $\matr{H}$ of the loss (assuming the loss is the log likelihood) as covariance matrix:
\begin{equation}
    p(\vec{\theta}|D) \approx \mathcal{N}(\vec{\hat{\theta}},\matr{H}^{-1}).
\end{equation}

\noindent Computing $\matr{H}^{-1}$ is usually intractable for large neural network architectures. Thus,  approximations are used, most of the time by analysing the variance of the gradient descent algorithm \cite{ritter2018a, MaddoxDL2019, Khan2018FastAS}. However, if those methods are able to capture the fine shape of one mode of the posterior, they cannot fit multiple modes.

Lakshminarayanan \etal~\cite{NIPS2017_7219}  proposed using warm restarts to obtain different point estimate networks instead of fitting a parametric distribution. This method, called deep ensembles; see Figure~\ref{fig:posteriorexpl} and Algorithm~\ref{alg:deepensembles}, has been used in the past to perform model averaging. The main contribution of \cite{NIPS2017_7219} was to show that it enables well-calibrated error estimates. While Lakshminarayanan \etal~\cite{NIPS2017_7219} claim that their method is non-Bayesian, it has been shown that their approach can still be understood from a Bayesian point of view \cite{WilsonDL2020, pearce2018uncertainty}. When regularization is used, the different point estimates should correspond to modes of a Bayesian posterior. 
This can be interpreted as approximating the posterior with a distribution parametrized as multiple Dirac deltas, \ie
\begin{equation}
    q_{\vec{\phi}}(\vec{\theta}) = \sum_{\vec{\theta}_i \in \vec{\phi}} \alpha_{\vec{\theta}_i} \delta_{\vec{\theta}_i}(\vec{\theta}) ,
\end{equation}

\noindent with the $\alpha_{\vec{\theta}_i}$ being positive constants such that their sum is equal to one. This approach can be seen as a form of variational inference. Note however that, for a variational distribution containing Dirac deltas, computing the ELBO in a sense that is meaningful for traditional optimization is impossible.

\begin{algorithm}[t]
\begin{algorithmic}
{\small
    \FOR{$i=0$ \TO $R$}
        \STATE Draw $\vec{\theta}_0 \sim Initial~probability~distribution$;
        \STATE $\epsilon_{t} = \epsilon_{0}$
        \FOR{$j=0$ \TO $N$}
            \STATE $f(\vec{\theta}_{i,j}) = \log(p(D_{\vec{y}}|D_{\vec{x}},\vec{\theta}_{i,j})) + \log(p(\vec{\theta}_{i,j}))$;
            \STATE $\Delta_{\theta} f = \text{backprop}_{\theta}(f)$;
            \STATE $\vec{\theta}_{i} = \vec{\theta}_{i} - \alpha_{\vec{\theta}} \Delta_{\theta} f$;
        \ENDFOR
    \ENDFOR
}
\end{algorithmic}
\caption{Deep ensembles.}
\label{alg:deepensembles}
\end{algorithm}

\section{Simplifying Bayesian Neural networks}
\label{sec:simplifications}

After training a BNN, one has to use Monte Carlo at evaluation time to estimate uncertainty. This is a major drawback of BNNs. For MCMC-based methods, storing a large set of parametrizations $\Theta$ is also not practical. This section presents mitigation strategies reported in the literature.

\subsection{Bayesian inference on the (n-)last layer(s) only}

\begin{figure}[b]
    \centering
        \includegraphics{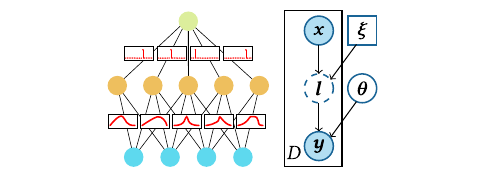}
    
    \caption{PGM and BNN corresponding to a last-layer model. }
    \label{fig:lastlayer}
\end{figure}

The architecture of deep neural networks makes it quite redundant to account for uncertainty for a large number of successive layers. Instead, recent research aims to use only a few stochastic layers, usually positioned at the end of the networks \cite{zeng2018relevance, brosse2020lastlayer}; see Figure~\ref{fig:lastlayer}. With only a few stochastic layers, training and evaluation can be drastically sped up while still obtaining meaningful results from a Bayesian perspective. This approach can be seen as learning a point estimate transformation followed by a shallow BNN.

Training a BNN with some non-stochastic layers is similar to learning the parameters for the prior presented in Section \ref{sec:learnprior}. The weights of the non-Bayesian layers should be considered as both prior and variational-posterior parameters.

\subsection{Bayesian teachers}

Using a BNN as a teacher is an idea derived from an approach used in Bayesian modeling \cite{10.1145/1102351.1102457}. The approach is to train a non-stochastic ANN to predict the marginal probability $p(\vec{y}|\vec{x},D)$ using a BNN as a teacher \cite{KorattikaraDL2015}. This is related to the idea of knowledge distillation \cite{hinton2015distilling,menonArxiv20} where possibly several pre-trained knowledge sources can be used to train a more functional system.

To do so, the KL-divergence between a parametric distribution $q_{\vec{\omega}}(\vec{y}|\vec{x})$, where $\vec{\omega}$ are the coefficients of the student network, and $p(\vec{y}|\vec{x},D)$ is minimized:
\begin{equation}
    \hat{\vec{\omega}} = \argmin_{\vec{\omega}} D_{KL}(p(\vec{y}|\vec{x},D)||q_{\vec{\omega}}(\vec{y}|\vec{x})).
\end{equation}

\noindent As this is intractable, Korattikara \etal \cite{KorattikaraDL2015} proposed a Monte Carlo approximation:
\begin{equation}
    \hat{\vec{\omega}} = \argmin_{\vec{\omega}} -\dfrac{1}{\left| \Theta \right|} \sum_{\vec{\theta}_i \in \Theta} \expec{p(\vec{y}|\vec{x}, \vec{\theta}_i)}{\log\left( q_{\vec{\omega}}(\vec{y}|\vec{x}) \right)}.
\end{equation}

\noindent Here, $\hat{\vec{\omega}}$ can be estimated using a training dataset $D'$ that contains only the features $\vec{x}$. During training, the probability $p(\vec{y}|\vec{x}, \vec{\theta})$ of the labels is given by the teacher BNN. Thus, $D'$ can be much larger than $D$. This helps the student network retain the calibration and uncertainty from the teacher.

Menon \etal~\cite{menonArxiv20} observed that, for classification problems, simply using the class probabilities output by a BNN teacher rather than one-hot labels helps the student to retain calibration and uncertainty from the teacher.

A Bayesian teacher can also be used to compress a large set of samples generated using MCMC \cite{Wang2018AdversarialDO}. Instead of storing $\Theta$, a generative model $G$ (\eg a GAN in \cite{Wang2018AdversarialDO}) is trained against the MCMC samples to generate the coefficients $\theta_i$ at evaluation time. This approach is similar to variational inference, with G representing a parametric distribution, but the proposed algorithm allows training a much more complex model than the distributions usually considered for variational inference.

\section{Performance metrics of Bayesian Neural Networks}
\label{sec:evaluation}

One big challenge with BNNs is how to evaluate their performance. They do not directly output a point estimate prediction $\vec{\hat{y}}$ but a conditional probability distribution $p(\vec{y}|\vec{x}, D)$, from which an optimal estimate $\vec{\hat{y}}$ can later be extracted. This means that both the predictive performance, \ie the ability of the model to give correct answers, and the calibration, \ie that the network is neither overconfident nor underconfident about its prediction, have to be assessed.

\textbf{The predictive performance}, sometimes called sharpness in statistics, of a network can be assessed by treating the estimator $\vec{\hat{y}}$ as the prediction. This procedure often depends on the type of data the network is meant to treat. Many different metrics, \eg mean square error (MSE), $\ell_n$ distances and cross-entropy, are used in practice. Covering these metrics is out of the scope of this tutorial. Instead, we refer the reader to \cite{Janocha2017} for more details.

\begin{figure}[t]
    \centering
    
    \includegraphics{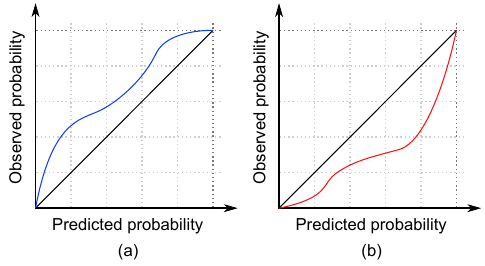}
    
    \caption{Examples of calibration curves for underconfident (a) and overconfident (b) models.}
    \label{fig:calibrationcurve}
\end{figure}

The standard method to assess \textbf{the model calibration} is a calibration curve, also called a reliability diagram \cite{Kendall2017, kuleshov2018accurate}. It is defined as a function $\check{p}: [0,1] \rightarrow [0,1]$ that represents the observed probability $\check{p}$, or empirical frequency, as a function of the predicted probability $\hat{p}$; see Figure~\ref{fig:calibrationcurve}. If $\check{p} < \hat{p}$, then the model is overconfident. Otherwise, it is underconfident. A well-calibrated model should have $\check{p} \cong \hat{p}$. Using this approach requires to first choose a set of events $\mathcal{E}$ with different predicted probabilities and then to measure the empirical frequency of each event using a test set $T$. 

For a binary classifier, the set of test events can be chosen as the set of all sets of datapoints with predicted probabilities of acceptance in interval $[p-\delta, p + \delta]$ for a chosen $\delta$, or alternatively $[0, p]$ or $[1-p, 1]$ for small datasets. The empirical frequency is given by:
\begin{equation}
\check{p} = \dfrac{ \sum_{\vec{\check{y}} \in T_{\vec{y}}} \vec{\check{y}} \cdot \mathbb{I}_{[\hat{p}-\delta, \hat{p} + \delta]}(\vec{\hat{y}}) }{ \sum_{\vec{\check{y}} \in T_{\vec{y}}} \mathbb{I}_{[\hat{p}-\delta, \hat{p} + \delta]}(\vec{\hat{y}}) }.
\end{equation}

\noindent
For multiclass classifiers, the calibration curve can be independently checked for each class against all the other classes. In this case, the problem is reduced to a binary classifier.

Regression problems are slightly more complex since the network does not output a confidence level, as in a classifier, but a distribution of possible outputs. The solution is to use an intermediate statistic with a known probability distribution. Assuming independence between the $\vec{\hat{y}}$ for a sufficiently large set of different randomly selected inputs $\vec{x}$, one can assume that the normalized sum of squared residuals ($\text{NSSR}$) follows a Chi-square law:
\begin{equation}
    \text{NSSR} = (\vec{\hat{y}} - \vec{\check{y}})^\intercal \matr{\Sigma}_{\vec{\hat{y}}}^{-1} (\vec{\hat{y}} - \vec{\check{y}}) \sim \chi_{Dim(\vec{y})}^2.
\end{equation}

\noindent This allows attributing to each data point in the test set $T$ a predicted probability that is the probability of observing a variance-normalized distance between the prediction and the true value equal to or lower than the measured NSSR. Formally, the predicted probability is computed as:
\begin{equation}
    \hat{p}_i = X_{\text{Dim}(\vec{y})}^2 \left( \text{NSSR}  \right) \quad \forall (\vec{y}_i,\vec{x}_i) \in T ,
\end{equation}

\noindent where $X_{\text{Dim}(\vec{y})}^2$ is the Chi-square cumulative distribution, with $\text{Dim}(\vec{y})$ degrees of freedom.
The observed probability can be computed as:
\begin{equation}
    \check{p}_i = \dfrac{1}{|T|} \sum_{j=1}^{|T|} \mathbb{I}_{[0,\infty)}(\hat{p}_j - \hat{p}_i).
\end{equation}

\noindent We present in the Supplementary Material a practical computation of such calibration curve for the sparse measure practical example (Practical example II).

Giving the whole calibration curve for a given stochastic model allows observing where the model is likely to be overconfident or underconfident. It also allows, to a certain extent, to recalibrate the model \cite{kuleshov2018accurate}. However, providing a summary measure to ease comparison or interpretation might also be necessary. The area under the curve (AUC) is a standard metric of the form:
\begin{equation}
    \text{AUC} = \int_{0}^{1} \check{p} d\hat{p}.
\end{equation}

\noindent An AUC of $0.5$ indicates that the model is, on average, well calibrated.

The distance from the actual calibration curve to the ideal calibration curve is also a good indicator for the calibration of a model:
\begin{equation}
    d(\check{p}, \hat{p}) = \sqrt{\int_{0}^{1} (\check{p} - \hat{p})^2 d\hat{p}}.
\end{equation}

\noindent When $d(\check{p}, \hat{p}) = 0$, then the model is perfectly calibrated.

Other measures have also been proposed. Examples include the expected calibration error and some discretized variants of the distance from the actual calibration curve to the ideal calibration curve~\cite{Nixon_2019_CVPR_Workshops}. 

\section{Conclusion}
\label{sec::conclusion}

This tutorial covers the design, training and evaluation of BNNs. While their underlying principle is simple, \ie just training an ANN with some probability distribution attached to its weights, designing efficient algorithms remains very challenging. Nonetheless, the potential applications of BNNs are huge. In particular, BNNs constitute a promising paradigm allowing the application of deep learning in areas where a system is not allowed to fail to generalize without emitting a warning. Finally, Bayesian methods can help design new learning and regularization strategies. Thus, their relevance extends to traditional point estimate models.\\

\noindent\fbox{%
    \parbox{0.48\textwidth}{\textbf{Online resources for the tutorial}:\\
    https://github.com/french-paragon/BayesianNeuralNetwork-Tutorial-Metarepos\\
Supplementary material, as well as additional practical examples for the covered material with the corresponding source code implementation, have been provided.}}

\section{Acknowledgments}
This material is partially based on research sponsored by the Australian Research Council  {https://www.arc.gov.au/}
(Grants DP150100294 and DP150104251), and Air Force Research Laboratory and DARPA {https://afrl.dodlive.mil/tag/darpa/} under agreement number FA8750-19-2-0501.

{\small
\bibliographystyle{IEEEtran}
\bibliography{refs}

\begin{thebibliography}{100}
\providecommand{\url}[1]{#1}
\csname url@samestyle\endcsname
\providecommand{\newblock}{\relax}
\providecommand{\bibinfo}[2]{#2}
\providecommand{\BIBentrySTDinterwordspacing}{\spaceskip=0pt\relax}
\providecommand{\BIBentryALTinterwordstretchfactor}{4}
\providecommand{\BIBentryALTinterwordspacing}{\spaceskip=\fontdimen2\font plus
\BIBentryALTinterwordstretchfactor\fontdimen3\font minus
  \fontdimen4\font\relax}
\providecommand{\BIBforeignlanguage}[2]{{%
\expandafter\ifx\csname l@#1\endcsname\relax
\typeout{** WARNING: IEEEtran.bst: No hyphenation pattern has been}%
\typeout{** loaded for the language `#1'. Using the pattern for}%
\typeout{** the default language instead.}%
\else
\language=\csname l@#1\endcsname
\fi
#2}}
\providecommand{\BIBdecl}{\relax}
\BIBdecl

\bibitem{szegedy2013intriguing}
C.~Szegedy, W.~Zaremba, I.~Sutskever, J.~Bruna, D.~Erhan, I.~Goodfellow, and
  R.~Fergus, ``Intriguing properties of neural networks,'' \emph{arXiv preprint
  arXiv:1312.6199}, 2013.

\bibitem{10.1145/3194085.3194087}
Q.~Rao and J.~Frtunikj, ``Deep learning for self-driving cars: Chances and
  challenges,'' in \emph{Proceedings of the 1st International Workshop on
  Software Engineering for AI in Autonomous Systems}, ser. SEFAIS ’18, 2018,
  pp. 35--38.

\bibitem{8241753}
J.~{Ker}, L.~{Wang}, J.~{Rao}, and T.~{Lim}, ``Deep learning applications in
  medical image analysis,'' \emph{IEEE Access}, vol.~6, pp. 9375--9389, 2018.

\bibitem{CAVALCANTE2016194}
R.~C. Cavalcante, R.~C. Brasileiro, V.~L. Souza, J.~P. Nobrega, and A.~L.
  Oliveira, ``Computational intelligence and financial markets: A survey and
  future directions,'' \emph{Expert Systems with Applications}, vol.~55, pp.
  194--211, 2016.

\bibitem{8371683}
H.~M.~D. Kabir, A.~Khosravi, M.~A. Hosen, and S.~Nahavandi, ``Neural
  network-based uncertainty quantification: A survey of methodologies and
  applications,'' \emph{IEEE Access}, vol.~6, pp. 36\,218--36\,234, 2018.

\bibitem{etzBayes}
A.~Etz, Q.~F. Gronau, F.~Dablander, P.~A. Edelsbrunner, and B.~Baribault, ``How
  to become a {B}ayesian in eight easy steps: An annotated reading list,''
  \emph{Psychonomic Bulletin \& Review}, vol.~25, pp. 219--234, 2018.

\bibitem{polson2017deep}
N.~G. Polson, V.~Sokolov \emph{et~al.}, ``Deep learning: a {B}ayesian
  perspective,'' \emph{Bayesian Analysis}, vol.~12, no.~4, pp. 1275--1304,
  2017.

\bibitem{LampinenDLReview2001}
J.~Lampinen and A.~Vehtari, ``{B}ayesian approach for neural networks—review
  and case studies,'' \emph{Neural Networks}, vol.~14, no.~3, pp. 257 -- 274,
  2001.

\bibitem{TitteringtonDL2004}
D.~M. Titterington, ``Bayesian methods for neural networks and related
  models,'' \emph{Statist. Sci.}, vol.~19, no.~1, pp. 128--139, 02 2004.

\bibitem{Goan2020}
E.~Goan and C.~Fookes, \emph{Bayesian Neural Networks: An Introduction and
  Survey}.\hskip 1em plus 0.5em minus 0.4em\relax Cham: Springer International
  Publishing, 2020, pp. 45--87.

\bibitem{WangDLSurvey2016}
H.~Wang and D.-Y. Yeung, ``A survey on bayesian deep learning,'' \emph{ACM
  Comput. Surv.}, vol.~53, no.~5, Sep. 2020.

\bibitem{WilsonDL2020}
\BIBentryALTinterwordspacing
A.~G. Wilson and P.~Izmailov, ``Bayesian deep learning and a probabilistic
  perspective of generalization,'' \emph{CoRR}, vol. abs/2002.08791, 2020.
  [Online]. Available: \url{http://arxiv.org/abs/2002.08791}
\BIBentrySTDinterwordspacing

\bibitem{Goodfellow-et-al-2016}
I.~Goodfellow, Y.~Bengio, and A.~Courville, \emph{Deep Learning}.\hskip 1em
  plus 0.5em minus 0.4em\relax MIT Press, 2016,
  \url{http://www.deeplearningbook.org}.

\bibitem{Yang2021}
S.~C.-H. Yang, W.~K. Vong, R.~B. Sojitra, T.~Folke, and P.~Shafto, ``Mitigating
  belief projection in explainable artificial intelligence via bayesian
  teaching,'' \emph{Scientific Reports}, vol.~11, no.~1, p. 9863, May 2021.

\bibitem{Guo2017}
C.~Guo, G.~Pleiss, Y.~Sun, and K.~Q. Weinberger, ``On calibration of modern
  neural network,'' in \emph{{Proceedings of the 34th International Conference
  on Machine Learning - Volume 70}}, ser. ICML’17, 2017, pp. 1321--1330.

\bibitem{Nixon_2019_CVPR_Workshops}
J.~Nixon, M.~W. Dusenberry, L.~Zhang, G.~Jerfel, and D.~Tran, ``Measuring
  calibration in deep learning,'' in \emph{The IEEE Conference on Computer
  Vision and Pattern Recognition (CVPR) Workshops}, June 2019.

\bibitem{DBLP:conf/iclr/HendrycksG17}
D.~Hendrycks and K.~Gimpel, ``A baseline for detecting misclassified and
  out-of-distribution examples in neural networks,'' in \emph{5th International
  Conference on Learning Representations, {ICLR} 2017, Conference Track
  Proceedings}, 2017.

\bibitem{Zhou:2012}
Z.-H. Zhou, \emph{Ensemble Methods: Foundations and Algorithms}, 1st~ed.\hskip
  1em plus 0.5em minus 0.4em\relax Chapman and Hall/CRC, 2012.

\bibitem{GALTON1907}
F.~Galton, ``{Vox Populi},'' \emph{Nature}, vol.~75, no. 1949, pp. 450--451,
  Mar 1907.

\bibitem{Breiman1996}
L.~Breiman, ``Bagging predictors,'' \emph{Machine Learning}, vol.~24, no.~2,
  pp. 123--140, Aug 1996.

\bibitem{doi:10.1162/neco.1992.4.3.448}
D.~J.~C. MacKay, ``A practical {B}ayesian framework for backpropagation
  networks,'' \emph{Neural Computation}, vol.~4, no.~3, pp. 448--472, 1992.

\bibitem{izmailov2021bayesian}
\BIBentryALTinterwordspacing
P.~Izmailov, S.~Vikram, M.~D. Hoffman, and A.~G. Wilson, ``What are {B}ayesian
  neural network posteriors really like?'' \emph{CoRR}, vol. abs/2104.14421,
  2021. [Online]. Available: \url{http://arxiv.org/abs/2104.14421}
\BIBentrySTDinterwordspacing

\bibitem{gal2015bayesian}
Y.~Gal and Z.~Ghahramani, ``Bayesian convolutional neural networks with
  {B}ernoulli approximate variational inference,'' in \emph{4th International
  Conference on Learning Representations (ICLR) workshop track}, 2016.

\bibitem{kingma2014stochastic}
D.~P. Kingma and M.~Welling, ``Stochastic gradient vb and the variational
  auto-encoder,'' in \emph{Second International Conference on Learning
  Representations, ICLR}, vol.~19, 2014.

\bibitem{robert2007bayesian}
C.~Robert, \emph{The Bayesian choice: from decision-theoretic foundations to
  computational implementation}.\hskip 1em plus 0.5em minus 0.4em\relax
  Springer Science \& Business Media, 2007.

\bibitem{MitrosDL2019}
J.~Mitros and B.~M. Namee, ``On the validity of {B}ayesian neural networks for
  uncertainty estimation,'' in \emph{AICS}, 2019.

\bibitem{kristiadi2020bayesian}
\BIBentryALTinterwordspacing
A.~Kristiadi, M.~Hein, and P.~Hennig, ``Being {B}ayesian, even just a bit,
  fixes overconfidence in {ReLU} networks,'' \emph{CoRR}, vol. abs/2002.10118,
  2020. [Online]. Available: \url{http://arxiv.org/abs/2002.10118}
\BIBentrySTDinterwordspacing

\bibitem{NIPS2019_9547}
Y.~Ovadia, E.~Fertig, J.~Ren, Z.~Nado, D.~Sculley, S.~Nowozin, J.~Dillon,
  B.~Lakshminarayanan, and J.~Snoek, ``Can you trust your model\textquotesingle
  s uncertainty? evaluating predictive uncertainty under dataset shift,'' in
  \emph{Advances in Neural Information Processing Systems 32}.\hskip 1em plus
  0.5em minus 0.4em\relax Curran Associates, Inc., 2019, pp. 13\,991--14\,002.

\bibitem{KIUREGHIAN2009105}
A.~D. Kiureghian and O.~Ditlevsen, ``Aleatory or epistemic? does it matter?''
  \emph{Structural Safety}, vol.~31, no.~2, pp. 105--112, 2009, risk Acceptance
  and Risk Communication.

\bibitem{depeweg2017decomposition}
S.~Depeweg, J.-M. Hernandez-Lobato, F.~Doshi-Velez, and S.~Udluft,
  ``Decomposition of uncertainty in {B}ayesian deep learning for efficient and
  risk-sensitive learning,'' in \emph{Proceedings of the 35th International
  Conference on Machine Learning}, ser. Proceedings of Machine Learning
  Research, vol.~80, 2018, pp. 1184--1193.

\bibitem{nofreelunch}
D.~H. {Wolpert}, ``The lack of a priori distinctions between learning
  algorithms,'' \emph{Neural Computation}, vol.~8, no.~7, pp. 1341--1390, 1996.

\bibitem{Kendall2017}
A.~Kendall and Y.~Gal, ``What uncertainties do we need in {B}ayesian deep
  learning for computer vision?'' in \emph{Proceedings of the 31st
  International Conference on Neural Information Processing Systems}, ser.
  NIPS’17, 2017, p. 5580–5590.

\bibitem{4049810}
T.~{Auld}, A.~W. {Moore}, and S.~F. {Gull}, ``{B}ayesian neural networks for
  internet traffic classification,'' \emph{IEEE Transactions on Neural
  Networks}, vol.~18, no.~1, pp. 223--239, 2007.

\bibitem{ZHANG2020113246}
X.~Zhang and S.~Mahadevan, ``{B}ayesian neural networks for flight trajectory
  prediction and safety assessment,'' \emph{Decision Support Systems}, vol.
  131, p. 113246, 2020.

\bibitem{doi:10.1080/15732479.2014.951867}
S.~Arangio and F.~Bontempi, ``Structural health monitoring of a cable--stayed
  bridge with {B}ayesian neural networks,'' \emph{Structure and Infrastructure
  Engineering}, vol.~11, no.~4, pp. 575--587, 2015.

\bibitem{BATENI2007102}
S.~M. Bateni, D.-S. Jeng, and B.~W. Melville, ``Bayesian neural networks for
  prediction of equilibrium and time-dependent scour depth around bridge
  piers,'' \emph{Advances in Engineering Software}, vol.~38, no.~2, pp.
  102--111, 2007.

\bibitem{hydrologyBNN}
X.~Zhang, F.~Liang, R.~Srinivasan, and M.~Van~Liew, ``Estimating uncertainty of
  streamflow simulation using bayesian neural networks,'' \emph{Water Resources
  Research}, vol.~45, no.~2, 2009.

\bibitem{Cobb_2019}
A.~D. Cobb, M.~D. Himes, F.~Soboczenski, S.~Zorzan, M.~D. O'Beirne, A.~G.
  Baydin, Y.~Gal, S.~D. Domagal-Goldman, G.~N. Arney, and D.~A. and, ``An
  ensemble of bayesian neural networks for exoplanetary atmospheric
  retrieval,'' \emph{The Astronomical Journal}, vol. 158, no.~1, p.~33, jun
  2019.

\bibitem{Aminian2001}
F.~Aminian and M.~Aminian, ``Fault diagnosis of analog circuits using
  {B}ayesian neural networks with wavelet transform as preprocessor,''
  \emph{Journal of Electronic Testing}, vol.~17, no.~1, pp. 29--36, Feb 2001.

\bibitem{Beker2020}
W.~Beker, A.~Wo{\l}os, S.~Szymku{\'{c}}, and B.~A. Grzybowski,
  ``Minimal--uncertainty prediction of general drug--likeness based on
  {B}ayesian neural networks,'' \emph{Nature Machine Intelligence}, vol.~2,
  no.~8, pp. 457--465, Aug 2020.

\bibitem{Gal2017}
Y.~Gal, R.~Islam, and Z.~Ghahramani, ``Deep {B}ayesian active learning with
  image data,'' in \emph{Proceedings of the 34th International Conference on
  Machine Learning - Volume 70}, ser. ICML’17, 2017, p. 1183–1192.

\bibitem{tran2019bayesian}
\BIBentryALTinterwordspacing
T.~Tran, T.-T. Do, I.~Reid, and G.~Carneiro, ``Bayesian generative active deep
  learning,'' \emph{CoRR}, vol. abs/1904.11643, 2019. [Online]. Available:
  \url{http://arxiv.org/abs/1904.11643}
\BIBentrySTDinterwordspacing

\bibitem{opper1998bayesian}
M.~Opper and O.~Winther, ``A {B}ayesian approach to on-line learning,''
  \emph{{On-line learning in neural networks}}, pp. 363--378, 1998.

\bibitem{10.5555/3327144.3327290}
H.~Ritter, A.~Botev, and D.~Barber, ``Online structured {L}aplace
  approximations for overcoming catastrophic forgetting,'' in \emph{Proceedings
  of the 32nd International Conference on Neural Information Processing
  Systems}, ser. NIPS’18, 2018, pp. 3742--3752.

\bibitem{10.1145/3234150}
S.~Pouyanfar, S.~Sadiq, Y.~Yan, H.~Tian, Y.~Tao, M.~P. Reyes, M.-L. Shyu, S.-C.
  Chen, and S.~S. Iyengar, ``A survey on deep learning: Algorithms, techniques,
  and applications,'' \emph{ACM Comput. Surv.}, vol.~51, no.~5, Sep. 2018.

\bibitem{Buntine_1994}
W.~L. Buntine, ``Operations for learning with graphical models,'' \emph{Journal
  of Artificial Intelligence Research}, vol.~2, pp. 159--225, Dec 1994.

\bibitem{wen2018flipout}
Y.~Wen, P.~Vicol, J.~Ba, D.~Tran, and R.~Grosse, ``Flipout: Efficient
  pseudo-independent weight perturbations on mini-batches,'' in
  \emph{International Conference on Learning Representations}, 2018.

\bibitem{ZhangBHRV17}
C.~Zhang, S.~Bengio, M.~Hardt, B.~Recht, and O.~Vinyals, ``Understanding deep
  learning requires rethinking generalization,'' in \emph{5th International
  Conference on Learning Representations, {ICLR}}, 2017.

\bibitem{carpenter2017stan}
B.~Carpenter, A.~Gelman, M.~D. Hoffman, D.~Lee, B.~Goodrich, M.~Betancourt,
  M.~Brubaker, J.~Guo, P.~Li, and A.~Riddell, ``Stan: A probabilistic
  programming language,'' \emph{Journal of statistical software}, vol.~76,
  no.~1, 2017.

\bibitem{StanPriors}
A.~Gelman and {other Stan developers}, ``Prior choice recommendations,'' 2020,
  retrieved from
  https://github.com/stan-dev/stan/wiki/Prior-Choice-Recommendations [last seen
  13.07.2020].

\bibitem{silvestro2020prior}
\BIBentryALTinterwordspacing
D.~Silvestro and T.~Andermann, ``Prior choice affects ability of {B}ayesian
  neural networks to identify unknowns,'' \emph{CoRR}, vol. abs/2005.04987,
  2020. [Online]. Available: \url{http://arxiv.org/abs/2005.04987}
\BIBentrySTDinterwordspacing

\bibitem{10.5555/2380985}
K.~P. Murphy, \emph{{Machine Learning: A Probabilistic Perspective}}.\hskip 1em
  plus 0.5em minus 0.4em\relax The MIT Press, 2012.

\bibitem{pourzanjani2017bayesian}
\BIBentryALTinterwordspacing
A.~A. Pourzanjani, R.~M. Jiang, B.~Mitchell, P.~J. Atzberger, and L.~R.
  Petzold, ``Bayesian inference over the {S}tiefel manifold via the {G}ivens
  representation,'' \emph{CoRR}, vol. abs/1710.09443, 2017. [Online].
  Available: \url{http://arxiv.org/abs/1710.09443}
\BIBentrySTDinterwordspacing

\bibitem{ba2016layer}
J.~L. Ba, J.~R. Kiros, and G.~E. Hinton, ``Layer normalization,'' \emph{CoRR},
  vol. arXiv:1607.06450, 2016, in NIPS 2016 Deep Learning Symposium.

\bibitem{qi2019small}
\BIBentryALTinterwordspacing
G.-J. Qi and J.~Luo, ``Small data challenges in big data era: A survey of
  recent progress on unsupervised and semi-supervised methods,'' \emph{CoRR},
  vol. abs/1903.11260, 2019. [Online]. Available:
  \url{http://arxiv.org/abs/1903.11260}
\BIBentrySTDinterwordspacing

\bibitem{NIPS2013_5073}
N.~Natarajan, I.~S. Dhillon, P.~K. Ravikumar, and A.~Tewari, ``Learning with
  noisy labels,'' in \emph{Advances in Neural Information Processing Systems
  26}.\hskip 1em plus 0.5em minus 0.4em\relax Curran Associates, Inc., 2013,
  pp. 1196--1204.

\bibitem{6685834}
B.~{Frenay} and M.~{Verleysen}, ``Classification in the presence of label
  noise: A survey,'' \emph{{IEEE} Transactions on Neural Networks and Learning
  Systems}, vol.~25, no.~5, pp. 845--869, 2014.

\bibitem{Tommi03oninformation}
A.~C. Tommi and T.~Jaakkola, ``On information regularization,'' in \emph{In
  Proceedings of the 19th UAI}, 2003.

\bibitem{sohn2020fixmatch}
\BIBentryALTinterwordspacing
K.~Sohn, D.~Berthelot, C.-L. Li, Z.~Zhang, N.~Carlini, E.~D. Cubuk, A.~Kurakin,
  H.~Zhang, and C.~Raffel, ``Fix{M}atch: Simplifying semi-supervised learning
  with consistency and confidence,'' \emph{CoRR}, vol. abs/2001.07685, 2020.
  [Online]. Available: \url{https://arxiv.org/abs/2001.07685}
\BIBentrySTDinterwordspacing

\bibitem{10.5555/1248547.1248632}
M.~Belkin, P.~Niyogi, and V.~Sindhwani, ``Manifold regularization: A geometric
  framework for learning from labeled and unlabeled examples,'' \emph{J. Mach.
  Learn. Res.}, vol.~7, pp. 2399--2434, Dec. 2006.

\bibitem{JMLR:v12:yu11a}
S.~Yu, B.~Krishnapuram, R.~Rosales, and R.~B. Rao, ``Bayesian co-training,''
  \emph{Journal of Machine Learning Research}, vol.~12, no.~80, pp. 2649--2680,
  2011.

\bibitem{6977247}
R.~{Kunwar}, U.~{Pal}, and M.~{Blumenstein}, ``Semi-supervised online
  {B}ayesian network learner for handwritten characters recognition,'' in
  \emph{2014 22nd International Conference on Pattern Recognition}, 2014, pp.
  3104--3109.

\bibitem{lee2013pseudo}
D.-H. Lee, ``Pseudo-label: The simple and efficient semi-supervised learning
  method for deep neural networks,'' in \emph{Workshop on challenges in
  representation learning, {ICML}}, vol.~3, 2013.

\bibitem{Li2019}
Z.~Li, B.~Ko, and H.-J. Choi, ``Naive semi-supervised deep learning using
  pseudo-label,'' \emph{Peer-to-Peer Networking and Applications}, vol.~12,
  no.~5, pp. 1358--1368, 2019.

\bibitem{bari2020multimix}
M.~S. Bari, M.~T. Mohiuddin, and S.~Joty, ``{MultiMix}: A robust data
  augmentation strategy for cross-lingual nlp,'' in \emph{ICML}, 2020.

\bibitem{NIPS2000_1876}
O.~Chapelle, J.~Weston, L.~Bottou, and V.~Vapnik, ``Vicinal risk
  minimization,'' in \emph{Advances in Neural Information Processing Systems
  13}.\hskip 1em plus 0.5em minus 0.4em\relax MIT Press, 2001, pp. 416--422.

\bibitem{xieArxiv19}
\BIBentryALTinterwordspacing
Q.~Xie, Z.~Dai, E.~H. Hovy, M.~Luong, and Q.~V. Le, ``Unsupervised data
  augmentation,'' \emph{CoRR}, vol. abs/1904.12848, 2019. [Online]. Available:
  \url{http://arxiv.org/abs/1904.12848}
\BIBentrySTDinterwordspacing

\bibitem{hospedales2020metalearning}
\BIBentryALTinterwordspacing
T.~Hospedales, A.~Antoniou, P.~Micaelli, and A.~Storkey, ``Meta-learning in
  neural networks: A survey,'' \emph{CoRR}, vol. abs/2004.05439, 2020.
  [Online]. Available: \url{http://arxiv.org/abs/2004.05439}
\BIBentrySTDinterwordspacing

\bibitem{5288526}
S.~J. {Pan} and Q.~{Yang}, ``A survey on transfer learning,'' \emph{IEEE
  Transactions on Knowledge and Data Engineering}, vol.~22, no.~10, pp.
  1345--1359, 2010.

\bibitem{qiu2020pretrained}
X.~Qiu, T.~Sun, Y.~Xu, Y.~Shao, N.~Dai, and X.~Huang, ``Pre-trained models for
  natural language processing: A survey,'' \emph{CoRR}, vol. abs/2003.08271,
  2020.

\bibitem{9086055}
L.~{Jing} and Y.~{Tian}, ``Self-supervised visual feature learning with deep
  neural networks: A survey,'' \emph{IEEE Transactions on Pattern Analysis and
  Machine Intelligence}, pp. 1--1, 2020.

\bibitem{han2021image}
X.-F. Han, H.~Laga, and M.~Bennamoun, ``Image-based 3d object reconstruction:
  State-of-the-art and trends in the deep learning era,'' \emph{IEEE
  transactions on pattern analysis and machine intelligence}, vol.~43, no.~5,
  pp. 1578--1604, 2021.

\bibitem{laga2020survey}
H.~Laga, L.~V. Jospin, F.~Boussaid, and M.~Bennamoun, ``A survey on deep
  learning techniques for stereo-based depth estimation,'' \emph{IEEE
  Transactions on Pattern Analysis and Machine Intelligence}, 2020.

\bibitem{DBLP:conf/iclr/GrantFLDG18}
E.~Grant, C.~Finn, S.~Levine, T.~Darrell, and T.~L. Griffiths, ``Recasting
  gradient-based meta-learning as hierarchical {B}ayes,'' in \emph{6th
  International Conference on Learning Representations, {ICLR} 2018, Vancouver,
  BC, Canada, April 30 - May 3, 2018, Conference Track Proceedings}, 2018.

\bibitem{9010283}
L.~{Beyer}, X.~{Zhai}, A.~{Oliver}, and A.~{Kolesnikov}, ``{S4L}:
  Self-supervised semi-supervised learning,'' in \emph{2019 IEEE/CVF
  International Conference on Computer Vision (ICCV)}, 2019, pp. 1476--1485.

\bibitem{10.1093/biomet/57.1.97}
W.~K. Hastings, ``{Monte Carlo sampling methods using Markov chains and their
  applications},'' \emph{Biometrika}, vol.~57, no.~1, pp. 97--109, 04 1970.

\bibitem{BleiStat2018}
D.~M. Blei, A.~Kucukelbir, and J.~D. McAuliffe, ``Variational inference: A
  review for statisticians,'' \emph{Journal of the American Statistical
  Association}, vol. 112, no. 518, pp. 859--877, 2017.

\bibitem{Bardenet2017}
R.~Bardenet, A.~Doucet, and C.~Holmes, ``On {Markov Chain Monte Carlo} methods
  for tall data,'' \emph{{J. Mach. Learn. Res.}}, vol.~18, no.~1, pp.
  1515--1557, Jan. 2017.

\bibitem{George92explainingthe}
E.~I. George, G.~Casella, and E.~I. George, ``Explaining the {G}ibbs sampler,''
  \emph{The American Statistician}, 1992.

\bibitem{doi:10.1080/00031305.1995.10476177}
S.~Chib and E.~Greenberg, ``Understanding the {M}etropolis-{H}astings
  algorithm,'' \emph{The American Statistician}, vol.~49, no.~4, pp. 327--335,
  1995.

\bibitem{neal2011mcmc}
R.~M. Neal \emph{et~al.}, ``{MCMC} using {H}amiltonian dynamics,''
  \emph{Handbook of Markov Chain Monte Carlo}, vol.~2, no.~11, p.~2, 2011.

\bibitem{hoffman2014no}
M.~D. Hoffman and A.~Gelman, ``The {No-U-Turn} sampler: adaptively setting path
  lengths in {Hamiltonian Monte Carlo},'' \emph{Journal of Machine Learning
  Research}, vol.~15, no.~1, pp. 1593--1623, 2014.

\bibitem{10.1214/aoms/1177729694}
S.~Kullback and R.~A. Leibler, ``On information and sufficiency,'' \emph{The
  Annals of Mathematical Statistics}, vol.~22, no.~1, pp. 79 -- 86, 1951.

\bibitem{6773024}
C.~E. Shannon, ``A mathematical theory of communication,'' \emph{The Bell
  System Technical Journal}, vol.~27, no.~3, pp. 379--423, 1948.

\bibitem{Hoffman2013}
M.~D. Hoffman, D.~M. Blei, C.~Wang, and J.~Paisley, ``Stochastic variational
  inference,'' \emph{J. Mach. Learn. Res.}, vol.~14, no.~1, pp. 1303--1347, May
  2013.

\bibitem{GravesNIPS2011}
A.~Graves, ``Practical variational inference for neural networks,'' in
  \emph{Advances in Neural Information Processing Systems 24}.\hskip 1em plus
  0.5em minus 0.4em\relax Curran Associates, Inc., 2011, pp. 2348--2356.

\bibitem{ghahramani00propagation}
Z.~Ghahramani and M.~J. Beal, ``Propagation algorithms for variational
  {B}ayesian learning,'' in \emph{Advances in Neural Information Processing
  Systems 13}.\hskip 1em plus 0.5em minus 0.4em\relax MIT Press, 2001, pp.
  507--513.

\bibitem{ritter2018a}
H.~Ritter, A.~Botev, and D.~Barber, ``A scalable laplace approximation for
  neural networks,'' in \emph{International Conference on Learning
  Representations}, 2018.

\bibitem{MaddoxDL2019}
W.~J. Maddox, P.~Izmailov, T.~Garipov, D.~P. Vetrov, and A.~G. Wilson, ``A
  simple baseline for {B}ayesian uncertainty in deep learning,'' in
  \emph{Advances in Neural Information Processing Systems 32}.\hskip 1em plus
  0.5em minus 0.4em\relax Curran Associates, Inc., 2019, pp. 13\,153--13\,164.

\bibitem{hernndezlobato2015probabilistic}
J.~M. Hern\'{a}ndez-Lobato and R.~P. Adams, ``Probabilistic backpropagation for
  scalable learning of {B}ayesian neural networks,'' in \emph{Proceedings of
  the 32nd International Conference on International Conference on Machine
  Learning - Volume 37}, ser. ICML’15, 2015, p. 1861–1869.

\bibitem{pmlr-v37-blundell15}
C.~Blundell, J.~Cornebise, K.~Kavukcuoglu, and D.~Wierstra, ``Weight
  uncertainty in neural network,'' in \emph{Proceedings of the 32nd
  International Conference on Machine Learning}, ser. Proceedings of Machine
  Learning Research, vol.~37, 2015, pp. 1613--1622.

\bibitem{kingma2019introduction}
D.~P. Kingma, M.~Welling \emph{et~al.}, ``An introduction to variational
  autoencoders,'' \emph{{Foundations and Trends{\textregistered} in Machine
  Learning}}, vol.~12, no.~4, pp. 307--392, 2019.

\bibitem{kingma2014adam}
D.~Kingma and J.~Ba, ``Adam: A method for stochastic optimization,''
  \emph{{International Conference on Learning Representations}}, 12 2014.

\bibitem{JMLR:v15:srivastava14a}
N.~Srivastava, G.~Hinton, A.~Krizhevsky, I.~Sutskever, and R.~Salakhutdinov,
  ``Dropout: A simple way to prevent neural networks from overfitting,''
  \emph{Journal of Machine Learning Research}, vol.~15, no.~56, pp. 1929--1958,
  2014.

\bibitem{GalDL2015}
Y.~Gal and Z.~Ghahramani, ``Dropout as a {B}ayesian approximation: Representing
  model uncertainty in deep learning,'' in \emph{Proceedings of the 33rd
  International Conference on Machine Learning - Volume 48}, ser. ICML’16,
  2016, p. 1050–1059.

\bibitem{Yingzhen2017}
Y.~Li and Y.~Gal, ``Dropout inference in {B}ayesian neural networks with
  alpha-divergences,'' in \emph{Proceedings of the 34th International
  Conference on Machine Learning - Volume 70}, ser. ICML’17, 2017, pp.
  2052--2061.

\bibitem{pmlr-v80-hron18a}
J.~Hron, A.~Matthews, and Z.~Ghahramani, ``Variational {B}ayesian dropout:
  pitfalls and fixes,'' in \emph{Proceedings of the 35th International
  Conference on Machine Learning}, ser. Proceedings of Machine Learning
  Research, vol.~80, 2018, pp. 2019--2028.

\bibitem{pmlr-v119-chan20a}
A.~Chan, A.~Alaa, Z.~Qian, and M.~Van Der~Schaar, ``Unlabelled data improves
  {B}ayesian uncertainty calibration under covariate shift,'' in
  \emph{Proceedings of the 37th International Conference on Machine Learning},
  ser. Proceedings of Machine Learning Research, vol. 119.\hskip 1em plus 0.5em
  minus 0.4em\relax Virtual: PMLR, 13--18 Jul 2020, pp. 1392--1402.

\bibitem{mandt2017stochastic}
S.~Mandt, M.~D. Hoffman, and D.~M. Blei, ``Stochastic gradient descent as
  approximate {B}ayesian inference,'' \emph{The Journal of Machine Learning
  Research}, vol.~18, no.~1, pp. 4873--4907, 2017.

\bibitem{welling2011bayesian}
M.~Welling and Y.~W. Teh, ``Bayesian learning via stochastic gradient
  {L}angevin dynamics,'' in \emph{Proceedings of the 28th international
  conference on machine learning}, ser. ICML ’11, 2011, pp. 681--688.

\bibitem{seedat2019calibrated}
\BIBentryALTinterwordspacing
N.~Seedat and C.~Kanan, ``Towards calibrated and scalable uncertainty
  representations for neural networks,'' \emph{CoRR}, vol. abs/1911.00104,
  2019. [Online]. Available: \url{http://arxiv.org/abs/1911.00104}
\BIBentrySTDinterwordspacing

\bibitem{NIPS2017_7219}
B.~Lakshminarayanan, A.~Pritzel, and C.~Blundell, ``Simple and scalable
  predictive uncertainty estimation using deep ensembles,'' in \emph{Advances
  in Neural Information Processing Systems 30}.\hskip 1em plus 0.5em minus
  0.4em\relax Curran Associates, Inc., 2017, pp. 6402--6413.

\bibitem{Khan2018FastAS}
M.~Khan, D.~Nielsen, V.~Tangkaratt, W.~Lin, Y.~Gal, and A.~Srivastava, ``Fast
  and scalable {B}ayesian deep learning by weight-perturbation in {A}dam,'' in
  \emph{Proceedings of the 35th International Conference on Machine Learning},
  ser. Proceedings of Machine Learning Research, vol.~80, 2018, pp. 2611--2620.

\bibitem{pearce2018uncertainty}
T.~Pearce, F.~Leibfried, A.~Brintrup, M.~Zaki, and A.~Neely, ``Uncertainty in
  neural networks: Approximately {B}ayesian ensembling,'' in \emph{AISTATS
  2020}, 2020.

\bibitem{zeng2018relevance}
\BIBentryALTinterwordspacing
J.~Zeng, A.~Lesnikowski, and J.~M. Alvarez, ``The relevance of {B}ayesian layer
  positioning to model uncertainty in deep {B}ayesian active learning,''
  \emph{CoRR}, vol. abs/1811.12535, 2018. [Online]. Available:
  \url{http://arxiv.org/abs/1811.12535}
\BIBentrySTDinterwordspacing

\bibitem{brosse2020lastlayer}
\BIBentryALTinterwordspacing
N.~Brosse, C.~Riquelme, A.~Martin, S.~Gelly, and Éric Moulines, ``On
  last-layer algorithms for classification: Decoupling representation from
  uncertainty estimation,'' \emph{CoRR}, vol. abs/2001.08049, 2020. [Online].
  Available: \url{http://arxiv.org/abs/2001.08049}
\BIBentrySTDinterwordspacing

\bibitem{10.1145/1102351.1102457}
E.~Snelson and Z.~Ghahramani, ``Compact approximations to {B}ayesian predictive
  distributions,'' in \emph{Proceedings of the 22nd International Conference on
  Machine Learning}, ser. ICML ’05, 2005, p. 840–847.

\bibitem{KorattikaraDL2015}
A.~Korattikara, V.~Rathod, K.~Murphy, and M.~Welling, ``Bayesian dark
  knowledge,'' in \emph{Proceedings of the 28th International Conference on
  Neural Information Processing Systems - Volume 2}, ser. NIPS’15, 2015, pp.
  3438--3446.

\bibitem{hinton2015distilling}
G.~Hinton, O.~Vinyals, and J.~Dean, ``Distilling the knowledge in a neural
  network,'' \emph{arXiv preprint arXiv:1503.02531}, 2015, in {NIPS 2014 Deep
  Learning Workshop}.

\bibitem{menonArxiv20}
\BIBentryALTinterwordspacing
A.~K. Menon, A.~S. Rawat, S.~J. Reddi, S.~Kim, and S.~Kumar, ``Why distillation
  helps: a statistical perspective,'' \emph{CoRR}, vol. abs/2005.10419, 2020.
  [Online]. Available: \url{https://arxiv.org/abs/2005.10419}
\BIBentrySTDinterwordspacing

\bibitem{Wang2018AdversarialDO}
K.-C. Wang, P.~Vicol, J.~Lucas, L.~Gu, R.~Grosse, and R.~Zemel, ``Adversarial
  distillation of {B}ayesian neural network posteriors,'' in \emph{Proceedings
  of the 35th International Conference on Machine Learning}, ser. Proceedings
  of Machine Learning Research, vol.~80, 2018, pp. 5190--5199.

\bibitem{Janocha2017}
\BIBentryALTinterwordspacing
K.~Janocha and W.~M. Czarnecki, ``On loss functions for deep neural networks in
  classification,'' \emph{Schedae Informaticae}, vol. 1/2016, 2017. [Online].
  Available: \url{http://dx.doi.org/10.4467/20838476SI.16.004.6185}
\BIBentrySTDinterwordspacing

\bibitem{kuleshov2018accurate}
V.~Kuleshov, N.~Fenner, and S.~Ermon, ``Accurate uncertainties for deep
  learning using calibrated regression,'' in \emph{Proceedings of the 35th
  International Conference on Machine Learning}, ser. Proceedings of Machine
  Learning Research, vol.~80, 2018, pp. 2796--2804.

\end{thebibliography}


\begin{thebibliography}{1}
\providecommand{\url}[1]{#1}
\csname url@samestyle\endcsname
\providecommand{\newblock}{\relax}
\providecommand{\bibinfo}[2]{#2}
\providecommand{\BIBentrySTDinterwordspacing}{\spaceskip=0pt\relax}
\providecommand{\BIBentryALTinterwordstretchfactor}{4}
\providecommand{\BIBentryALTinterwordspacing}{\spaceskip=\fontdimen2\font plus
\BIBentryALTinterwordstretchfactor\fontdimen3\font minus
  \fontdimen4\font\relax}
\providecommand{\BIBforeignlanguage}[2]{{%
\expandafter\ifx\csname l@#1\endcsname\relax
\typeout{** WARNING: IEEEtran.bst: No hyphenation pattern has been}%
\typeout{** loaded for the language `#1'. Using the pattern for}%
\typeout{** the default language instead.}%
\else
\language=\csname l@#1\endcsname
\fi
#2}}
\providecommand{\BIBdecl}{\relax}
\BIBdecl

\bibitem{726791}
Y.~Lecun, L.~Bottou, Y.~Bengio, and P.~Haffner, ``Gradient-based learning
  applied to document recognition,'' \emph{Proceedings of the IEEE}, vol.~86,
  no.~11, pp. 2278--2324, 1998.

\bibitem{6248110}
D.~Ciregan, U.~Meier, and J.~Schmidhuber, ``Multi-column deep neural networks
  for image classification,'' in \emph{2012 IEEE Conference on Computer Vision
  and Pattern Recognition}, 2012, pp. 3642--3649.

\bibitem{10.5555/1248547.1248632}
M.~Belkin, P.~Niyogi, and V.~Sindhwani, ``Manifold regularization: A geometric
  framework for learning from labeled and unlabeled examples,'' \emph{J. Mach.
  Learn. Res.}, vol.~7, pp. 2399--2434, Dec. 2006.

\bibitem{hoffman2014no}
M.~D. Hoffman and A.~Gelman, ``The {No-U-Turn} sampler: adaptively setting path
  lengths in {Hamiltonian Monte Carlo},'' \emph{Journal of Machine Learning
  Research}, vol.~15, no.~1, pp. 1593--1623, 2014.

\end{thebibliography}
}


\appendices

\section{Implementing parameter-efficient normal distributions for variational inference}
\label{app:matrSimpl}

Given a random vector $\vec{\varepsilon}$ in which each independent and identically distributed component follows a standard normal distribution $\mathcal{N}(0,1)$, one can obtain a sample $\vec{\theta}$ from a normal distribution $\mathcal{N}(\vec{\mu},\matr{\Sigma})$ using the following formulas:

\begin{equation}
    \vec{\theta} = \sqrt{\matr{\Sigma}} \vec{\varepsilon} + \vec{\mu},
\end{equation}

\noindent where $\sqrt{\matr{\Sigma}}$ is a matrix such that $\sqrt{\matr{\Sigma}} \sqrt{\matr{\Sigma}}^{\top} = \matr{\Sigma}$. When a variational inference algorithm needs to learn the covariance matrix of $\vec{\theta}$, it is often more convenient to learn $\sqrt{\matr{\Sigma}}$. Assuming $\vec{\theta}$ has $n$ entries, $\sqrt{\matr{\Sigma}} \sqrt{\matr{\Sigma}}^{\top}$ is the Cholesky decomposition of $\matr{\Sigma}$. As such, $\sqrt{\matr{\Sigma}}$ is a lower triangular matrix and $\mathcal{O}(n^2)$ variational parameters are required to learn the exact covariance matrix. This becomes exceedingly computationally expensive rather quickly when $n$ becomes large.

\begin{figure}
    \centering
    \includegraphics{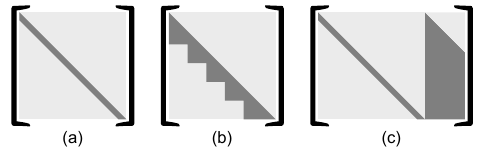}
    \caption{Nonzero entries in $\sqrt{\matr{\Sigma}}$ when learning a diagonal (a), block diagonal (b) or diagonal plus low rank (c) approximation of $\matr{\Sigma}$}
    \label{fig:matrSimplifications}
\end{figure}

A straightforward simplification is to only consider a diagonal approximation of $\matr{\Sigma}$. This can be enforced by learning only the diagonal coefficients of $\sqrt{\matr{\Sigma}}$, meaning only $\mathcal{O}(n)$ variational parameters are required (Figure~\ref{fig:matrSimplifications}a). This approach can be extended to learn more correlation coefficients by learning a block diagonal~\cite{ritter2018a} covariance matrix, which can be done by learning the corresponding lower triangular entries in $\sqrt{\matr{\Sigma}}$ (Figure~\ref{fig:matrSimplifications}b). If the maximal size of the nonzero blocks is fixed to be $w$, $\mathcal{O}(w \cdot n)$ variational parameters are required. The major drawback of this model is that the index of two given parameters determines whether their covariance can be learned by the variational distribution. This is not always ideal, as it is hard to predict which parameters will be the most correlated and need to be positioned close to one another. An alternative is to learn a diagonal plus low rank approximation of $\matr{\Sigma}$ \cite{MaddoxDL2019}. This is done by sampling a vector $\vec{\varepsilon}$ with $n+r$ (instead of $n$) components. $\sqrt{\matr{\Sigma}}$ is then defined as:

\begin{equation}
    \sqrt{\matr{\Sigma}} = \left[ D, ~ L\right],
\end{equation}

\noindent where $D$ is a diagonal matrix of size $n\times n$ and $L$ is a lower triangular matrix of size $n\times r$ (Figure~\ref{fig:matrSimplifications}c). This means that the model has more flexibility to learn the correlation between all the components of $\vec{\theta}$ while only requiring $\mathcal{O}(n \cdot r)$ variational parameters.

\section{A proof of Equation \ref{eq:elbo_trick}}
\label{app:proofProbInt}

Let us assume that we have a probability space $(\Omega, \mathcal{F}, \operatorname{P})$, where $\Omega$ is a set of outcomes, $\mathcal{F}$ is a $\sigma$-algebra of $\Omega$ representing possible events and $\operatorname{P}$ is a measure defined on $\mathcal{F}$ and which assigns a value of $1$ to $\Omega$, representing the probability of an event. In addition, assume that we have a probability distribution $q_{\phi}(\vec{\theta})$ for a given random variable $\vec{\theta}$, a probability distribution $q(\varepsilon)$ for a given random variable $\varepsilon$ and a functional relation $t(\varepsilon, \vec{\phi})$ such that $t(\varepsilon, \vec{\phi})$ is distributed according to $q_{\phi}(\vec{\theta})$ and $t(\varepsilon, \vec{\phi})$ is a bijection with respect to $\varepsilon$. Thus, we have:
\begin{equation}\small
    \operatorname{P}(\vec{\theta}^{-1}(t(E, \vec{\phi}))) = \operatorname{P}(\varepsilon^{-1}(E)) \quad \forall E \in \varepsilon(\mathcal{F}),
\end{equation}

\noindent with 
\[
\begin{array}{c}
     \varepsilon(\mathcal{F}) = \left\{ \{\varepsilon(\omega): \omega \in e\} : e \in \mathcal{F} \right\},  \\
     \\
     t(E, \vec{\phi}) = \{ t(\varepsilon, \vec{\phi}) : \varepsilon \in E \} , \\
     \\
     \varepsilon^{-1}(E) = \bigcup_{e \in \mathcal{F} \land \varepsilon(e) \subseteq E} e , \\
     \\
     \vec{\theta}^{-1}(t(E, \vec{\phi})) = \bigcup_{e \in \mathcal{F} \land \vec{\theta}(e) \subseteq t(E, \vec{\phi})} e.
\end{array}
\]

\noindent Since $t(\varepsilon, \vec{\phi})$ is a bijection with respect to $\varepsilon$, we have $\varepsilon^{-1}(E) = \vec{\theta}^{-1}(t(E, \vec{\phi}))$. This implies:
\begin{equation}\small
    \int_{\vec{\theta} \in t(E, \vec{\phi})} q_{\phi}(\vec{\theta}) d\vec{\theta} = \int_{\varepsilon \in E} q(\varepsilon) d\varepsilon \quad \forall E \in \varepsilon(\mathcal{F}).
\end{equation}

\noindent which in turn implies: 
\begin{equation}\small
    q_{\phi}(\vec{\theta}) d\vec{\theta} = q(\varepsilon) d\varepsilon
\end{equation}

\noindent for non-degenerated probability distributions $q_{\phi}(\vec{\theta})$ and $q(\varepsilon)$.

Now, given a differentiable function $f(\vec{\phi},\vec{\theta})$, we have:
\begin{equation}\small
    \int_{\vec{\theta} \in t(\varepsilon(\Omega), \vec{\phi})} f(\vec{\phi},\vec{\theta}) q_{\phi}(\vec{\theta}) d\vec{\theta} = \int_{\varepsilon \in \varepsilon(\Omega)} f(\vec{\phi},t(\varepsilon, \vec{\phi})) q(\varepsilon) d\varepsilon
\end{equation}

\noindent which implies Equation (\ref{eq:elbo_trick}).

\end{document}


\title{Hands-on Bayesian Neural Networks -- Supplementary material}

\newcommand{\shortauthors}{Jospin et al.}

\maketitle

\section{Practical example -- Bayesian MNIST}

Over the years, MNIST \cite{726791} has become the most renown toy dataset in deep learning. Coding a handwritten digit classifier based on this dataset is now the Hello World of deep neural network programming. The first practical example we introduce for this tutorial is thus just a plain old classifier for MNIST implemented as a BNN. The code is available on github [\url{https://github.com/french-paragon/BayesianMnist}]. The setup is as follows;

\textbf{The purpose} is to show a Bayesian Neural Network (BNN) hello world project.

\textbf{The problem} is to train a BNN to perform hand-written digit recognition.

\textbf{The dataset} we are going to use is MNIST. However, to evaluate how the proposed BNN reacts to unseen data, we will remove one of the classes (we used the digit 5 for this experiment) from the training set so that the network never sees it during training.

\textbf{The stochastic model} is the plain Bayesian regression presented in Section~IV-B of the main paper. We use a normal distribution as a prior for the network parameters $\vec{\theta}$, with a standard deviation of $5$ for the weights and $25$ for the bias as we expect the scale of the bias to have a slightly more variability than the weights. This is because, in a RELU or leaky-RELU network, the weights influence the local change in slope of a layer function while the bias indicates the position of those inflection points.

\begin{figure}[b]
    \centering
    \includegraphics[width=0.48\textwidth]{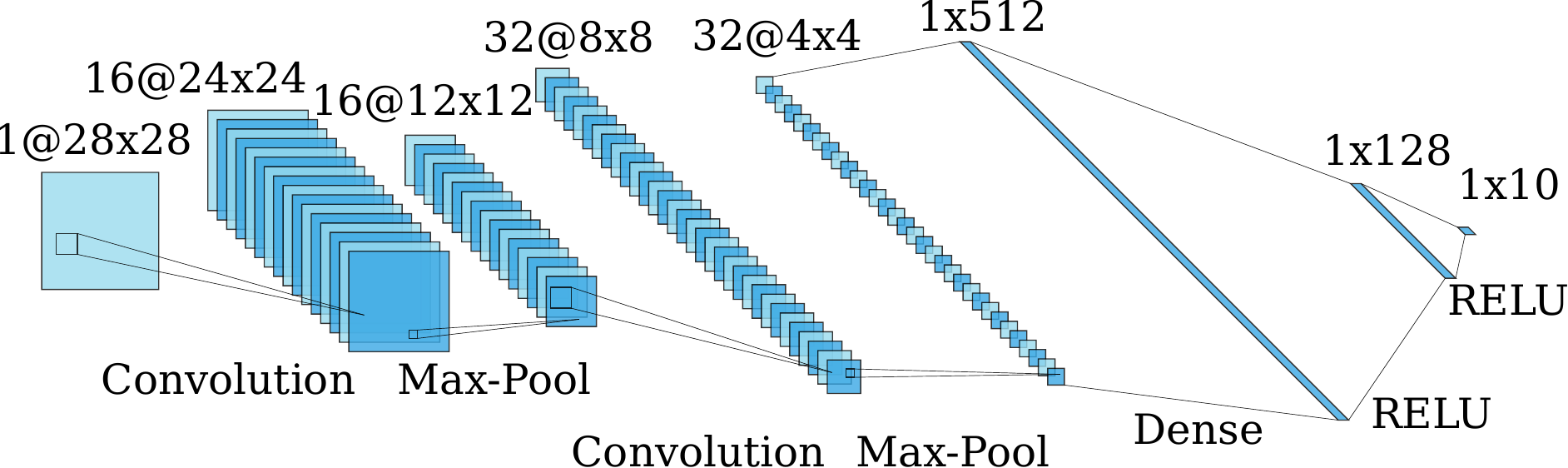}
    \caption{The neural network architecture used for the Bayesian MNIST practical example.}
    \label{fig:bayes_mnist_functional}
\end{figure}

\textbf{The functional model} is a standard convolutional neural network with two convolutional layers followed by two fully connected layers; see Figure~\ref{fig:bayes_mnist_functional}.

\begin{figure*}[t]
    \centering
    \begin{subfigure}[t]{\textwidth}
        \includegraphics[width=0.48\textwidth]{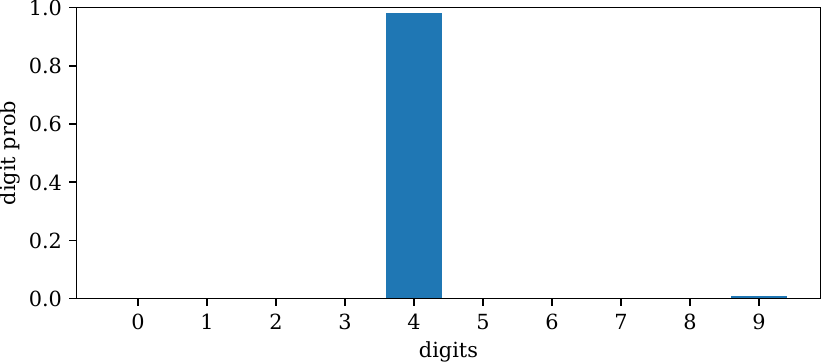}\hfill
        \includegraphics[width=0.48\textwidth]{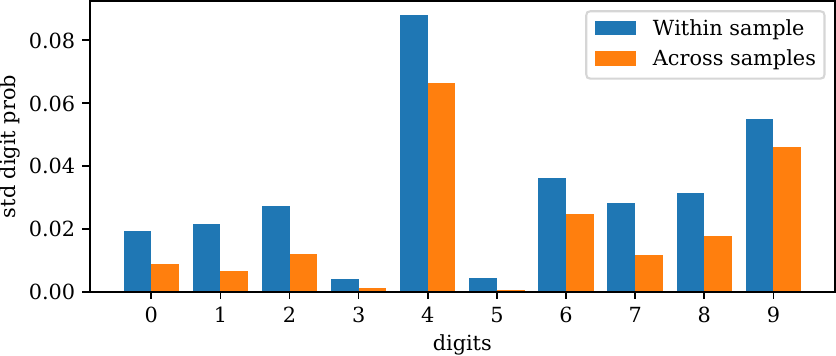}
        \caption{Results on a seen class}
        \label{fig:mnist_results:seen}
    \end{subfigure}
    
    \vspace{10pt}
    
    \begin{subfigure}[t]{\textwidth}
        \includegraphics[width=0.48\textwidth]{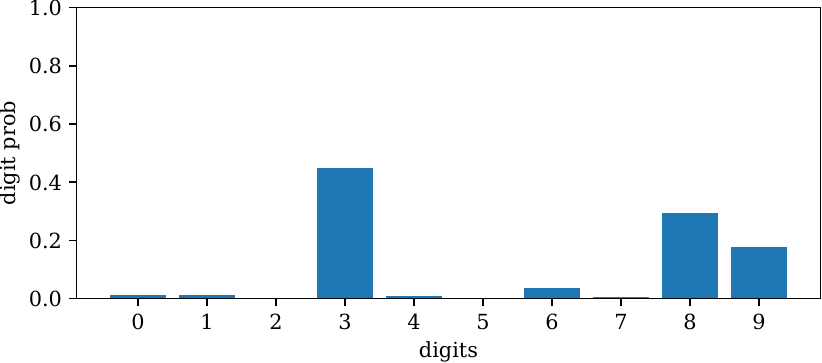}\hfill
        \includegraphics[width=0.48\textwidth]{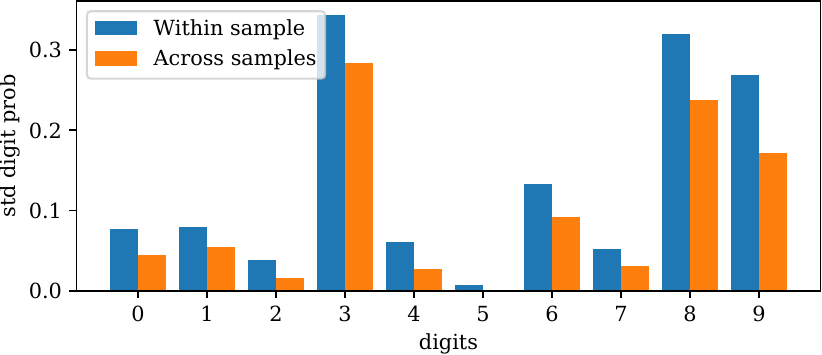}
        \caption{Results on the unseen class}
        \label{fig:mnist_results:unseen}
    \end{subfigure}
    
    \vspace{10pt}
    
    \begin{subfigure}[t]{\textwidth}
        \includegraphics[width=0.48\textwidth]{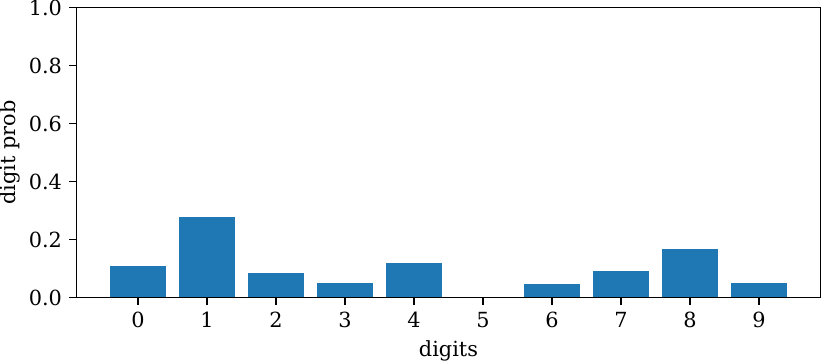}\hfill
        \includegraphics[width=0.48\textwidth]{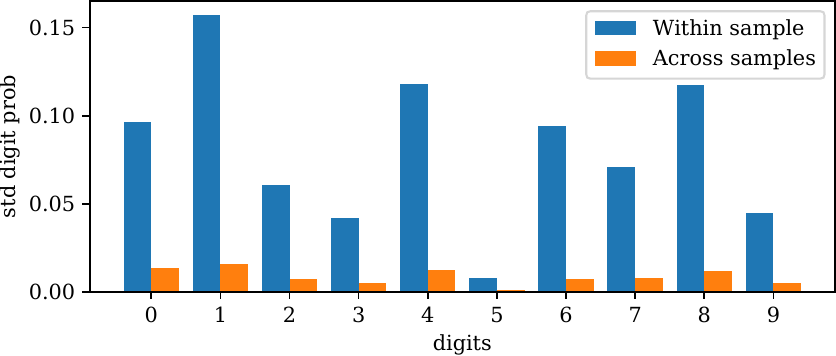}
        \caption{Results on white noise}
        \label{fig:mnist_results:noise}
    \end{subfigure}
    \caption{Average prediction of the Bayesian MNIST practical example for (\subref{fig:mnist_results:seen}) one of the seen class, (\subref{fig:mnist_results:unseen}) the unseen class and (\subref{fig:mnist_results:noise}) white noise input.}
    \label{fig:mnist_results}
\end{figure*}

\subsection{Training}

We used Variational Inference to train this BNN. We use a Gaussian distribution, with diagonal covariance, as a variational posterior (see Section~V-C of the main paper). Since we expect the posterior to be multi-modal, we train an ensemble of BNNs instead of a single one (see Section~V-E2b of the main paper). Note that the top performing method on the MNIST leaderboard is actually an ensemble-based method \cite{6248110}. (The authors of \cite{6248110} designate their model as a multi-column neural network, but the idea is equivalent to an ensemble.)

\subsection{Results}

We tested the final BNN against \textbf{(1)} the test set restricted to the classes the network has been trained on, \textbf{(2)} the test set restricted to the class the network has not been trained on, and \textbf{(3)} pure white noise. For each case, we report:
\begin{itemize}
    \item  The average probabilities predicted by the BNN for all input images in the considered class (Fig.~\ref{fig:mnist_results}), 
    \item The average standard deviation of the network on a single sample, and 
    \item The standard deviation of the prediction for all samples. 
\end{itemize}

\noindent While not as informative and rigorous as a collection of calibration curves, these measures do indicate whether the BNN uncertainty matches the natural variability of the dataset. The reported results show that for a class seen during training, the network gives the correct prediction and is confident about its results. The variability per sample and the variability across samples are coherent with one another and quite small. When presented with a digit from a class unseen during training, the network attempts to match it to digits that share similarities. However, the predicted low probabilities along with the high per-sample and across-sample standard deviation show that the network is aware that it does not know what these unseen digits are. As for the white noise, the average output is constant, meaning the networks is clear about the fact this is not a character.

\section{Practical example -- Sparse measure}
\label{toy_example:sparse_measure}

The second practical example we introduce is entitled ``Sparse measure''. Below, we describe in detail the definition of the stochastic model for this example. We also provide a Python implementation of this example in order to show how the different hypotheses we present in this Appendix translate to actual code [\url{https://github.com/french-paragon/sparse_measure_bnn}].

\textbf{The purpose} is to present a small model illustrating how different training strategies can be integrated in a BNN. 

\textbf{The problem} is to learn a function $f: \mathbb{R}^n \rightarrow \mathbb{R}^m$ based on a dataset $D = \{ (\vec{x}_i, \vec{y}_i) | i \in [1,N],  \vec{x}_i \in \mathbb{R}^n, \vec{y}_i \in \mathbb{R}^m\}$ where for each tuple $(\vec{x}_i, \vec{y}_i)$, only certain elements of $\vec{y}_i$ are actually measured. This scenario corresponds to any application of machine learning where measuring a raw signal, represented by the $\vec{x}_i$'s, is easy but one wants to train an algorithm to reconstruct a derived signal, represented by the $\vec{y}_i$'s, which is much harder to acquire. On top of that, the elements of $\vec{y}_i$ are measured with a certain level of uncertainty. Also, one element of $\vec{y}_i$, which is constant but unknown, has a much higher level of uncertainty (\eg it has been found that one of the instruments used for data acquisition is defective, but no one remembers for which experiment it has been used). It is also known that the function $f$ is continuous and will, with very high probability, map any given input to a point near a known hyperplane of $\mathbb{R}^m$.

\textbf{The dataset} is generated at random for each experiment using python. Inputs are sampled from a zero-mean multivariate normal distribution with a random correlation matrix. The input is then multiplied by a random projection matrix to $\mathbb{R}^3$ before a non-linear function is applied. A random set of three orthonormal vectors in $\mathbb{R}^m$ is then generated and used to reproject the values in the final output space, constrained on a known hyperplane. A final non-linear function is applied on the input and its results are added to the previously generated outputs with a very low gain factor to ensure that the actual output is still likely to be near the known hyperplane. Finally, noise is added to the training data, with one channel receiving much more noise than the other channels. We set $n=m=9$. The measured values for the training set are then selected at random, the other values being set to 0. A set of binary masks are returned with the data to indicate which values have actually been measured.

\textbf{The stochastic model} is the most interesting part of this case study. Note that learning one function $f: \mathbb{R}^n \rightarrow \mathbb{R}^m$ is equivalent to learning $m$ functions $f: \mathbb{R}^n \rightarrow \mathbb{R}$. This means that a standard Bayesian regression would be sufficient to address this learning problem. However, one can do much more since  the functions under consideration can be correlated with one another.

\begin{figure}
    \centering
    \scalebox{1.3}{\begin{tikzpicture}[scale=1,y=1cm] 
    
    \draw[\pgmLineColor,line width=0.5pt] (0.85,1.75) circle (0.25);
    \draw (0.85,1.75) node[anchor=center,align=center]{$\vec{\theta}$};

    \draw[line width=0.5pt,->] (0.68, 1.55) -- (0.2, 0.9);

    \draw[\pgmLineColor,line width=0.5pt] (-1.25,-0.6) rectangle (0.4, 2.1);
    \draw (-1.23,-0.2) node[anchor=south west,align=center]{\footnotesize $D$};
    
    \fill[\pgmBgColor] (0,1.75) circle (0.25);
    \draw[line width=0.75pt] (0,1.75) circle (0.25);
    \draw (0,1.75) node[anchor=center,align=center]{$\vec{x}$};

    \draw[line width=0.5pt,->] (-0.13, 1.55) -- (-0.65, 0.9);
    \draw[line width=0.5pt,->] (-1.6, 1.5) -- (-1.05, 0.9);
    \draw[line width=0.5pt,->] (-0.6, 0.75) -- (-0.25, 0.75);
    \draw[line width=0.5pt,->] (-0.25, 1.75) -- (-1.45, 1.75);
    \draw[line width=0.5pt,->] (0.25, 1.75) -- (0.6, 1.75);
    
    \draw[\pgmLineColor,line width=0.75pt] (-1.95,2) rectangle (-1.45,1.5);
    \draw (-1.7,1.75) node[anchor=center,align=center]{$\vec{\xi}$};
    
    \draw[\pgmLineColor,line width=0.75pt, dashed] (-0.85,0.75) circle (0.25);
    \draw (-0.85,0.75) node[anchor=center,align=center]{$\vec{l}$};

    \draw[\pgmLineColor,line width=0.75pt] (0,0.75) circle (0.25);
    \draw (0,0.75) node[anchor=center,align=center]{$\vec{y}$};
    
    \draw[line width=0.5pt,->] (-1.45, -0.25) -- (-0.25, -0.25);
    
    \draw[\pgmLineColor,line width=0.75pt] (-2.95,1) rectangle (-2.45,0.5);
    \draw (-2.7,0.75) node[anchor=center,align=center]{$\vec{s}$};

    \draw[line width=0.5pt,->] (-2.7, 0.5) -- (-2.7, 0);

    \draw[\pgmLineColor,line width=0.75pt, dashed] (-2.7,-0.25) circle (0.25);
    \draw (-2.7,-0.25) node[anchor=center,align=center]{$\vec{p_{o}}$};

    \draw[line width=0.5pt,->] (-2.45, -0.25) -- (-1.95, -0.25);

    \draw[\pgmLineColor,line width=0.75pt, dashed] (-1.7,-0.25) circle (0.25);
    \draw (-1.7,-0.25) node[anchor=center,align=center]{$\vec{\sigma}_{\vec{\Tilde{y}}}$};

    \draw[line width=0.5pt,->] (0, 0.5) -- (0, 0);

    \fill[\pgmBgColor] (0,-0.25) circle (0.25);
    \fill[white] (-0.25,0) rectangle (0.0,-0.5);
    \draw[line width=0.75pt] (0,-0.25) circle (0.25);
    \draw (0,-0.25) node[anchor=center,align=center]{$\vec{\tilde{y}}$};

\end{tikzpicture}}
    \caption{BBN corresponding to the stochastic model used for the sparse measures example.}
    \label{fig:sparse_measure_pgm}
\end{figure}
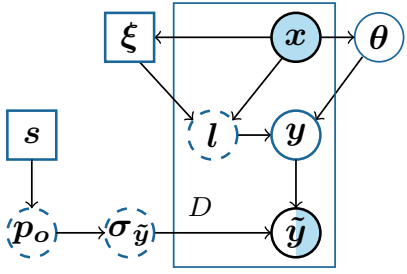

First,  the model can be extended to account for the \textbf{noisy labels}. To learn which output channel is actually more noisy than the others, a \textbf{meta-learning} approach can be used with \textbf{hierarchical Bayes} applied not on the model parameters but on the noise model. If applied to standard point estimate networks, this approach would be seen as learning a loss. The prior can be extended with a \textbf{consistency condition} to account for the fact that the unknown function projects points near a known hyperplane. Last, but not least, a \textbf{semi-supervised learning} model can be used to share information across the training samples. The PGM corresponding to this approach is depicted in Fig.~\ref{fig:sparse_measure_pgm}. We kept the plate for the dataset $D$ as we will use a second consistency condition to implement the semi-supervised learning strategy. We also chose to use a last layer-only BNN, mostly to illustrate how it can be done. For the hierarchy of noise quantification variables, $\vec{p_o}$ and $\vec{\sigma}_{\vec{\Tilde{y}}}$, only point estimates will be considered to simplify the model.

\begin{figure}[b]
    \centering
    \includegraphics[width=0.48\textwidth]{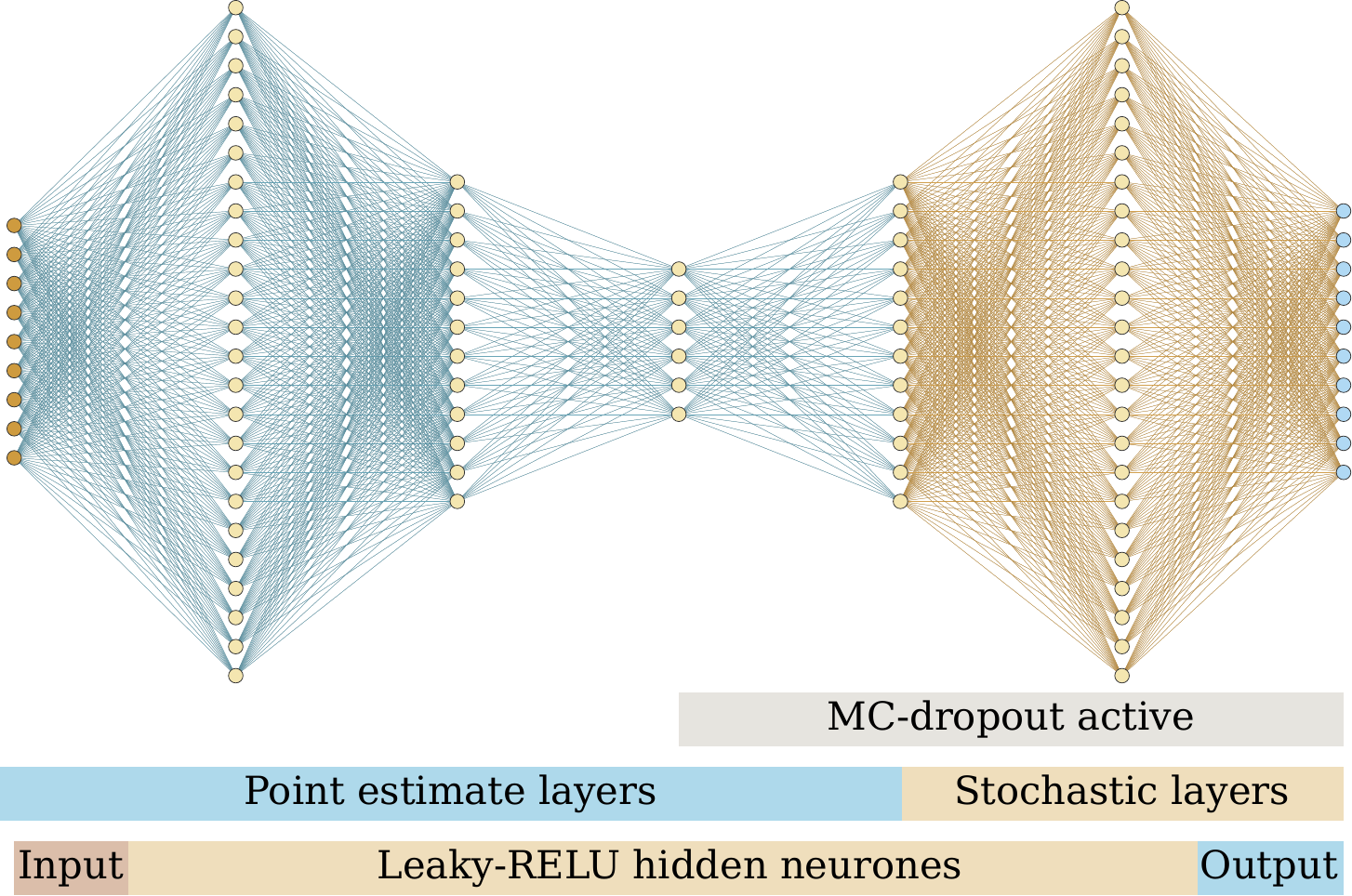}
    \caption{The neural network architecture used for the sparse measure example.}
    \label{fig:sparse_measure_functional}
\end{figure}

\textbf{The functional model} $\NN_{\vec{\xi}, \vec{\theta}}$ is a feedforward artificial neural network with five hidden layers (Fig.~\ref{fig:sparse_measure_functional}). The first four layers are point-estimate layers, with parameters $\vec{\xi}$, while the last hidden layer and the output layer are stochastic, with parameters $\vec{\theta}$. We also have three Monte-Carlo dropout layers at the end of the network. The hourglass shape of the network is supposed to match the expected behaviour of the studied function, where the information is approximately located on a lower dimensional space.

\subsection{Base stochastic model}

The base stochastic model is built as a Bayesian regression with a last-layer BNN architecture. The base probability of $\vec{\theta}$, without the consistency conditions that will be applied later on, has been set to a normal distribution with mean 0 and a standard deviation of $1$ for the weights and $10$ for the bias. The prior distribution of the output $\vec{y}$ knowing the input $\vec{x}$ and the parameters $\vec{\theta}$ and $\vec{\xi}$ is then given by:

\begin{equation}
    p(\vec{y} | \vec{x}, \vec{\theta}, \vec{\xi}) = \mathcal{N}(\NN_{\vec{\xi},\vec{\theta}}(\vec{x}),\sigma_{\vec{y}}).
\end{equation}

\noindent Here, $\sigma_{\vec{y}}$, which we set to $0.1$, represent the small uncertainty due to the fact that the functional model will never be able to perfectly fit the actual function $f$. 

\subsection{Noisy labels model}

For this regression problem, we assume that the measurements have been corrupted by a zero-mean Gaussian noise. This is not only a convenient formulation but also a reasonable approximation of the true noise model for continuous measurements (due to the central limit theorem). The mean of the noisy measurements $\vec{\Tilde{y}}$ is thus the unobserved, true function value $y$. For the standard deviation $\vec{\sigma}_{\vec{\Tilde{y}}}$, we assume that it is $\sigma_{in} = 0.1$ for inliers and $\sigma_{out} = 5$ for outliers. The problem here is that we do not know which channel has been corrupted by the additional noise. We could just set for all channels a constant standard deviation that is slightly above $\sigma_{in}$, but instead we added an additional variable, $\vec{p_o}$, a vector representing the probability that a given channel is an outlier. Since a single channel is corrupted by noise, we could constrain $\vec{p_o}$ to lie on the probability simplex. Instead, to let the model generalize to cases with more than a single noisy channel, we will impose that $\vec{p_o} \in [0,1]^m$. For convenience, we will consider an unconstrained variable $\vec{s} \in \mathbb{R}^m$ and define $\vec{p_o}$ as a function of $\vec{s}$ to simplify the optimisation:
\begin{equation}
    \vec{p_o} = \dfrac{1}{1+e^{-\vec{s}}}.
\end{equation}
\noindent The value of $\sigma$ is then determined by a Bernoulli distribution with parameter $\vec{p_o}$:
\begin{equation}
    \begin{array}{l}
         \vec{b} \sim \text{Bernoulli}(\vec{p_o}), \\
         \vec{\sigma}_{\vec{\Tilde{y}}} = \vec{b} \cdot \sigma_{out} + (1-\vec{b}) \cdot \sigma_{in}.
    \end{array}
\end{equation}
The only thing left is to determine the prior for $\vec{s}$. Assuming we have no idea which channel is noisy and which is not, a reasonable hypothesis is a Gaussian with mean $0$ and large covariance matrix $\matr{\Sigma}_{\vec{s}}$ (we choose $\matr{\Sigma}_{\vec{s}} = 2500\matr{I})$). Yet, since we know that only one channel is noisy, we can add a consistency condition to enforce that $\left\lVert \vec{p_o} \right\rVert_{1} \approx 1 $, or any expected number of noisy channels:
\begin{equation}
    p(\vec{s}) \propto \mathcal{N}(0,\matr{\Sigma}_{\vec{s}}) exp \left[ - \gamma_{s} \left( \left\lVert \dfrac{1}{1+e^{-\vec{s}}} \right\rVert_{1} - 1 \right)^2\right].
    \label{eq:sconsistentprior}
\end{equation}
\noindent Here, $\gamma_{s}$ is a scaling factor representing the confidence one has in the model capacity to fit the correct number of noisy channels. We set $\gamma_{s} = 3000$ as we are almost certain the model should be able to fit the single noisy channel. The probability of $\vec{\Tilde{y}}$ given $\vec{y}$ and $\vec{\sigma}_{\vec{\Tilde{y}}}$ is then given by:
\begin{equation}
    p(\vec{\Tilde{y}}|\vec{y},\vec{\sigma}_{\vec{\Tilde{y}}}) = \mathcal{N}(\vec{y}, \vec{\sigma}_{\vec{\Tilde{y}}}).
\end{equation}

\noindent Since we are not so interested in the complete posterior distribution for $\vec{s}$ or even $\vec{\sigma}_{\vec{\Tilde{y}}}$, we will learn point estimates of those variables instead and then derive a regularization term from the log of Equation~\eqref{eq:sconsistentprior}. The final contribution for the learnable parameters $s$ to the total loss is $\log(p(\vec{s}))$ plus an unknown constant that one can just ignore:
\begin{equation}
     \dfrac{1}{2} \left( \vec{s}^{\top} \matr{\Sigma}_{\vec{s}}^{-1} \vec{s} \right) + \gamma_{s} \left( \left\lVert \dfrac{1}{1+e^{-\vec{s}}} \right\rVert_{1} - 1 \right)^2.
\end{equation}

\noindent This is implemented in the |outlierAwareSumSquareError| learnable loss in the included Python implementation.

\subsection{Consistency condition}

The function $f$ is likely to map a given input near a known subspace of $\mathbb{R}^m$, which can be represented as $\text{span}(\matr{S})$ where $\matr{S} \in \mathbb{R}^{m \times 3}$ is a given orthonormal matrix. The distance $d_{\vec{y},\matr{S}}$ between a prediction $\vec{y}$ and $\text{span}(\matr{S})$ is then given by:
\begin{equation}
    d_{\vec{y},\matr{S}} = \left\lVert \vec{y} - \matr{S} \matr{S}^{\top} \vec{y} \right\rVert_{2}.
\end{equation}

\noindent We  assume a priori that $d_{\vec{y},\matr{S}}$ knowing $\vec{x}$ follows a normal distribution. We then use this to derive a consistency condition for $p(\vec{\theta}, \vec{\xi} | D_{\vec{x}})$:
\begin{equation}
    p(\vec{\theta}, \vec{\xi} | D_{\vec{x}}) \propto p(\vec{\theta})p(\vec{\xi}) e^{- \sum_{\vec{x} \in D_{\vec{x}}} \dfrac{1}{\sigma_d^2} \left\lVert \NN_{\vec{\xi}, \vec{\theta}}(\vec{x}) - \matr{S} \matr{S}^{\top} \NN_{\vec{\xi}, \vec{\theta}}(\vec{x}) \right\rVert^{2}_{2} }
\end{equation}

\noindent where $\sigma_d$ is the prior standard deviation of $d_{\vec{y},\matr{S}}$, which is set to $1$.

The additional contribution to the loss by the consistency condition is thus:
\begin{equation}
    \sum_{\vec{x} \in D_{\vec{x}}} \dfrac{1}{\sigma_d^2} \left\lVert \NN_{\vec{\xi}, \vec{\theta}}(\vec{x}) - \matr{S} \matr{S}^{\top} \NN_{\vec{\xi}, \vec{\theta}}(\vec{x}) \right\rVert^{2}_{2}.
\end{equation}

\subsection{Semisupervised learning strategy}

As stated in Section~IV-D1 of the main paper, there are two ways one can implement a semi-supervised learning strategy for a BNN. The first one is to use a consistency condition (\ie a data-driven regularization). The second one is to assume some kind of dependence across the samples. We use the former for this practical example as data-driven regularization is generally easier to implement.

The intuition behind this example is that if two points from the input space are close to one another then their image by $f$ should also be close to one another. The simplest approach to formally measure this is  to assume that the norm of the difference in the output space normalized by the distance in the input space is small. This sets the consistency condition function $C(\vec{\theta}, \vec{\xi}, \vec{x})$ to:
\begin{equation}
\label{eq:naiverelativedistance}
    C(\vec{\theta}, \vec{\xi}, \vec{x}) = \dfrac{1}{2} \sum_{\vec{x'} \in D_{\vec{x}} \setminus \left\{\vec{x}\right\}} \dfrac{\left\lVert  \NN_{\vec{\xi}, \vec{\theta}}(\vec{x}) - \NN_{\vec{\xi}, \vec{\theta}}(\vec{x'}) \right\rVert^{2}_{2}}{\left\lVert  \vec{x} - \vec{x'}\right\rVert^{2}_{2}}.
\end{equation}

\noindent Averaging this consistency condition over the whole training set $D_{\vec{x}}$ is equivalent to computing the sum over all output channels of $\NN_{\vec{\xi}, \vec{\theta}}$ of the quadratic form of the Laplacian matrix of the dense graph spanned by the training set $D_{\vec{x}}$, where the weights of the edges are given by the inverse Euclidean distances of the points. Graph Laplacian regularization is itself a common method to implement semi-supervised learning \cite{10.5555/1248547.1248632}.

\begin{figure}[tb]
    \centering
    \includegraphics[width=0.48\textwidth]{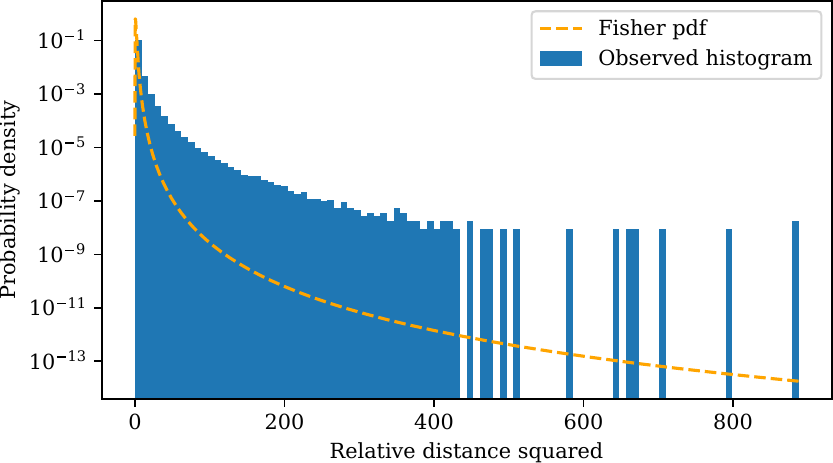}
    \caption{Comparison between the observed distribution of relative distances squared in the test set and the Fisher–Snedecor distribution of parameters 9 and 9.}
    \label{fig:fisherlaw}
\end{figure}

This approach can be further improved. Assume that the differences between the components of two random input vectors $\vec{x}$ and $\vec{x'}$, as well as the differences between the components of their images by $f$, are normally distributed, then:
\begin{equation}
    \dfrac{ \sigma_{\Delta x}^2 \left\lVert  f(\vec{x}) - f(\vec{x'}) \right\rVert^{2}_{2}}{ \sigma_{\Delta f}^2 \left\lVert  \vec{x} - \vec{x'}\right\rVert^{2}_{2}} \sim F(n,m),
\end{equation}
\noindent where $F(n,m)$ is a Fisher–Snedecor distribution with parameters $n$ and $m$, $\sigma_{\Delta x}$ is the prior standard deviation of the difference of two random inputs, and $\sigma_{\Delta f}$ is the prior standard deviation of the difference of the corresponding outputs. As shown in Fig.~\ref{fig:fisherlaw}, such assumptions are not perfect but still match the data reasonably well, especially since those assumptions are just a prior and not an exact model. The most important point to notice is that both the observed and prior distributions have a heavy tail. This would not be the case with the consistency condition proposed in Equation~\eqref{eq:naiverelativedistance}. This naive approach would  penalize outliers too heavily. We thus implemented the following, corrected, consistency condition function in the sparse measures example:

\begin{equation}
\label{eq:correctrelativedistance}
    \sum_{\vec{x'} \in D_{\vec{x}} \setminus \left\{\vec{x}\right\}} \left(\dfrac{n + m}{2} - 1\right) log(1+F) - \left(\dfrac{n}{2} - 1\right) \log(F) ,
\end{equation}

\noindent where $F = \lambda_{\Delta} \dfrac{\left\lVert  \NN_{\vec{\xi}, \vec{\theta}}(\vec{x}) - \NN_{\vec{\xi}, \vec{\theta}}(\vec{x'}) \right\rVert^{2}_{2}}{\left\lVert  \vec{x} - \vec{x'}\right\rVert^{2}_{2}}$ is the Fisher statistic. 

The only parameter one has to define for this prior is $\lambda_{\Delta}$, which is the ratio of $\sigma_{\Delta x}^2$ and $\sigma_{\Delta f}^2$. We set $\lambda_{\Delta} = 1$.

\subsection{Results}

\begin{figure}[tb]
    \centering
    \includegraphics[width=0.48\textwidth]{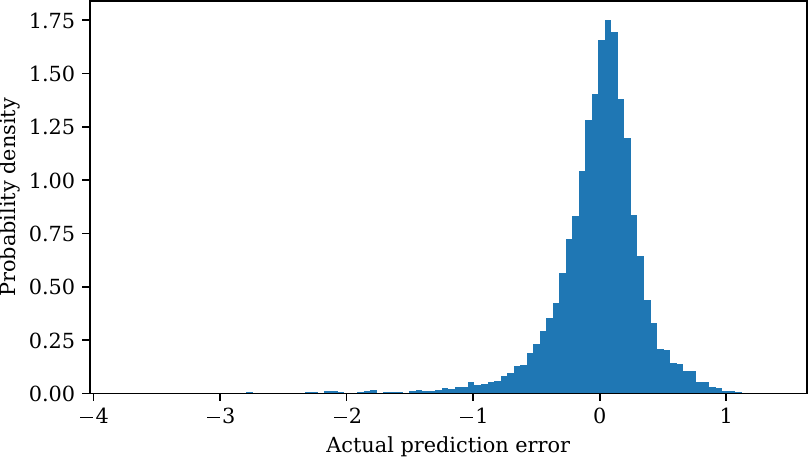}
    \caption{Prediction error distribution for the sparse measure example.}
    \label{fig:sparsemeserrordist}
\end{figure}

\begin{figure}[tb]
    \centering
    \includegraphics[width=0.4\textwidth]{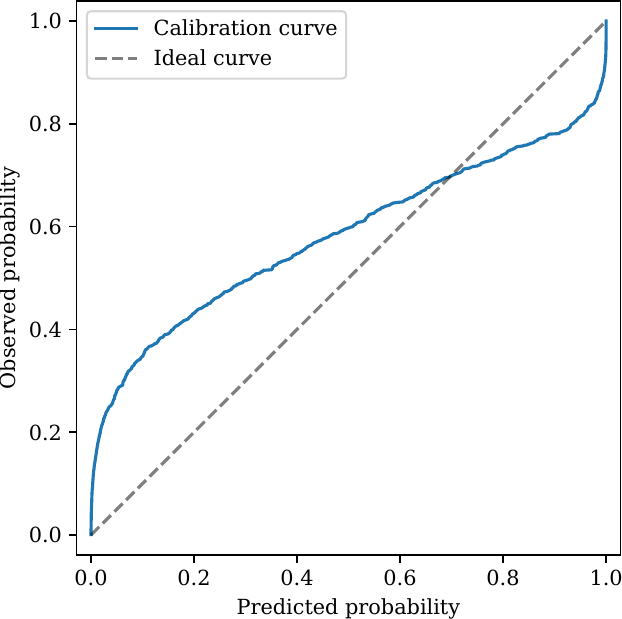}
    \caption{Calibration curve for the sparse measure example.}
    \label{fig:sparsemescalibcurve}
\end{figure}

The estimator for $\vec{y}$ reaches a RMSE of around $0.4$ with an error distribution approximating a normal distribution. Variations can be observed over mutliple runs, as a different function $f$ is generated at random each times (Fig.~\ref{fig:sparsemeserrordist}). The Python code also computes the calibration curve using the hypothesis  for regression models presented in Section~VII of the main paper. The results (Fig.~\ref{fig:sparsemescalibcurve}) show that the model tends to be slightly underconfident, except for large outliers where the model is overconfident. This is probably due to the fact that our simple variational inference model cannot fit the exact posterior and thus tends to be slightly too pessimistic about its predictions to compensate for the outliers it is unable to efficiently fit.

\section{Practical example -- Paperfold}
\label{toy_example:paperfold}

\begin{figure}[b]
    \centering
    \begin{subfigure}[t]{0.1\textwidth}
    \centering
        \includegraphics[height=2.3cm]{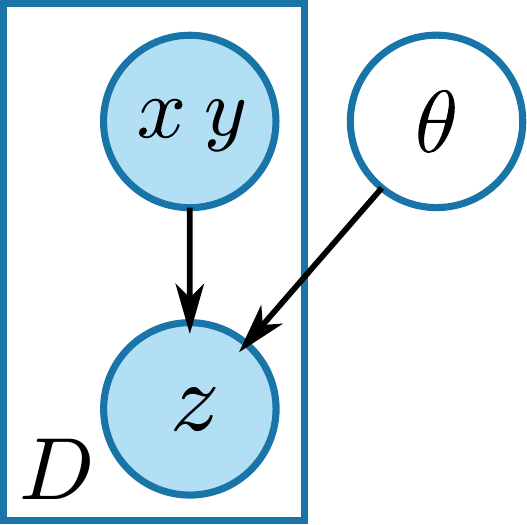}
        \caption{}
        \label{fig:models:stochastic}
    \end{subfigure}
    \hfill
    \begin{subfigure}[t]{0.34\textwidth}
    \centering
        \includegraphics[height=2.3cm]{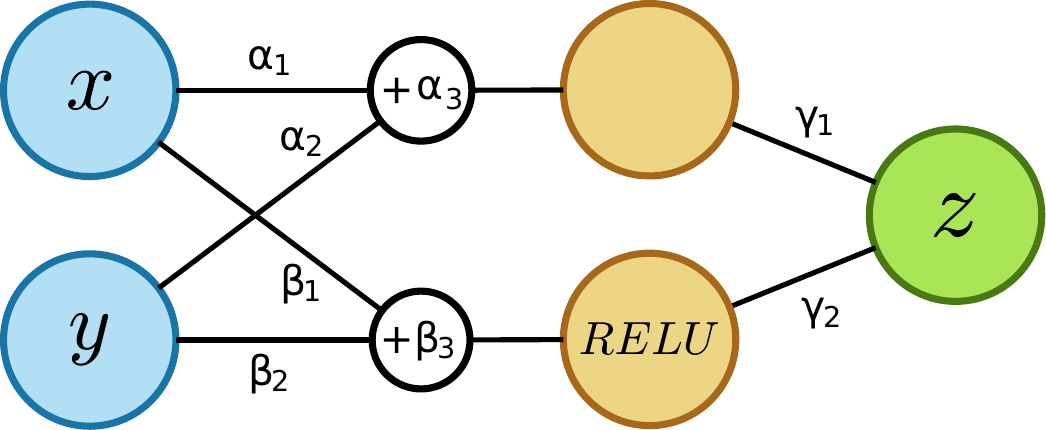}
        \caption{}
        \label{fig:models:functionnal}
    \end{subfigure}
    \caption{Stochastic (\subref{fig:models:stochastic}) and functional (\subref{fig:models:functionnal}) model for the paper-fold case study.}
    \label{fig:models}
\end{figure}

\begin{figure*}[t]
    \centering
    \begin{subfigure}[t]{0.32\textwidth}
        \includegraphics[width=\textwidth]{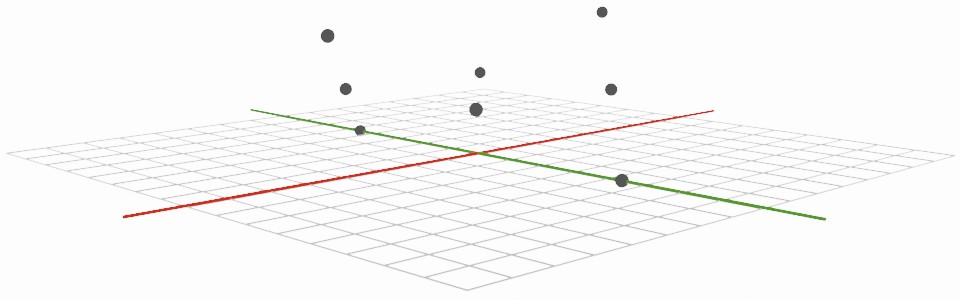}
        \caption{Raw data}
    \end{subfigure}
    \hfill
    \begin{subfigure}[t]{0.32\textwidth}
        \includegraphics[width=\textwidth]{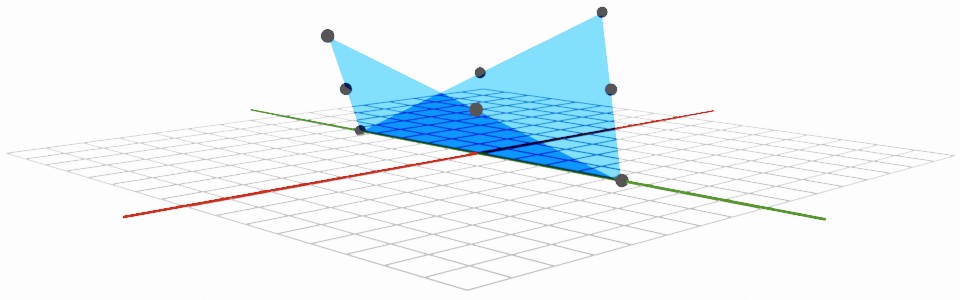}
        \caption{First possible fold}
    \end{subfigure}
    \hfill
    \begin{subfigure}[t]{0.32\textwidth}
        \includegraphics[width=\textwidth]{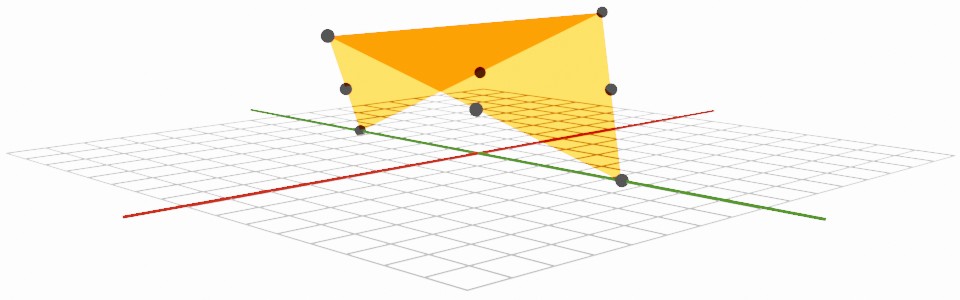}
        \caption{Second possible fold}
    \end{subfigure}
    \caption{The dataset and two possible folding behaviors of the model for the paperfold case study.}
    \label{fig:paperfold_fitting}
\end{figure*}

The last practical example we introduce is entitled ''Paperfold''. The code is available on github [\url{https://github.com/french-paragon/paperfold-bnn}].

\textbf{The purpose} of this case study is to compare different inference approaches for BNNs. We introduce a model, and a corresponding dataset, which is small enough to make all existing training methods for BNNs tractable when used for the corresponding architecture. This makes the number of parameters small enough such that the posterior and the corresponding approximations are low dimensional. Thus, they can be easily visualized and sampled without issues using exact MCMC methods. On the other hand, the model is complex enough to display most of the issues one would encounter when studying a BNN posterior.

\textbf{The problem} is to fit a one dimensional function of two parameters of the form $z = f(x,y)$. 

\textbf{The dataset} is made of eight $(x,y,z)$ samples grouped into the vectors $\vec{x},\vec{y},\vec{z}$ with:

\begin{equation}
    \vec{x} = \left( \begin{array}{r}
    \sqrt{2} \\
    \dfrac{1}{2}\sqrt{2} \\
    0 \\
    -\dfrac{1}{2}\sqrt{2} \\
    -\sqrt{2} \\
    -\dfrac{1}{2}\sqrt{2} \\
    0 \\
    \dfrac{1}{2}\sqrt{2}
    \end{array}\right), ~ \vec{y} = \left( \begin{array}{r}
    0 \\
    \dfrac{1}{2}\sqrt{2} \\
    \sqrt{2} \\
    \dfrac{1}{2}\sqrt{2} \\
    0 \\
    -\dfrac{1}{2}\sqrt{2} \\
    -\sqrt{2} \\
    -\dfrac{1}{2}\sqrt{2}
    \end{array}\right),~ \vec{z} = \left( \begin{array}{r}
    1 \\
    \dfrac{1}{2} \\
    0 \\
    \dfrac{1}{2} \\
    1 \\
    \dfrac{1}{2} \\
    0 \\
    \dfrac{1}{2}
    \end{array}\right).
\end{equation}

\textbf{The stochastic model} (Fig.~\ref{fig:models:stochastic}) is the usual fully supervised Baysian regression. We assume each measure in $\vec{z}$ has a small uncertainty $\varepsilon_i \sim \mathcal{N}(0,\sigma_{\vec{z}}^2)$, with $\sigma_{\vec{z}}$ being a small positive standard deviation that one can set depending on the experiment. By default, we assume that $\sigma_{\vec{z}} = 0.1$.

\textbf{The functional model} (Fig.~\ref{fig:models:functionnal}) is a two-branch and two-layer feedforward neural network. The first layer branch has no non-linearities afterwards, while the second branch is followed by a $RELU$ non-linearity. The second layer simply takes the weighed sum of both branches and thus has no bias. This model reduces to a function of the form:
\begin{equation}
    z = \gamma_1 (\alpha_1 x + \alpha_2 y + \alpha_3) + \gamma_2 \text{RELU}(\beta_1 x + \beta_2 y + \beta_3).
\end{equation}

\noindent Here,  $\vec{\alpha}$ represents the parameters of the first branch of the first layer, $\vec{\beta}$ the parameters of the second branch of the first layer, and $\vec{\gamma}$ the parameters of the second layer. This simple model represents a planar function with a single fold along the line of equation $\beta_1 x + \beta_2 y + \beta_3 = 0$.

\begin{figure*}[b]
    \centering
    \includegraphics[width=\textwidth]{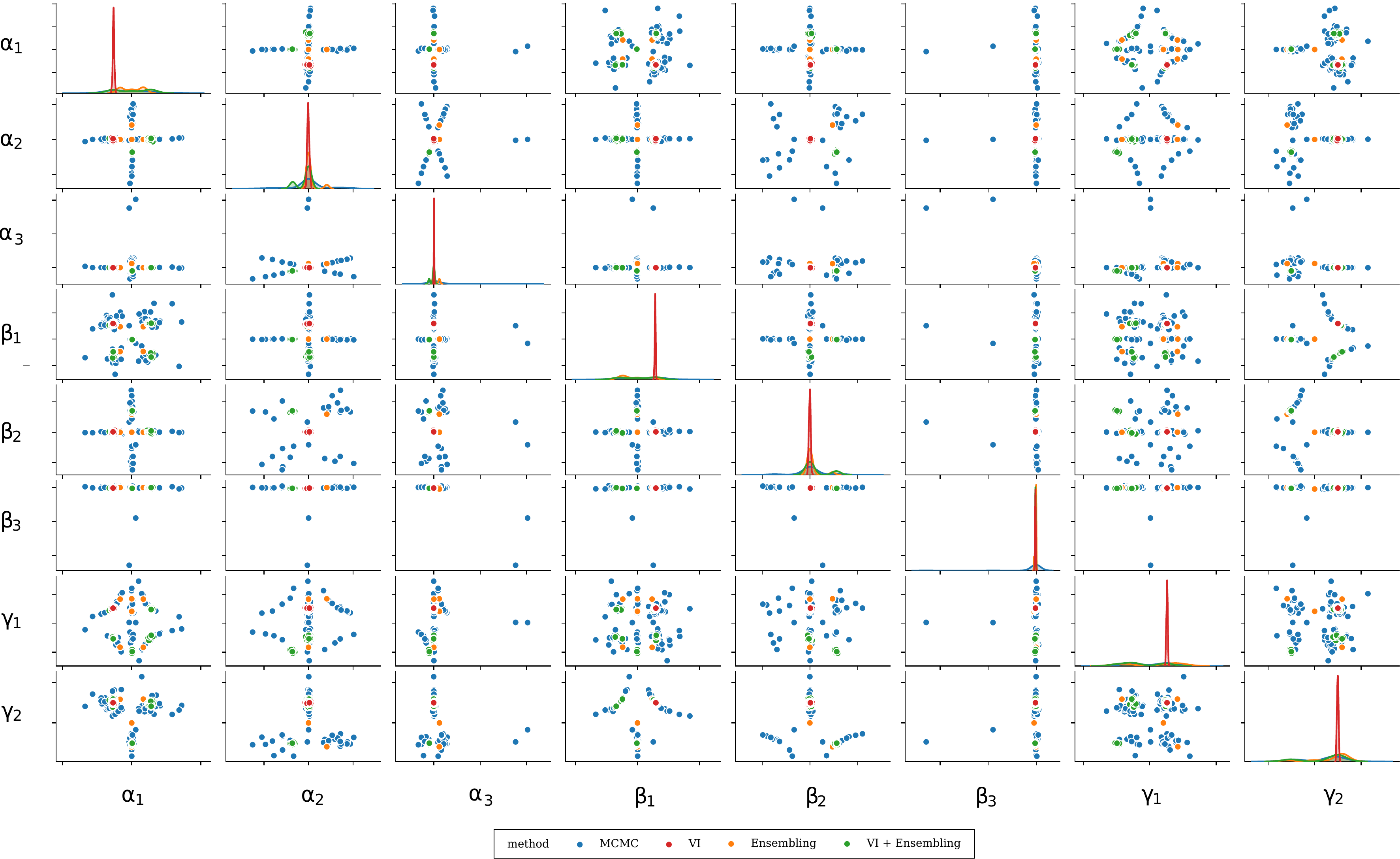}
    \caption{Pairplot of the samples from the posterior using four different approaches.}
    \label{fig:posterior_samples}
\end{figure*}

This model has always two and only two non-equivalent ways of fitting the proposed dataset (see Fig.~\ref{fig:paperfold_fitting}). The model also exhibits some weight-space symmetry. In other words, the half planes fitted by the first and second branches of the first layer can be swapped, resulting in a different parametrization but equivalent fitting of the data. It also exhibits a scaling symmetry between $\vec{\beta}$ and $\vec{\alpha}$ against $\vec{\gamma}$.

To sample the posterior, we first have to choose a prior for the parameters $\vec{\alpha}$, $\vec{\beta}$ and $\vec{\gamma}$ of the BNN. The following procedure works for any choice of prior $p(\vec{\alpha}, \vec{\beta}), p(\vec{\gamma})$ where the probability distribution of the parameters of the first layer is independent of the probability distribution of the parameters of the second layer. For simplicity and if not specified otherwise, we assume a normal prior with diagonal constant covariance:
\begin{equation}
    \left( \begin{array}{r}
         \vec{\alpha}\\
         \vec{\beta} \\ 
         \vec{\gamma}
    \end{array} \right) \sim \mathcal{N}(0,\sigma^2 \matr{I}),
\end{equation}

\noindent with $\sigma$ a positive standard deviation. By default, we assume that we do not know much about the model and set $\sigma = 5$ for this case study.

\subsection{Comparison of training methods}

\begin{figure*}[p]
    \centering
    
    \begin{subfigure}[t]{0.48\textwidth}
        \includegraphics[width=0.47\textwidth]{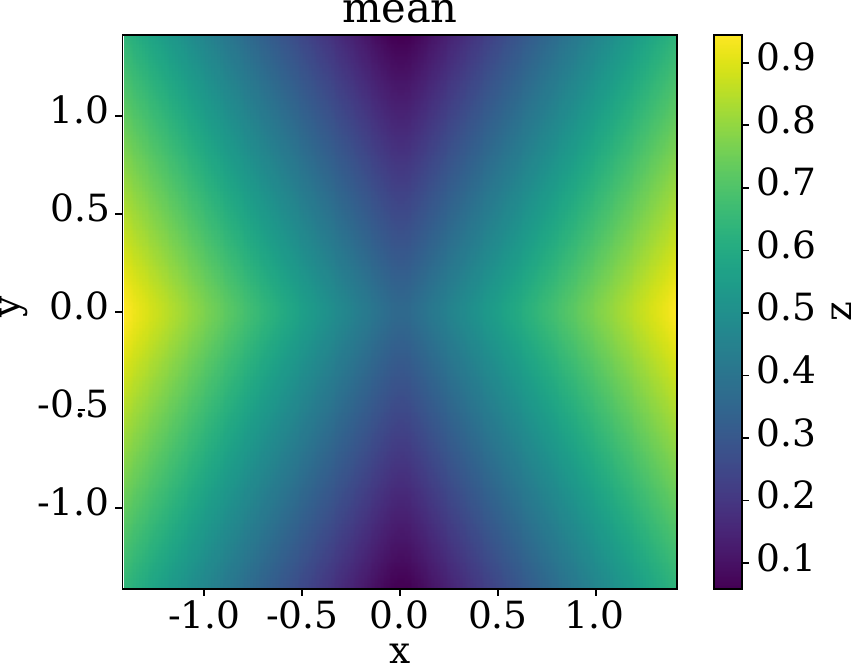}
        \hfill
        \includegraphics[width=0.47\textwidth]{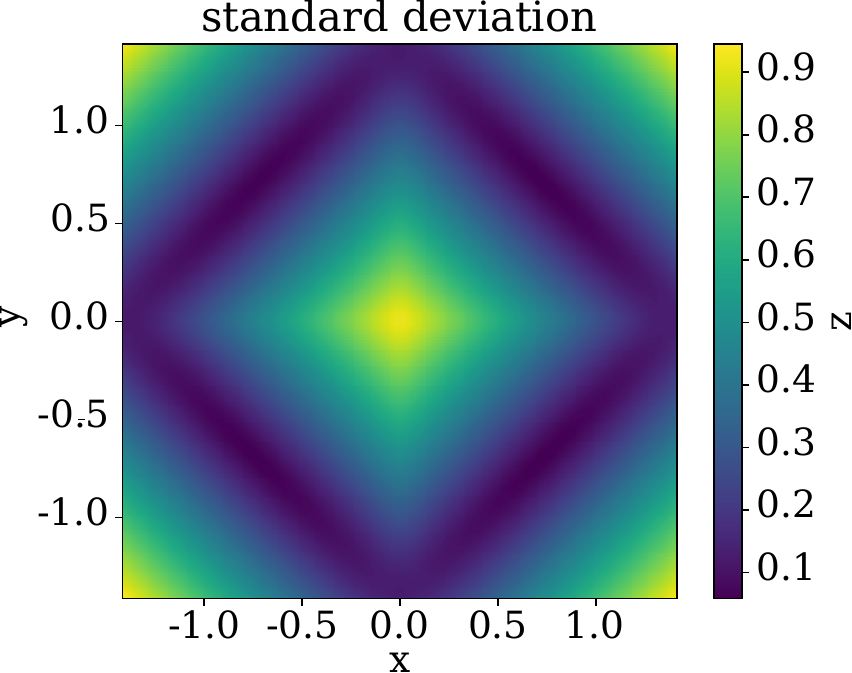}
        \caption{}
        \label{fig:paperfoldmarginals:mcmc_results}
    \end{subfigure}
    \hfill
    \begin{subfigure}[t]{0.48\textwidth}
        \includegraphics[width=0.47\textwidth]{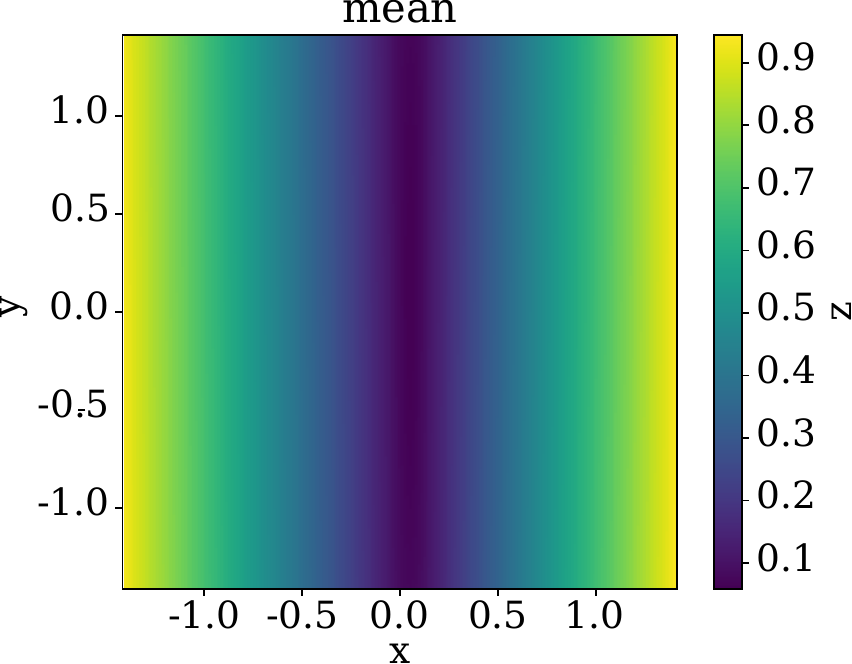}
        \hspace{5pt}
        \includegraphics[width=0.47\textwidth]{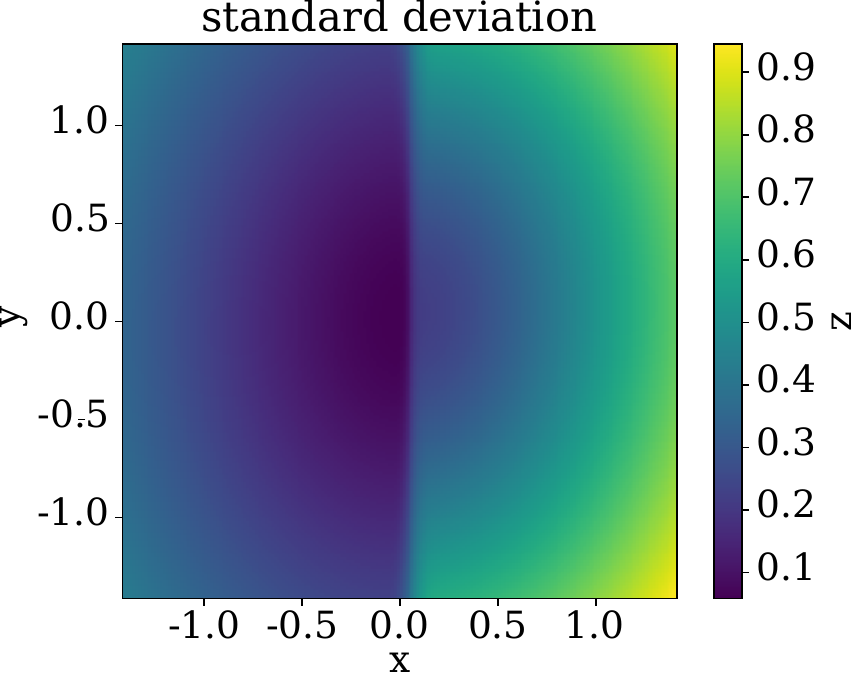}
        \caption{}
        \label{fig:paperfoldmarginals:vi_results}
    \end{subfigure}
    
    \begin{subfigure}[t]{0.48\textwidth}
        \includegraphics[width=0.47\textwidth]{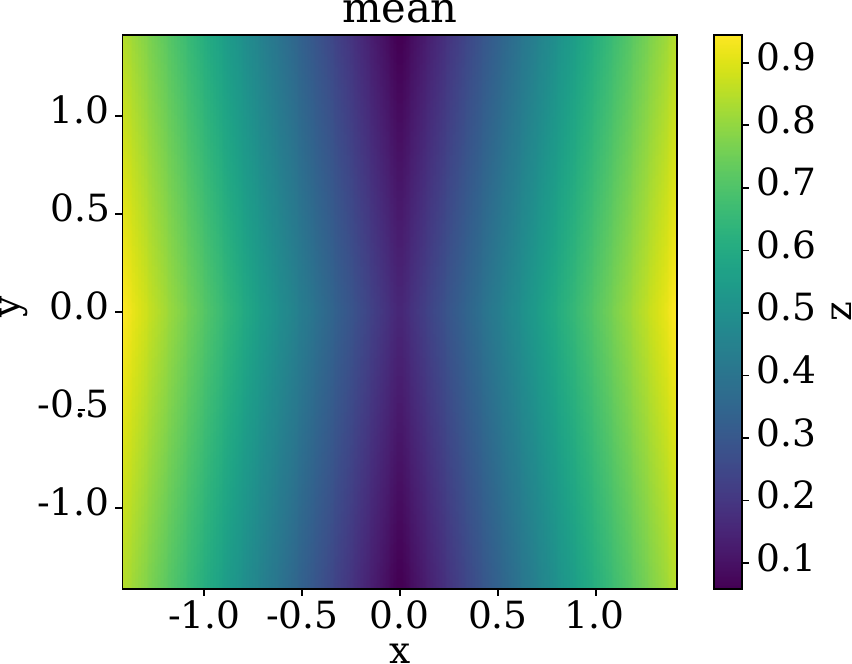}
        \hfill
        \includegraphics[width=0.47\textwidth]{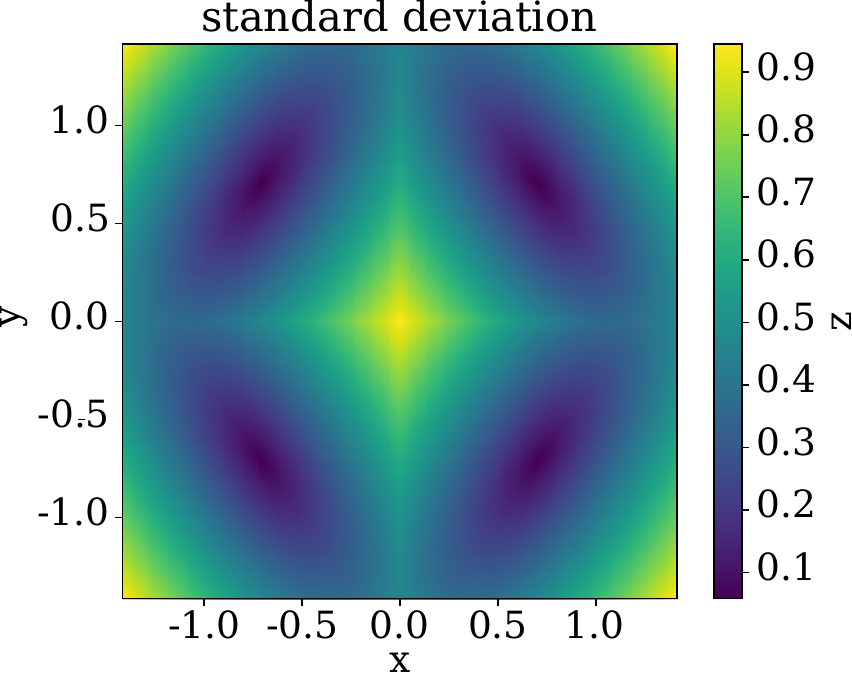}
        \caption{}
        \label{fig:paperfoldmarginals:ensembles_results}
    \end{subfigure}
    \hfill
    \begin{subfigure}[t]{0.48\textwidth}
        \includegraphics[width=0.47\textwidth]{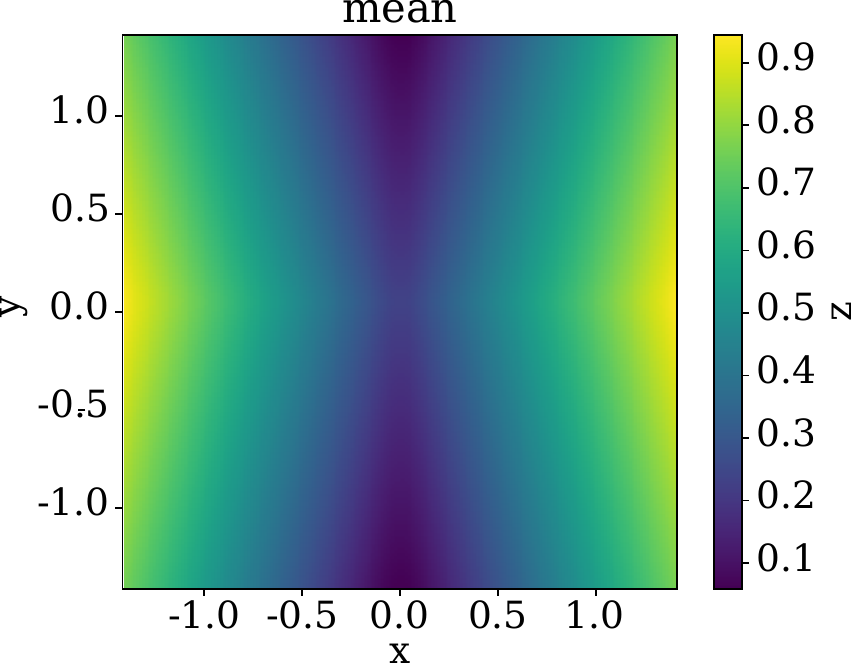}
        \hfill
        \includegraphics[width=0.47\textwidth]{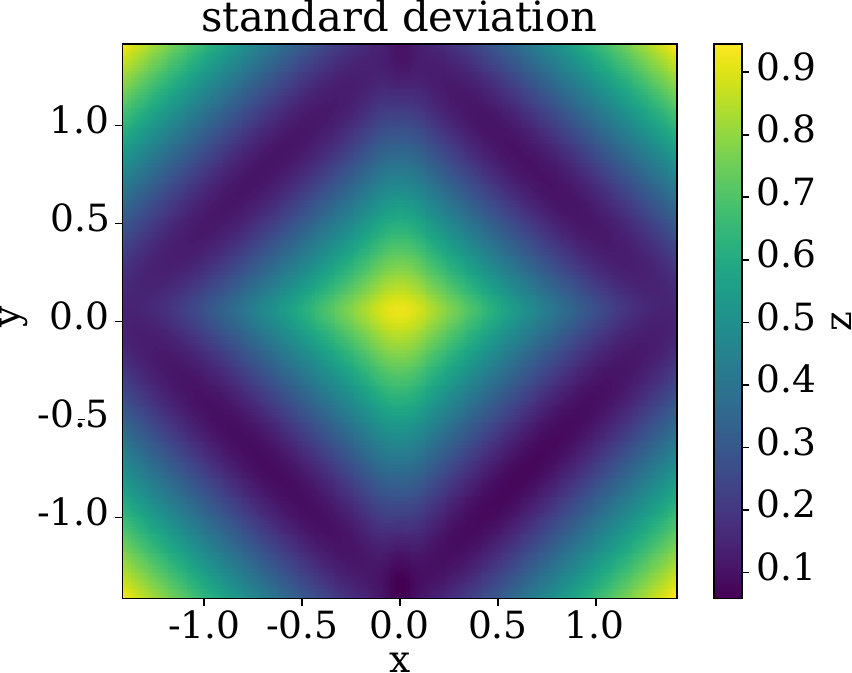}
        \caption{}
        \label{fig:paperfoldmarginals:vi_ensembles_results}
    \end{subfigure}
    
        \caption{Mean and standard deviation of the marginal distribution of $z$ knowing $D$, $x$ and $y$ for the paperfold model using (\subref{fig:paperfoldmarginals:mcmc_results}) MCMC, (\subref{fig:paperfoldmarginals:vi_results}) variational inference, (\subref{fig:paperfoldmarginals:ensembles_results}) ensembling and (\subref{fig:paperfoldmarginals:vi_ensembles_results}) variational inference + ensembling.}
        \label{fig:paperfoldmarginals}
\end{figure*}

\begin{figure*}[p]
    \centering
    \begin{subfigure}[t]{0.44\textwidth}
        \includegraphics[width=\textwidth]{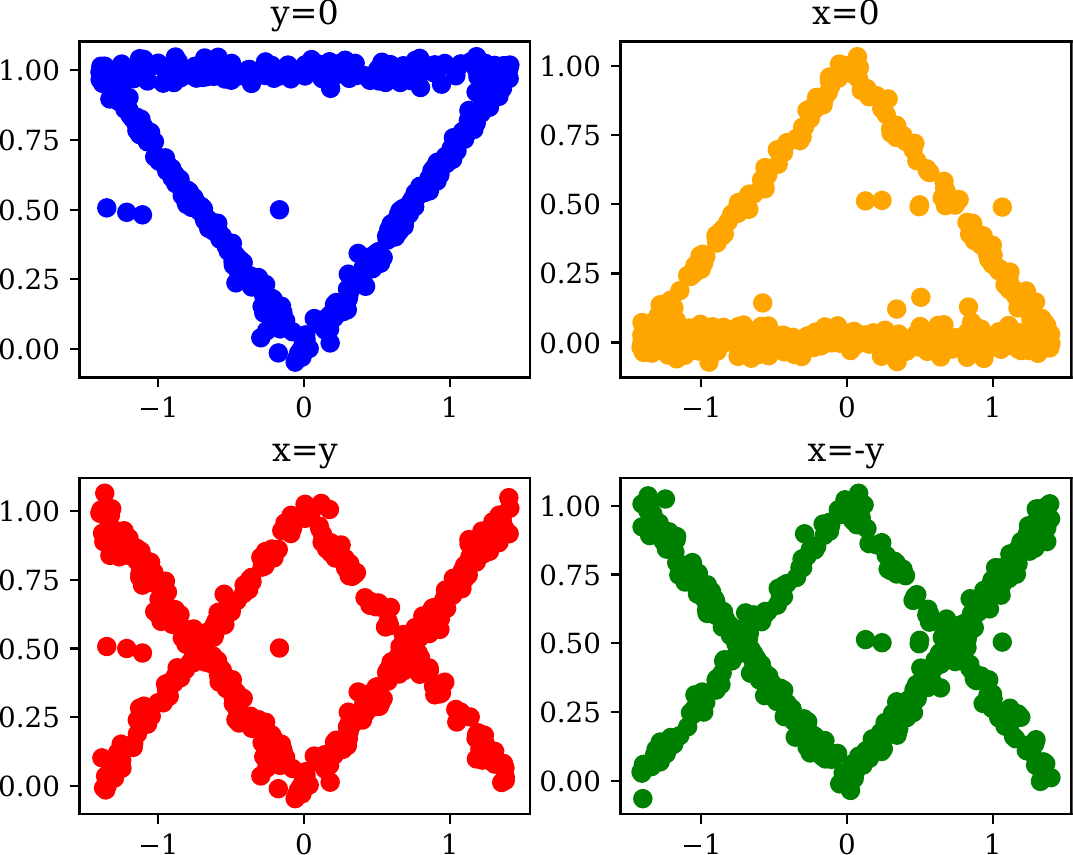}
        \caption{}
        \label{fig:paperfoldcuts:mcmc_results}
    \end{subfigure}
    \hfill
    \begin{subfigure}[t]{0.44\textwidth}
        \includegraphics[width=\textwidth]{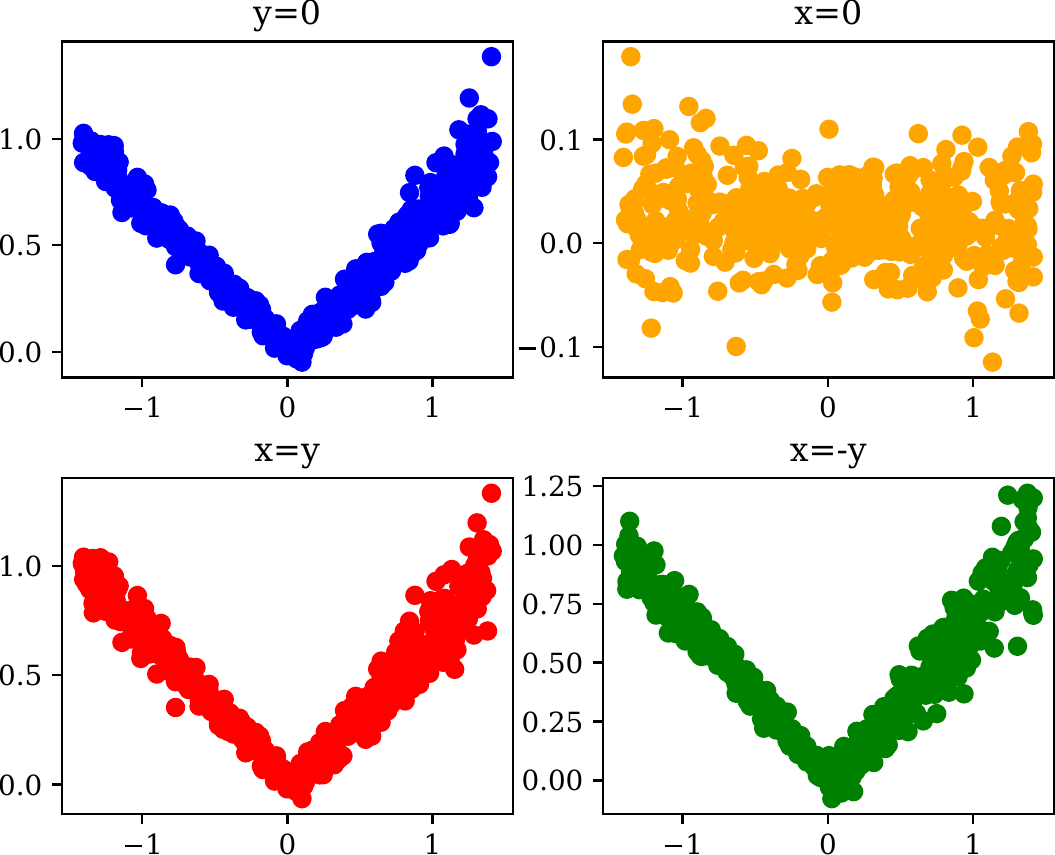}
        \caption{}
        \label{fig:paperfoldcuts:vi_results}
    \end{subfigure}
    
    \begin{subfigure}[t]{0.44\textwidth}
        \includegraphics[width=\textwidth]{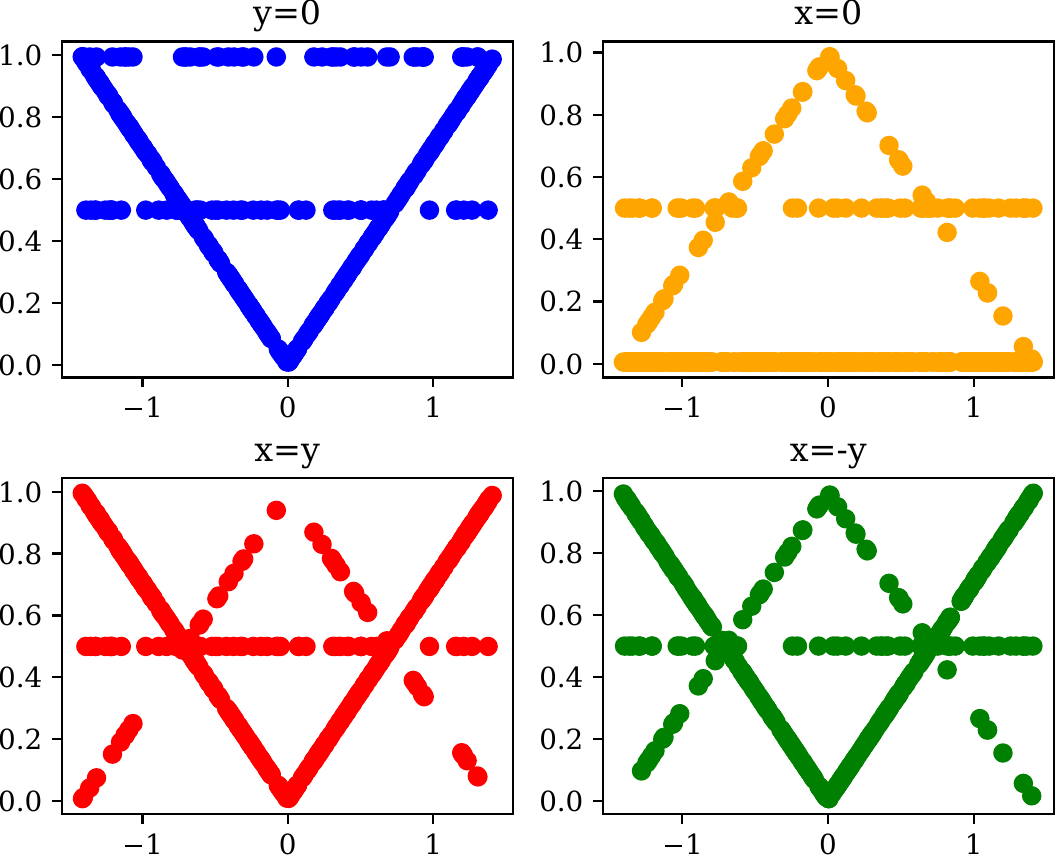}
        \caption{}
        \label{fig:paperfoldcuts:ensembles_results}
    \end{subfigure}
    \hfill
    \begin{subfigure}[t]{0.44\textwidth}
        \includegraphics[width=\textwidth]{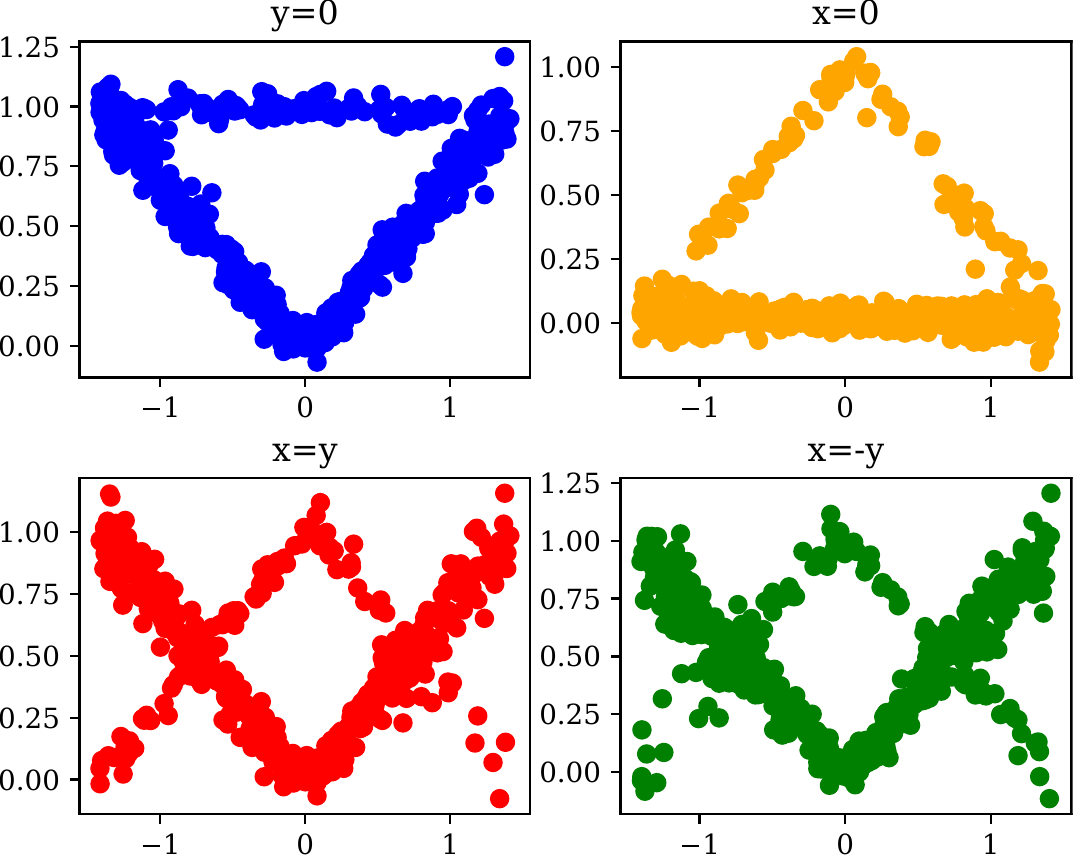}
        \caption{}
        \label{fig:paperfoldcuts:vi_ensembles_results}
    \end{subfigure}
    
    \caption{Predicted $z$ values across four different lines for the (\subref{fig:paperfoldcuts:mcmc_results}) MCMC, (\subref{fig:paperfoldcuts:vi_results}) variational inference, (\subref{fig:paperfoldcuts:ensembles_results}) ensembling and (\subref{fig:paperfoldcuts:vi_ensembles_results}) variational inference + ensembling version of the paperfold BNN.}
    \label{fig:paperfoldcuts}
\end{figure*}

\begin{figure}
    \centering
    \begin{subfigure}[t]{0.48\textwidth}
    \centering
        \includegraphics[width=\textwidth]{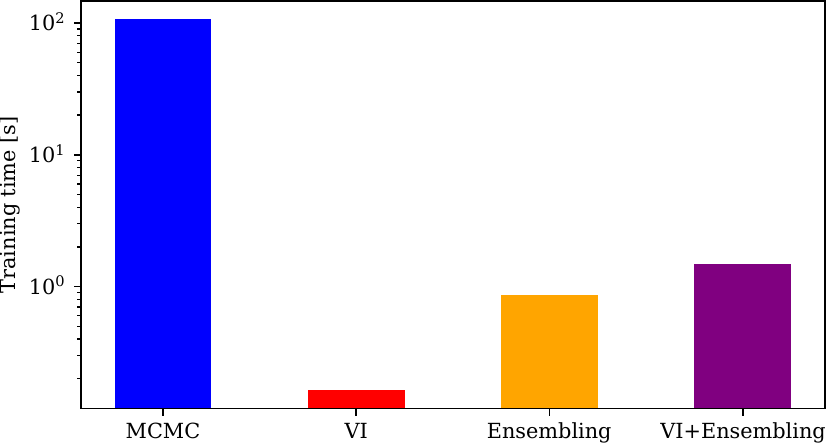}
        \caption{}
        \label{fig:statefiescomp:traintime}
    \end{subfigure}
    
    \begin{subfigure}[t]{0.48\textwidth}
    \centering
        \includegraphics[width=\textwidth]{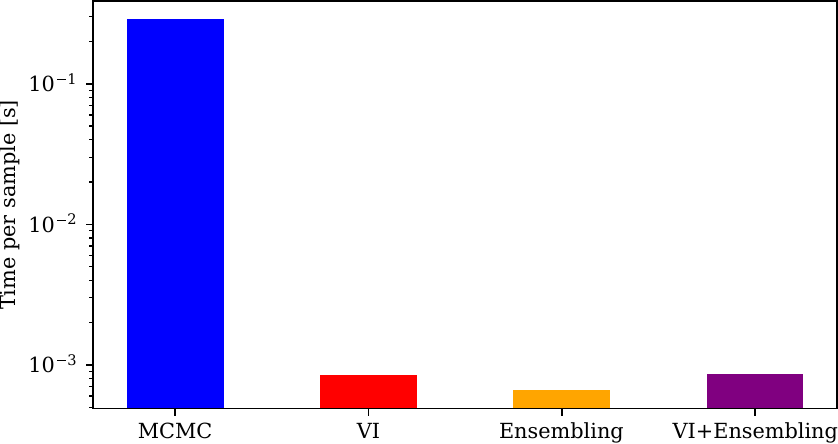}
        \caption{}
        \label{fig:statefiescomp:sampletime}
    \end{subfigure}
    \caption{Comparison between (\subref{fig:statefiescomp:traintime}) training time and (\subref{fig:statefiescomp:sampletime}) inference time for the paperfold example with standard algorithms for the different training approaches.}
    \label{fig:statefiescomp}
\end{figure}

We implemented four different samplers for the paperfold example (Fig.~\ref{fig:posterior_samples}):
\begin{itemize}
    \item The NUTS sampler \cite{hoffman2014no}, whose implementation is provided by the Pyro Python library. This is a state-of-the-art general purpose MCMC sampler, which was used mainly as a way of generating samples from the true posterior. It should serve as a base of comparison with other methods, because this MCMC sampler is much slower than the other methods we consider (Fig.~\ref{fig:statefiescomp}).
    
    \item A variational inference model based on a Gaussian approximation of the posterior.
    
    \item An ensemble based approach, and 
    
    \item An ensemble of variational inference based models. 
\end{itemize}

\subsubsection{MCMC}

The MCMC sampler provides a very good approximation of samples from the exact posterior. In Fig.~\ref{fig:posterior_samples}, one can get an appreciation of the complexity of the exact posterior even for this small and simple model. Each row and column represent one variable with the graph at the intersection showing the samples projected in the 2-dimensional plane spanned by those variables. The samples form perpendicular and diagonal structures, which probably correspond to the two non-equivalent ways of fitting the data. The symmetric lines are caused by the weight-space symmetry. The hyperbolas correspond to equivalent parametrizations of the network under scaling symmetry. Finally, the few outliers that are visible correspond to a series of small modes in the posterior around the parametrizations fitting the constant function $f(x,y) = 0.5$.

When looking at the actual aggregate results (Fig.~\ref{fig:paperfoldmarginals:mcmc_results}), one can see that the average prediction of the BNN fits the data with a regular function similar to a second degree polynomial. The uncertainty, as measured by the standard deviation a posteriori, is low along the lines of the square where the data points lie and increases linearly with distance, creating a diamond shape. Sampling the predicted value for $z$ along a series of lines in the $(x,y)$ plane of coordinates (Fig.~\ref{fig:paperfoldcuts:mcmc_results}) shows that both folds appear as clearly distinct from one another, but a bit of uncertainty remains around the exact position of these folds.

\subsubsection{Variational inference}

As a variational inference approximation to the posterior, we used a normal distribution with a learnable mean and a diagonal covariance matrix. It is pretty clear that for this specific example, which has been designed to generate such a specific problem, this method leads to a very poor approximation of the actual posterior since it can only fit an unimodal distribution (Fig.~\ref{fig:posterior_samples}). In terms of the marginal, only one of the two possible folds (Fig.~\ref{fig:paperfoldcuts:vi_results}) has been fitted. This translates to a mean estimate for $z$ as well as an uncertainty level and distribution well off the actual posterior (Fig.~\ref{fig:paperfoldmarginals:vi_results}). The actual side of the fold fitted by the RELU branch is also visible as its uncertainty is higher than on the other side of the fold. This highlights the major limitations of simple variational inference methods for BNNs. Since  the underlying models are complex and the data can be fitted in many non-equivalent manners, simple variational posteriors lack some expressive power to provide a good estimate of the actual BNN uncertainty. It is still important to keep in mind that this example has been specifically designed  to make those problems apparent and that, in practice, mean field Gaussian and similar variational inference approximations can still lead to good estimations of the actual uncertainty. An example of this is the sparse measure example provided in Appendix~\ref{toy_example:sparse_measure}. 
Variational inference is also several orders of magnitude faster than MCMC (Fig.~\ref{fig:statefiescomp}).

\subsubsection{Ensembling}

The most straightforward way to perform ensemble learning with an artificial neural network architecture is just to restart the learning procedure multiple times and each time add the final model to the ensemble. Using this strategy to learn the posterior for the paperfold example already gives quite good results, even better than naive variational inference as the samples can belong to different modes of the posterior (Fig.~\ref{fig:posterior_samples}). The marginal mean and standard deviation a posteriori are clearly different from the ones obtained via MCMC, but the diamond shape can be recongnized in both predictions (Fig.~\ref{fig:paperfoldmarginals:ensembles_results}). The training is slightly longer than with basic variational inference as it has to be run multiple times. However, the time per sample when computing the marginal is extremely low (Fig.~\ref{fig:statefiescomp}). The main drawback of this approach is that it provides no estimation of the local uncertainty around the different modes of the posterior as shown by Fig.~\ref{fig:paperfoldcuts:ensembles_results}. 

\subsubsection{Variational Inference + Ensembling}

Combining the benefits of a simple variational inference and ensembling can be done in a straightforward manner by  learning an ensemble of variational approximations. The resulting distribution is a Gaussian mixture, which can approximate the shape of multiple modes of the posterior. In this example, it cannot match the complex shapes of the exact posterior (Fig.~\ref{fig:posterior_samples}). Despite those limitations, the resulting marginal for $z$ is a very good match with the one generated via a MCMC sampler (Fig.~\ref{fig:paperfoldmarginals} and \ref{fig:paperfoldcuts}). In addition, the time per sample when computing the marginal is still very similar to the single Gaussian variational approximation (Fig.~\ref{fig:statefiescomp}). This shows the huge benefits of ensembling as a tool for approximate Bayesian methods in deep learning.

{\small
\bibliographystyle{IEEEtran}
\bibliography{refs}
}